\documentclass[preprint,12pt,nopreprintline]{elsarticle}
\PassOptionsToPackage{hyphens}{url}
\usepackage{times}
\usepackage[T1]{fontenc}
\usepackage{microtype}
\usepackage{graphicx}
\usepackage{amsmath,amssymb}
\usepackage{newtxmath}
\usepackage{hyphenat}
\AddToHook{env/abstract/begin}{\tolerance=700\emergencystretch=.15em}
\AddToHook{cmd/thebibliography/after}{\tolerance=1000\emergencystretch=2em}
\usepackage{array,booktabs}
\usepackage{placeins}
\usepackage{needspace}
\usepackage{url}
\usepackage[hidelinks]{hyperref}
\usepackage{longtable,fancyvrb}

\makeatletter
\newenvironment{keeptable}{\par\vspace{4pt}\noindent\begin{minipage}{\linewidth}\def\@captype{table}\setlength{\abovecaptionskip}{2pt}\centering}{\end{minipage}\par\vspace{2pt}}
\makeatother

\newcommand{\suppsection}[1]{\ref{#1}}
\biboptions{sort&compress}
\hypersetup{pdftitle={From Search to Signal: Search-Value Credit for Evolutionary Automatic Heuristic Design},pdfauthor={Yilun Yuan, Tianyu Zhou, Zhenzhou Tang},pdfsubject={Preprint with appendices A--E}}
\newcommand{\FirstAuthorName}{Yilun Yuan}
\newcommand{\SecondAuthorName}{Tianyu Zhou}
\newcommand{\CorrespondingAuthorName}{Zhenzhou Tang}
\newcommand{\FirstAuthorEmail}{25451354046@stu.wzu.edu.cn}
\newcommand{\SecondAuthorEmail}{25451354054@stu.wzu.edu.cn}
\newcommand{\CorrespondingAuthorEmail}{tzz@wzu.edu.cn}
\newcommand{\AuthorInstitution}{Wenzhou University}
\newcommand{\AuthorCity}{Wenzhou}
\newcommand{\AuthorPostcode}{325035}
\newcommand{\AuthorCountry}{China}

\begin{document}
\begin{frontmatter}
\title{From Search to Signal: Search-Value Credit for Evolutionary Automatic Heuristic Design}
\author[aff1]{\FirstAuthorName}
\ead{\FirstAuthorEmail}
\author[aff1]{\SecondAuthorName}
\ead{\SecondAuthorEmail}
\author[aff1]{\CorrespondingAuthorName\corref{cor1}}
\ead{\CorrespondingAuthorEmail}
\affiliation[aff1]{organization={\AuthorInstitution},
  city={\AuthorCity}, postcode={\AuthorPostcode}, country={\AuthorCountry}}
\cortext[cor1]{Corresponding author.}

\begin{abstract}
Large language models generate candidate programs for automatic heuristic design, while evolutionary search evaluates and selects them to guide further generation. Online post-training uses rewards from candidate evaluations to adapt the generator. Retained programs can also guide later generation as parents. This motivates rewards that reflect their reuse opportunities beyond immediate fitness. We propose Search-value credit (SVC), which constructs response-level rewards from actual admission to the retained set and a short-term parent-exposure proxy. Admission determines reward eligibility; the proxy weights eligible responses by reuse opportunities under existing parent-selection rules. Rewards are computed after candidate evaluation, without evaluating descendants. SVC retains the search and learning procedures of the Co-Evolution of Algorithms and Language Model (CALM) framework. Across four combinatorial optimization tasks with the same base model and response budget, SVC achieves the best mean final search performance on three tasks among six conditions. Under identical prompts and parents on traveling salesman and capacitated vehicle routing tasks, SVC-trained generators yield a higher mean proportion of responses whose programs outperform the best prompting parent than those trained with CALM's native reward. These findings support reward construction using search retention and selection information.

\end{abstract}
\end{frontmatter}
\section{Introduction}
\label{sec:introduction}
Automatic heuristic design (AHD) searches for programs that implement decision rules or components of optimization algorithms \citep{burke2013hyperheuristics}. Large language models (LLMs) can generate candidate programs using task descriptions and previously evaluated programs as context. FunSearch, EoH and ReEvo evaluate generated programs and use the results to guide subsequent proposals through program databases, evolving populations or reflection \citep{romeraparedes2024funsearch,liu2024eoh,ye2024reevo}. The resulting heuristics can be applied to multiple problem instances. These methods keep the LLM parameters fixed during search and adapt subsequent prompts through the selection of programs and evaluation feedback.

Online post-training updates the generator's parameters as search proceeds. EvoTune trains generation policies using preferences collected during evolutionary exploration, and CALM combines evolutionary heuristic search with Group Relative Policy Optimization (GRPO) \citep{surina2025evotune,huang2026calm,shao2024deepseekmath}. In CALM, the search selects parent programs and variation instructions, and the LLM generates candidate programs from the resulting prompt. The candidates are evaluated both to decide which programs to retain and to construct rewards for training the LLM. These rewards guide parameter updates, after which the model generates further candidates for the search. We study how to construct training rewards from the evaluations and search context so that the generator can better support subsequent search.

We aim to make these rewards reflect a candidate's potential use in subsequent search. CALM already uses generation context by comparing candidate performance with that of the best prompting parent \citep{huang2026calm}. Improvement over a parent, admission to the retained set, and opportunities to participate in later generation describe different properties. In the inherited search, admission depends on whether an eligible candidate's fitness is already represented in the retained set. A candidate can be admitted without improving its parents, and a retained program outside the highest-fitness parent pool can still be selected through the idea-distance route of crossover. Conversely, a candidate that improves a weak parent can repeat a retained fitness value and be rejected. These possibilities motivate using admission and parent-selection information when constructing training feedback. Section~\ref{sec:admission} shows that the two events disagree frequently in the inherited search, so a reward built on parent-relative improvement alone would ignore this separation.

We propose \textbf{Search-value credit (SVC)}, which constructs training rewards from actual candidate admission and a short-term parent-exposure proxy, with the aim of making generator feedback reflect candidates' potential use in subsequent search. Actual admission gates credit to eligible responses, while the proxy weights eligible responses by their opportunities for reuse as parents under the inherited selection rules. The proxy is computed by inserting each candidate separately into the retained set before the current group is processed. It includes the primary-parent route and both fitness-based and idea-distance-based routes for the second parent in crossover. SVC uses existing evaluations and search metadata, retaining CALM's search rules and GRPO learning procedure. It requires no additional candidate execution.

The evaluation connects the reward construction to search outcomes through three levels: complete searches and selected-program tests assess the heuristics found by search; common-context generation and population-reset searches examine trained checkpoints under fixed or reinitialized conditions; and reward contrasts on shared records identify changes in the training signal. The main searches cover the traveling salesman problem (TSP), capacitated vehicle routing problem (CVRP), online bin packing (OBP) and orienteering (OP), with seven seeds per task and condition at the same response budget. Comparisons include a frozen generator, CALM's native reward and alternative reward constructions. Among these is a local Tail-weighted control that adapts the exponential weighting of TTT-Discover \citep{yuksekgonul2026tttdiscover}. Tests of programs selected by search assess their performance on held-out instances and additional sizes. Common-context generation on TSP and CVRP compares checkpoints under fixed prompts and parents. TSP searches with reinitialized populations and further training disabled examine whether a trained checkpoint can carry benefits into a fresh search. Reward recomputation on shared records examines how the construction changes training feedback.

SVC achieves the lowest mean final costs on TSP and CVRP and the lowest mean final gap on OBP among the six local conditions. Relative to Tail-weighted, it also obtains a higher mean final prize on OP, although the descriptive 95\% intervals for all four paired endpoint differences include zero. On common TSP and CVRP contexts, SVC generates parent-improving responses more often on average than Native CALM, but less often than Tail-weighted. Comparisons over the full search trajectories and tests of selected programs also vary by task and instance size.

\Needspace{7\baselineskip}
Our contributions are:

\begin{enumerate}
\item \textbf{A response-level reward construction for evolutionary AHD.} SVC combines actual admission with a short-term parent-exposure proxy to translate search retention and selection information into online generator training. Its central design choice is to give feedback about candidates' potential participation in subsequent search using information available when they are evaluated.
\item \textbf{An empirical characterization of search and generation under this reward.} The joint evaluation reveals different mean orderings for search endpoints and immediate parent-improvement yields, together with task- and size-dependent outcomes for the selected programs. These observations clarify what the measured benefits establish about SVC and why endpoint performance alone is insufficient to characterize changes in the generator.
\item \textbf{An analysis of the learning signal produced by SVC.} Recomputing rewards on the same recorded candidate groups shows that SVC produces fewer groups with zero task advantages than admission-only and no-distance-credit variants. This identifies how the reward components affect the availability of nonzero task feedback, while distinguishing that effect from evidence that the credited candidates improve later search.
\end{enumerate}

The following sections position SVC relative to prior work, define its reward construction and approximation, describe the evaluation protocols, and present the results and their implications.

\section{Related work}
\label{sec:related}
\subsection{Evolutionary search and generator adaptation}

LLM-based automatic heuristic design uses evaluated programs to guide the generation of further candidates. FunSearch maintains a program database, EoH evolves natural-language descriptions together with executable heuristics, and ReEvo adds reflections on the relative performance of candidate programs \citep{romeraparedes2024funsearch,liu2024eoh,ye2024reevo}. These methods adapt the information supplied to a fixed generator during search. LLaMEA iteratively generates, mutates and selects algorithms using runtime feedback to design metaheuristics \citep{vanstein2024llamea}, whereas MCTS-AHD organizes all generated heuristics in a tree so that temporarily underperforming heuristics can still be developed \citep{zheng2025mctsahd}. AlphaEvolve improves programs through direct code changes guided by feedback from one or more evaluators, targeting scientific and computational problems \citep{novikov2025alphaevolve}. These approaches illustrate how the organization of search can change which programs are proposed and retained.

Search procedures also determine how evaluated candidates inform later proposals. OPRO supplies previously evaluated solutions and their values as context for subsequent proposals \citep{yang2024opro}. LLM4AD provides separate interfaces for search methods, design tasks and language models \citep{liu2024llm4ad}, while the comparative study of Zhang et al. examines the contribution of evolutionary search across AHD methods and generators \citep{zhang2024evolutionarysearch}. These works motivate distinguishing the generator from the search procedure in which it operates.

EvoTune adds parameter adaptation by training generation policies on preference pairs accumulated through evolutionary exploration \citep{surina2025evotune}. ThetaEvolve similarly combines program search with test-time reinforcement learning \citep{wang2025thetaevolve}, whereas task-specific fine-tuning for algorithm design \citep{liu2025finetuning} and Evolution Fine-Tuning \citep{lee2026evolutionfinetuning} train generators on search-derived data before they are used in new searches. The former combines diversity-aware rank-based data selection with preference optimization; the latter converts evolutionary trajectories across tasks into supervision. CALM couples the evolution of heuristic programs with online generator training using GRPO \citep{huang2026calm,shao2024deepseekmath}. SVC builds on CALM and studies how the evaluations produced by this coupled process are converted into response-level training rewards.

\subsection{Execution feedback and reward construction}

Execution provides feedback about generated programs, but the learning signal depends on how that feedback is used. CodeRL trains a code generator with reinforcement learning and a learned critic that predicts functional correctness using unit-test feedback \citep{le2022coderl}. RLTF uses unit-test feedback at multiple levels of granularity in online code-model training \citep{liu2023rltf}. Other learning procedures use rewards to filter generated examples, as in reinforced self-training \citep{gulcehre2023rest}, or optimize preferences directly, as in DPO \citep{rafailov2023dpo}. These choices differ both in the feedback retained and in how it affects the generator.

For group-relative learning, reward construction and normalization jointly determine the advantages used for training. GRPO is a variant of proximal policy optimization that estimates its baseline from group scores rather than a learned critic \citep{schulman2017ppo,shao2024deepseekmath}. GDPO normalizes each reward component within its response group, combines the resulting advantages, and then normalizes across the training batch \citep{liu2026gdpo}. Zero-variance groups are addressed through different interventions: DAPO oversamples and filters out prompts whose sampled answers are all correct or all incorrect \citep{yu2025dapo}, while RL-ZVP constructs correctness-directed, entropy-scaled token advantages for such groups \citep{le2026rlzvp}. These methods change how group-relative training signals are computed or which groups are used.

In evolutionary AHD, CALM already uses generation context: its reward compares candidate performance with that of the best prompting parent \citep{huang2026calm}. Parent-relative performance and participation in later search describe different candidate properties. SVC uses actual admission and the inherited parent-selection rules to construct feedback intended to reflect a candidate's potential use in subsequent search. The search procedure and GRPO learner are inherited from CALM; the intervention concerns the response reward.

\subsection{Learning signals for search outcomes}

Selection outcomes have also been used to adapt generation strategies in evolutionary computation \citep{qin2009sade}. More broadly, reinforcement learning can assist evolutionary algorithms through solution generation, operator selection and parameter adaptation \citep{song2024rlea}. Recent learning-based discovery systems use search information in several parts of the learning and search process. HeurAgenix evolves heuristics and selects among them, with an option to fine-tune a lightweight selector \citep{yang2025heuragenix}, and AHD Agent learns a multi-turn policy that can generate heuristics or invoke tools to obtain evidence from the solving environment \citep{lv2026ahdagent}. PACEvolve++ trains a strategic advisor while a separate model implements candidate solutions. Its subset leave-one-out component measures a candidate's marginal contribution to the maximum score in candidate subsets and is combined with a group-relative component after each is standardized \citep{yan2026pacevolvepp}. The mixing coefficient changes with training stage. TTT-Discover combines test-time learning with reuse of search states and an entropic objective that emphasizes high rewards \citep{yuksekgonul2026tttdiscover}. Its reuse mechanism incorporates the best observed one-step child reward when selecting states to expand. None of these systems, however, reports a reward for the candidate generator that is gated by the actual admission outcome of the inherited search and weighted by that search's own parent-selection rules; this combination is the object studied here.

Our Tail-weighted control adapts TTT-Discover's exponential weighting to an AHD-specific contextual utility within CALM, followed by the inherited GRPO normalization. Within evolutionary AHD with an online-trained LLM generator, SVC constructs a response-level reward by combining actual candidate admission with a short-term parent-exposure proxy derived from the search's parent-selection rules. This construction is intended to make training feedback reflect candidates' potential use in subsequent search. The implemented proxy describes selection-based exposure; its relationship to useful search outcomes requires empirical evaluation through complete searches, selected-program tests and generator evaluations.

\section{Method}
\label{sec:method}
\subsection{Problem setting and search--learning loop}

We study automatic heuristic design in which an LLM generates heuristic programs for a target optimization task. Each program implements a decision rule or a component of a prescribed solver, which produces solutions when run on problem instances. An evaluator executes the candidate under a fixed task protocol and aggregates its performance over the search evaluation instances into a fitness value $f$, with higher values indicating better performance. The search seeks high-performing programs within a fixed candidate budget. Online post-training adapts the generator using the experience collected during this search. Our question is how to turn that experience into training feedback that helps the generator support subsequent search. The fitness of a selected program and the generator's behavior under a given prompt are distinct evaluation targets.

We build on CALM's evolutionary search and online training framework \citep{huang2026calm}. Let $S_t$ denote the external search state used to construct the prompt for response group $t$, after any required state maintenance. This state includes retained programs and relevant search history; $\theta_t$ denotes the generator parameters used to sample the group. Search control selects an operator and its parent programs, denoted jointly by $\kappa_t$, and constructs a prompt $x_t=\mathcal C(S_t,\kappa_t)$. The prompt combines the task description, operator instructions and selected parent information, and requests a design description with executable code. The generator samples $G\geq2$ responses $y_{t,i}\sim\pi_{\theta_t}(\cdot\mid x_t)$. Evaluation records whether each response satisfies the prescribed checks and, when successful, the extracted program's fitness. Reward computation can additionally access search metadata that is not included in the prompt.

The retained set $\mathcal A_t$ and the main parent pool $\mathcal H_t$ play different roles. The latter contains up to the $K=10$ highest-fitness retained programs and supplies parents for simplification, injection and replacement, as well as the primary parent for crossover. These parents are sampled with probabilities proportional to inverse fitness rank. For each crossover proposal, the second-parent route is chosen with equal probability: fitness-rank sampling from $\mathcal H_t$, or idea-distance-rank sampling from the full retained set $\mathcal A_t$, subject to CALM's rejection rules. During candidate processing, CALM admits an eligible evaluated program when its fitness differs from every currently retained fitness by at least $10^{-7}$. Candidates are processed in sequence, so admission can depend on earlier candidates in the same group. The retained set can grow beyond $K$; stagnation-dependent reductions retain the best and initialization programs. Admission does not require improvement over the prompt parents. A retained program outside the main parent pool can still participate through the distance-based crossover route.

Candidate evaluations inform both retained-set updates and generator training. Let $E_t$ collect the response evaluation records, $M_t$ the search context before candidate processing, and $\mathbf a_t$ the response admission indicators recovered from the actual retained-set update. A reward construction $\phi$ maps this information to response-level rewards:
\begin{equation}
\mathbf r_t=\phi(E_t,M_t,\mathbf a_t).
\label{eq:reward-construction}
\end{equation}
The learner then normalizes these rewards within the response group:
\begin{equation}
\widehat{\mathbf A}_t=\mathcal N_G(\mathbf r_t).
\label{eq:search-learning-loop}
\end{equation}

Here $\mathcal N_G$ subtracts the group mean and divides by the sample standard deviation plus a small constant. The resulting group-relative advantages weight the sampled responses in the inherited GRPO learning procedure, which also includes policy--reference KL regularization \citep{shao2024deepseekmath}. Fitness $f_{t,i}$ evaluates a program on the task, reward $r_{t,i}$ supplies training feedback for its response, and advantage $\widehat A_{t,i}$ expresses that reward relative to the other responses in the group. SVC changes $\phi$ while retaining the search rules and learning procedure. Parameter updates change the distribution of subsequent candidates, which can in turn change later retained sets and prompts. The following subsection defines SVC's reward construction, designed to reflect candidates' potential use in subsequent search through actual admission and a short-term parent-exposure proxy.

\subsection{Search-value credit}
\label{sec:svc}

Search-value credit (SVC) constructs response-level training rewards from actual admission and a short-term parent-exposure proxy. Its aim is to make feedback to the generator reflect candidates' potential use in subsequent search. In CALM, retained programs can be selected as parents and included in subsequent prompts. Admission identifies the programs added to the retained set, and the parent-selection rules govern their opportunities for reuse. SVC uses admission to determine reward eligibility and the exposure proxy to weight eligible responses according to these selection rules. The required admission flags and exposure weights are available after the current response group has been evaluated, without waiting for descendants to be generated or evaluated.

\begin{figure}[!htbp]
\centering
\includegraphics[width=\linewidth]{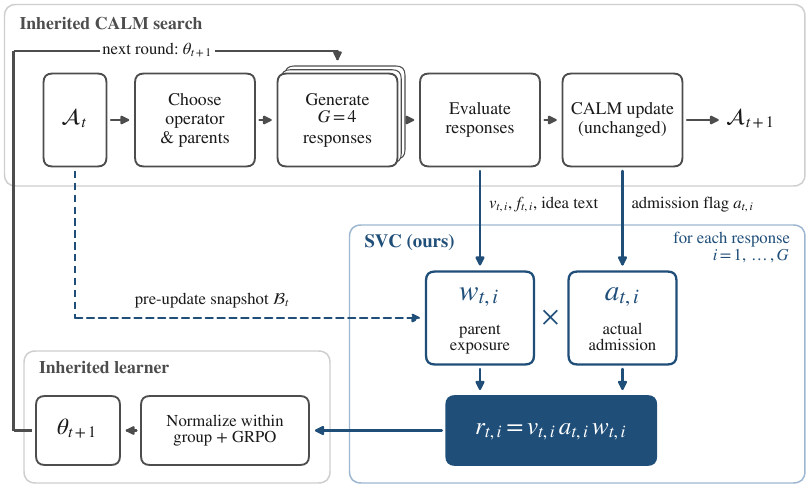}
\caption{SVC reward construction within CALM for response group $t$. Gray components are inherited from CALM and unchanged: the search step from the retained set $\mathcal A_t$ to $\mathcal A_{t+1}$ and the GRPO learner. Blue components are added by SVC and computed for each response $i$. Evaluation supplies the validity indicator $v_{t,i}$, the fitness $f_{t,i}$ and the idea text; the admission flag $a_{t,i}$ is the actual outcome of the CALM update. The parent-exposure weight $w_{t,i}$ is computed after evaluation by inserting the response alone into the ordered pre-update snapshot $\mathcal B_t$ and applying the parent-selection rules that CALM uses; it is a proxy for reuse opportunities and is zero for responses that are not eligible for this hypothetical insertion. The dashed line marks this reference snapshot, not a data flow. The learner normalizes $r_{t,i}=v_{t,i}a_{t,i}w_{t,i}$ within the group, so a response with zero reward receives a negative advantage whenever another response in the group has a positive reward.}
\label{fig:svc-overview}
\end{figure}

Figure~\ref{fig:svc-overview} summarizes the reward construction and its place in the search--learning loop.

\FloatBarrier

\subsubsection{Connecting rewards to the actual search update}

Reward eligibility depends on both evaluation validity and the actual retained-set update. Let $v_{t,i}\in\{0,1\}$ indicate that response $y_{t,i}$ is marked valid, underwent program evaluation, and has a finite recorded fitness $f_{t,i}$. The response-level admission flag $a_{t,i}\in\{0,1\}$ is recovered from that update as described below. Given a nonnegative exposure weight $w_{t,i}$, the task reward is
\begin{equation}
r_{t,i}=v_{t,i}a_{t,i}w_{t,i}. \label{eq:svc-m1}
\end{equation}

CALM processes evaluated candidates in sequence. A program that passes the preceding checks is added if its fitness differs from the fitness of every currently retained program by at least $\tau=10^{-7}$. Earlier candidates in the group can therefore affect the admission of later candidates. To recover $a_{t,i}$, the implementation matches the hash of each complete response against the response hashes associated with newly retained program objects. Identical valid responses can share an admission flag even when only one program is added. An invalid duplicate still receives zero reward because it does not satisfy $v_{t,i}=1$. The flag therefore indicates a match to a newly retained program; it does not identify a separate insertion for each response. Invalid responses and valid responses with $a_{t,i}=0$ receive zero task reward. The main method adds no separate failure penalty.

This admission gate links reward eligibility to the observed search update. The exposure weight then distinguishes eligible responses by their opportunities for reuse, evaluated against a common snapshot of the retained programs.

\subsubsection{Computing exposure against a shared snapshot}

The exposure calculation uses an ordered snapshot $\mathcal B_t=(b_1,\ldots,b_n)$ of retained programs with finite fitness. This snapshot is taken from $M_t$ immediately before the group's candidates are processed. The following indicator specifies whether a candidate is eligible for hypothetical insertion into the snapshot:

\begin{equation}
e_{t,i}=
\begin{cases}
1,&v_{t,i}=1\ \text{and}\ |f_{t,i}-f(b)|\geq\tau\quad\forall b\in\mathcal B_t,\\
0,&\text{otherwise}.
\end{cases} \label{eq:svc-m2}
\end{equation}

For an empty snapshot, the condition on fitness differences holds. If $e_{t,i}=0$, the exposure weight is zero. Otherwise, appending candidate $i$ alone gives the ordered hypothetical collection $\mathcal B_t^{(i)}$. This calculation leaves the actual search state unchanged. Because every candidate is evaluated against the same snapshot, the exposure calculation omits competition among new candidates in the group. The admission flag in \eqref{eq:svc-m1} still reflects the actual sequential update. For example, consider two distinct valid responses with the same fitness. If that fitness does not duplicate any value in the snapshot under the stated tolerance, both satisfy $e_{t,i}=1$, although only the first program is admitted.

For the following calculation, fix a candidate with $e_{t,i}=1$ and suppress the indices $(t,i)$. The hypothetical collection contains $N=n+1$ programs indexed by $k\in\{1,\ldots,N\}$; the appended candidate has index $k_\star=N$. These collection indices are distinct from the response index $i$. Let $\rho_k$ denote program $k$'s rank in descending fitness order, with ties resolved by collection order. The main parent pool contains the $m=\min(K,N)$ highest-ranked programs, indexed by $\mathcal H=\{k:\rho_k\leq m\}$. 

\Needspace{7\baselineskip}
Their sampling weights are
\begin{equation}
p_k=
\begin{cases}
\displaystyle\frac{\rho_k^{-1}}{\sum_{\ell=1}^{m}\ell^{-1}},&k\in\mathcal H,\\
0,&k\notin\mathcal H.
\end{cases} \label{eq:svc-m3}
\end{equation}

The experiments use $K=10$. CALM samples one parent from this distribution for each single-parent operator and uses the same distribution to sample the primary parent for crossover. The corresponding primary-parent weight forms the first exposure component. Crossover introduces an additional parent, so the exposure calculation also includes its two second-parent selection routes.

\subsubsection{Accounting for the crossover routes}

For each crossover proposal, CALM selects one of two routes with probability $1/2$ each. The fitness route samples the second parent from the main parent pool; the distance route samples from the full retained collection according to idea-distance ranks. For the latter route, let $T_k$ be the set of lowercase word tokens in program $k$'s idea description. The directed distance from primary parent $j$ to program $k$ is
\begin{equation}
 d(j,k)=
 \begin{cases}
 |T_k\setminus T_j|/|T_k|,&|T_k|>0,\\
 0,&|T_k|=0.
 \end{cases}
 \label{eq:idea-distance}
\end{equation}
This distance measures differences between idea texts. Let $\delta_{k\mid j}\in\{1,\ldots,N\}$ rank program $k$ by decreasing distance from $j$. Ties are resolved by the original order of $\mathcal B_t^{(i)}$, assigning a distinct ordinal rank to each program. The normalized inverse-rank weight is
\begin{equation}
 q_{k\mid j}=\frac{\delta_{k\mid j}^{-1}}{\sum_{\ell=1}^{N}\delta_{\ell\mid j}^{-1}}.
 \label{eq:svc-m4}
\end{equation}
The denominator includes all $N$ programs, including $j$ itself and any other program at zero distance from $j$.

Both second-parent routes contribute only when crossover is proposed. Their weights therefore include the crossover proposal probability, computed from the hypothetical collection size. Let $(u_s,u_i,u_r,u_c)$ be the configured nonnegative weights for simplification, injection, replacement and crossover. Crossover is disabled when fewer than two programs are available, giving $\widetilde u_c(N)=u_c\mathbf1\{N\geq2\}$. If the hypothetical collection is smaller than the main-pool limit and injection is enabled, its weight is raised to the largest current operator weight:
\begin{equation}
 \widetilde u_i(N)=
 \begin{cases}
 \max\{u_s,u_i,u_r,\widetilde u_c(N)\},&N<K\ \text{and}\ u_i>0,\\
 u_i,&\text{otherwise}.
 \end{cases}
 \label{eq:injection-weight}
\end{equation}
After these adjustments, the total weight is $Z(N)=u_s+\widetilde u_i(N)+u_r+\widetilde u_c(N)$. Normalizing the crossover weight yields the proposal probability
\begin{equation}
 c(N)=
 \begin{cases}
 \widetilde u_c(N)/Z(N),&Z(N)>0,\\
 0,&Z(N)=0.
 \end{cases}
 \label{eq:svc-m5}
\end{equation}
The experiments use weights $(1,1,2,4)$ and $K=10$, giving $c(N)=0$ for $N<2$, $4/11$ for $2\leq N<K$, and $1/2$ for $N\geq K$. When the total weight is zero, the reward observer sets $c(N)=0$; the actual search procedure requires a positive total weight. Adjusting $c(N)$ changes the crossover proposal probability, while the two routes within crossover each retain branch probability $1/2$.

The distance-based route contributes through primary parents other than the candidate itself and only at positive idea distance. Averaging over primary-parent proposals, before multiplication by the crossover route probability, gives
\begin{equation}
 D_{k_\star}=\sum_{\substack{j\in\mathcal H,\ j\ne k_\star\\ d(j,k_\star)>0}}
 p_j q_{k_\star\mid j}.
 \label{eq:distance-component}
\end{equation}
With this notation, the complete exposure definition separates the primary-parent, fitness-based second-parent and distance-based second-parent components:
\begin{equation}
 w_{t,i}=
 \begin{cases}
 p_{k_\star}+\dfrac{c(N)}2p_{k_\star}+\dfrac{c(N)}2D_{k_\star},&e_{t,i}=1,\\[3pt]
 0,&e_{t,i}=0.
 \end{cases}
 \label{eq:svc-m6}
\end{equation}

All quantities in the first branch are computed using candidate $i$'s hypothetical collection. The first term is its primary-parent weight, and the second is its fitness-based second-parent weight. The third averages its distance-based second-parent weight over possible primary parents. The primary term has coefficient one because each of the four operators uses a primary parent at the proposal stage. A candidate outside the main parent pool can still receive positive exposure through the distance route and, if $a_{t,i}=1$, positive reward under \eqref{eq:svc-m1}.

Equation \eqref{eq:svc-m6} combines proposal weights while holding the reference state fixed. The fitness-based second-parent term includes proposals in which the same program occupies both parent positions. The distance term excludes self-pairs and zero-distance pairs after computing $q$, without renormalizing the remaining weights. The calculation does not account for the conditional distribution induced by crossover retries and prompt rejection. It also omits fallback to creation, subsequent retained-set reductions, and the duration of each program's retention. For distance ties, exposure computation uses stable sorting, whereas the native selector does not specify a stable sorting algorithm. These approximations limit $w_{t,i}$ to a proxy for reuse opportunities; it is neither the exact probability of inclusion in the next successfully constructed prompt nor an estimate of descendant improvement. SVC uses this proxy to construct immediate training feedback. Its effects on generation and search are evaluated empirically.

\subsubsection{Turning rewards into learning signals}

Equation~\eqref{eq:svc-m1} combines validity, actual admission and the exposure defined in \eqref{eq:svc-m6} to assign a task reward to each response. The inherited GRPO procedure \citep{shao2024deepseekmath} then normalizes the rewards within the group of responses sampled for the same prompt. For $G\geq2$, the group mean is $\bar r_t=G^{-1}\sum_{i=1}^G r_{t,i}$ and the sample standard deviation is
\begin{equation}
 s_t=\sqrt{\frac1{G-1}\sum_{i=1}^G(r_{t,i}-\bar r_t)^2}.
 \label{eq:sample-sd}
\end{equation}
Centering each reward and scaling by this group statistic gives the response-level task advantage:
\begin{equation}
 \widehat A_{t,i}=\frac{r_{t,i}-\bar r_t}{s_t+10^{-4}}.
 \label{eq:svc-m7}
\end{equation}

The denominator uses the sample standard deviation, with the stabilizer added after taking the square root. The resulting response-level advantages weight the sampled responses in the existing policy update, which also includes KL regularization relative to the reference policy. A response with zero reward has negative advantage if another response in its group has positive reward. A valid response with an admission flag and positive reward also has negative advantage if its reward is below the group mean. Constant rewards produce zero task advantages, but do not imply zero parameter updates. Thus, eligibility for a positive task reward does not determine the sign of the group-relative advantage.

SVC uses CALM's search and online training framework \citep{huang2026calm}, retaining its admission rule, parent-selection distributions, idea distance, and GRPO learning procedure. The reward construction introduced here combines admission flags recovered from the actual update with parent-exposure weights computed by inserting each candidate separately into the pre-update snapshot. The resulting feedback reflects both the observed search update and the opportunities for reuse represented by the inherited selection rules.

Implementation details, the complete pseudocode, and a constructed numerical example are provided in \suppsection{app:svc-implementation}.

\section{Experimental setup}
\label{sec:design}

\subsection{Tasks and comparisons}

Qwen2.5-7B-Instruct \citep{yang2024qwen25} serves as the program generator for four optimization tasks: the traveling salesman problem (TSP), capacitated vehicle routing (CVRP), online bin packing (OBP), and orienteering (OP). Generated programs select the next node in TSP, score feasible bins in OBP, and provide heuristic information to fixed ant colony optimization solvers \citep{dorigo1996antsystem} in CVRP and OP.

SVC and five control conditions share the same base model and CALM search rules. \textbf{Frozen} disables model training, and \textbf{Native CALM} uses CALM's original reward. \textbf{Validity-only} assigns rewards based on whether a generated program executes successfully and returns a finite score. \textbf{Factorized Validity--Quality} combines a signed validity signal with RMS-scaled, parent-relative score differences. \textbf{Tail-weighted} combines a tie-aware rank utility for validity, archive-score novelty and within-group response repetition with an RMS-scaled positive frontier gap, then applies adaptive exponential weighting. This local control adapts TTT-Discover's concentration rule. All conditions with training enabled use the same training settings. The control rewards are defined in~\suppsection{app:control-rewards}.

For each task, all six conditions use the same seven random seeds, giving 168 searches in total. Each search evaluates 500 groups of four responses, for a budget of 2,000 responses including invalid outputs. Model training uses low-rank adaptation (LoRA) \citep{hu2022lora} and GRPO on NVIDIA RTX 4090 and RTX 3090 GPUs. Dataset details, solver settings, and execution configurations are provided in \suppsection{app:protocol}.

\subsection{Search outcomes and selected-program tests}

The three subsections that follow correspond to the three evaluation levels summarized in Section~\ref{sec:introduction}: search outcomes and selected programs, checkpoint behavior under fixed or reinitialized conditions, and reward contrasts on shared records.

Search performance is measured by the final best fitness and the mean best-so-far fitness over the 500 response groups. Each search contributes one program for testing: the valid program with the highest training fitness among those admitted during search. This program is evaluated on separate test instances at the training problem sizes and larger sizes, without reselection based on test performance. All CVRP selected-program tests use a common float64 evaluation protocol; dataset sizes and evaluation details are given in \suppsection{app:protocol}.

\subsection{Generation and population-reset evaluations}

Common-context generation evaluates final model checkpoints using the same prompts and parent programs, with further training disabled. The prompt sets contain 128 prompts for TSP and 64 for CVRP, taken from the corresponding Frozen search trajectories with seed 42. Each checkpoint generates four responses per prompt. The TSP comparison includes SVC and all four controls with training enabled; the CVRP comparison includes SVC, Native, Validity-only, and Tail. Each of these conditions contributes seven checkpoints, and each task has one Frozen checkpoint as a reference. The reported metrics are the validity rate, the proportion of all responses that yield valid programs outperforming the best parent in the prompt, and the proportion of prompts for which at least one of the four responses yields such an improvement.

Population-reset evaluation initializes a new TSP search for each of three SVC checkpoints, with further training disabled. Each search uses the same 2,000-response budget and is compared with the Frozen search using the corresponding seed. This evaluates whether the trained model improves search performance after the population is reset. Sampling settings and execution details are documented in \suppsection{app:protocol}.

\subsection{Ablation studies and reward analysis}

Ablation studies on TSP examine SVC's reward construction under the same seven seeds and search budget. Three variants are compared with SVC: one adds failure penalties; another replaces exposure weighting with a binary reward for valid programs admitted during search; and the third removes the distance-based contribution to the exposure reward while retaining distance-based parent selection. Evaluation covers complete searches, selected-program tests, and common-context generation. A supplementary variant replaces positive exposure values with rank-based weights and adds failure penalties; it changes two factors and is evaluated through searches and selected-program tests.

Shared-record replay applies eight reward mappings to the same recorded response groups from the 28 SVC searches across all four tasks, without generating candidates or updating models. The analysis measures associations between the resulting group-relative advantages and two aspects of subsequent search use: whether a candidate is later selected as a parent and whether it is an ancestor of a later program that improves the best fitness found so far. A separate OBP analysis examines search plateaus and groups in which all task advantages are zero. Replay protocols, label construction, and supplementary diagnostics are specified in \suppsection{app:protocol} and \suppsection{app:credit-analysis}.

\subsection{Metrics and statistical reporting}

Task performance is measured by tour cost for TSP, route cost for CVRP, the percentage gap to the volume lower bound for OBP, and collected prize for OP. Lower costs and gaps and higher collected prize indicate better performance. Results are first summarized within each run or checkpoint and then averaged with equal weight across seeds. Comparisons between conditions with matching seeds use paired differences and descriptive 95\% $t$ intervals. Differences in generation rates and OBP gaps are reported in percentage points. The unit of replication is the search run or model checkpoint, rather than an individual response or test instance. Each Frozen generation reference comprises one checkpoint, and the three population-reset searches are reported individually. Metric formulas, OBP aggregation rules, and supplementary statistical procedures are provided in \suppsection{app:protocol}.

\section{Results and analysis}
\label{sec:results}
\subsection{Complete-search performance}
\label{sec:search}

\begin{figure}[!htbp]
\centering
\includegraphics[width=\linewidth]{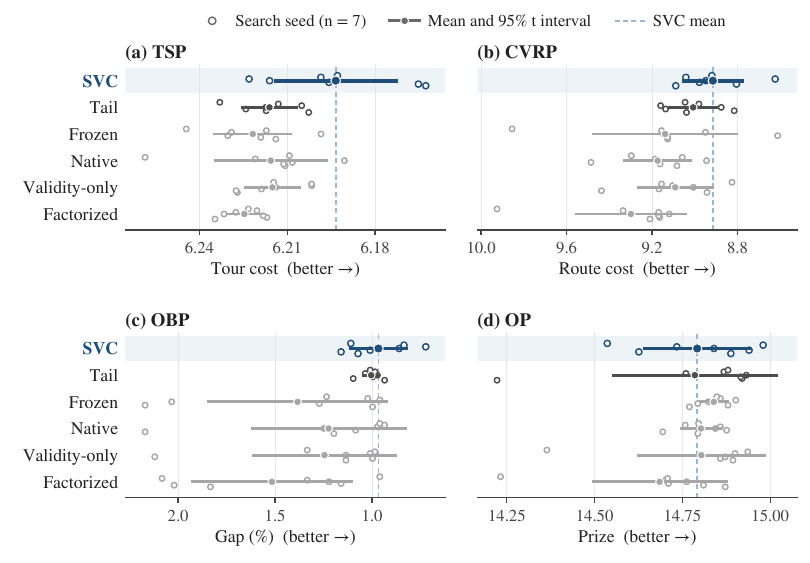}
\caption{Search endpoints after 2,000 responses for SVC and the five local controls. SVC (dashed line) has the best mean on TSP, CVRP and OBP. Open circles are the seven search seeds, vertically offset only to separate points; filled markers and bars give the mean and descriptive 95\% Student-$t$ interval over seeds. Each panel keeps the task's own metric and is oriented so that right is better (cost and gap axes reversed). Paired comparisons are reported in Table~\ref{tab:main-results}.}
\label{fig:search-endpoints}
\end{figure}

At a budget of 2,000 responses per search, SVC achieves the lowest mean final cost on TSP and CVRP and the lowest mean final gap on OBP among the six conditions using Qwen2.5-7B-Instruct (Table~\ref{tab:main-results} and Figure~\ref{fig:search-endpoints}). The mean final values are 6.19345, 8.91526, and 0.96734\%, respectively. These endpoints measure the quality of the best programs found during search. On OP, SVC obtains a mean final prize of 14.79200, exceeding Tail and Factorized but falling below Frozen, Native, and Validity-only.

Relative to Tail, SVC reduces mean final tour cost by 0.02263 on TSP, route cost by 0.09246 on CVRP, and gap by 0.03739 percentage points on OBP, while increasing mean final prize by 0.00657 on OP. The TSP reduction occurs in six of seven paired runs. The corresponding counts are four for CVRP and three each for OBP and OP, so the improvements in the latter two task means do not extend to most seed pairs. Table~\ref{tab:main-results} reports the paired differences and their descriptive 95\% intervals; all four intervals include zero.

The reductions relative to Factorized are larger: 0.03122 in mean final TSP cost, 0.38375 in CVRP cost, and 0.54876 percentage points in OBP gap. Their descriptive 95\% paired intervals are [0.00983, 0.05261], [0.09794, 0.66956], and [0.11560, 0.98192], respectively. SVC has lower costs in all seven TSP and CVRP pairs and a lower gap in six OBP pairs. The intervals for cost reductions relative to Frozen on TSP and Native on CVRP also lie above zero. The remaining local-control endpoint intervals, including all five OP comparisons, contain zero (Table~\ref{tab:svc-paired-final}).

\begin{table}[!htbp]
\centering
\caption{Search endpoints after 2,000 responses (mean $\pm$ sample SD over seven search seeds). SVC has the lowest mean cost on TSP and CVRP and the lowest mean gap on OBP among the six local conditions. Relative to Tail, it has lower mean costs and gap and a higher mean OP prize. Bold marks the best mean among the six local conditions; it does not denote significance. Paired benefit is control minus SVC for cost and gap and SVC minus control for prize; OBP differences are in percentage points.}
\label{tab:main-results}
\small
\begin{tabular*}{\linewidth}{@{\extracolsep{\fill}}lcccc@{}}
\toprule
 & TSP & CVRP & OBP & OP \\
Condition & tour cost $\downarrow$ & route cost $\downarrow$ & gap (\%) $\downarrow$ & prize $\uparrow$ \\
\midrule
\textbf{SVC} & \textbf{6.193}\,{\scriptsize$\pm$0.023} & \textbf{8.92}\,{\scriptsize$\pm$0.16} & \textbf{0.97}\,{\scriptsize$\pm$0.16} & 14.79\,{\scriptsize$\pm$0.17} \\
\addlinespace[1pt]
Tail & 6.216\,{\scriptsize$\pm$0.011} & 9.01\,{\scriptsize$\pm$0.13} & 1.00\,{\scriptsize$\pm$0.05} & 14.79\,{\scriptsize$\pm$0.26} \\
Frozen & 6.222\,{\scriptsize$\pm$0.015} & 9.14\,{\scriptsize$\pm$0.37} & 1.38\,{\scriptsize$\pm$0.51} & \textbf{14.84}\,{\scriptsize$\pm$0.05} \\
Native & 6.216\,{\scriptsize$\pm$0.021} & 9.17\,{\scriptsize$\pm$0.18} & 1.22\,{\scriptsize$\pm$0.43} & 14.80\,{\scriptsize$\pm$0.06} \\
Validity-only & 6.215\,{\scriptsize$\pm$0.011} & 9.09\,{\scriptsize$\pm$0.20} & 1.25\,{\scriptsize$\pm$0.40} & 14.80\,{\scriptsize$\pm$0.20} \\
Factorized & 6.225\,{\scriptsize$\pm$0.007} & 9.30\,{\scriptsize$\pm$0.28} & 1.52\,{\scriptsize$\pm$0.45} & 14.69\,{\scriptsize$\pm$0.21} \\
\midrule
API system$^{\dagger}$ & 6.234\,{\scriptsize$\pm$0.007} & 9.08\,{\scriptsize$\pm$0.08} & 1.21\,{\scriptsize$\pm$0.47} & 14.78\,{\scriptsize$\pm$0.07} \\
\midrule
\multicolumn{5}{@{}l}{\textit{Paired endpoint benefit, SVC vs.\ Tail (positive favors SVC)}} \\
\addlinespace[1pt]
Mean benefit & $+0.0226$ & $+0.092$ & $+0.037$ & $+0.007$ \\
95\% interval & {\footnotesize$[-0.0003,\,0.0455]$} & {\footnotesize$[-0.107,\,0.292]$} & {\footnotesize$[-0.111,\,0.186]$} & {\footnotesize$[-0.303,\,0.316]$} \\
W/T/L & 6/0/1 & 4/0/3 & 3/0/4 & 3/0/4 \\
\bottomrule
\end{tabular*}

\par\medskip
{\footnotesize\raggedright $^{\dagger}$GPT-4o-mini: seven search seeds for each of the four tasks; values are mean $\pm$ sample SD. Parameters remain fixed during search.\par}
\end{table}

\subsubsection{Performance across the search budget}

The mean best-so-far metric across the 500 response groups gives a different view of the search. Relative to Tail, SVC has lower budget-averaged costs on TSP (6.21416 versus 6.22708) and CVRP (9.19736 versus 9.31018), but a higher budget-averaged gap on OBP and a lower budget-averaged prize on OP. The ordering on the latter two tasks therefore differs from the endpoint results. Using the raw task metric and computing SVC minus Tail, the paired mean differences are \ensuremath{-}0.01292 in TSP cost, \ensuremath{-}0.11282 in CVRP cost, +0.07828 percentage points in OBP gap, and \ensuremath{-}0.03513 in OP prize; all four paired intervals include zero. These budget-averaged metrics provide complementary evidence of the search dynamics.

Comparisons at 400, 1,000, and 2,000 responses further show how the relative outcomes change with budget. The mean OBP gap is higher for SVC than for Tail at the first two positions, before becoming lower at the endpoint. On OP, SVC has a higher mean prize at 400 responses, a lower mean at 1,000, and a slightly higher mean at 2,000. SVC has lower mean costs than Tail on TSP and CVRP at all three positions.

Complete paired endpoint, budget-average, intermediate-budget, and second-half-improvement results are provided in \suppsection{app:search-results}; budget summaries and endpoint ranks appear in \suppsection{app:budget-ranks}.

\subsubsection{API system comparison}

The GPT-4o-mini API experiments also cover all four tasks with seven search seeds per task, for 28 searches in total. Task means average the seven runs, and paired comparisons use the matching search seeds. Against this API system, whose parameters remain fixed during search, SVC reduces mean final tour cost by 0.04028 on TSP and route cost by 0.16754 on CVRP. The descriptive 95\% paired intervals for these reductions are [0.01946, 0.06111] and [0.02300, 0.31207]. SVC also has a lower mean final OBP gap and a higher mean final OP prize, although both paired intervals include zero. For budget-averaged performance, the TSP cost-reduction interval remains above zero; the CVRP and OBP intervals contain zero. On OP, the API system obtains the higher budget-averaged mean prize, with a paired interval that also contains zero (Tables~\ref{tab:svc-paired-final} and~\ref{tab:svc-paired-budget-mean}).

\subsection{Performance of the selected programs}
\label{sec:program-tests}

To examine whether the search improvements extend to new instances, we evaluate the programs selected by training fitness on held-out test sets. The programs obtained with SVC achieve lower mean costs than those obtained with Tail on every TSP and CVRP test set, including instances larger than those used during search. They also achieve the lowest mean gap among the six conditions on four of the six OBP test sets. All comparisons use seven matched search seeds. The intervals below are descriptive 95\% intervals for paired cost or gap reductions (control minus SVC), or prize increases (SVC minus control).

\begin{figure}[!htbp]
\centering
\includegraphics[width=\linewidth]{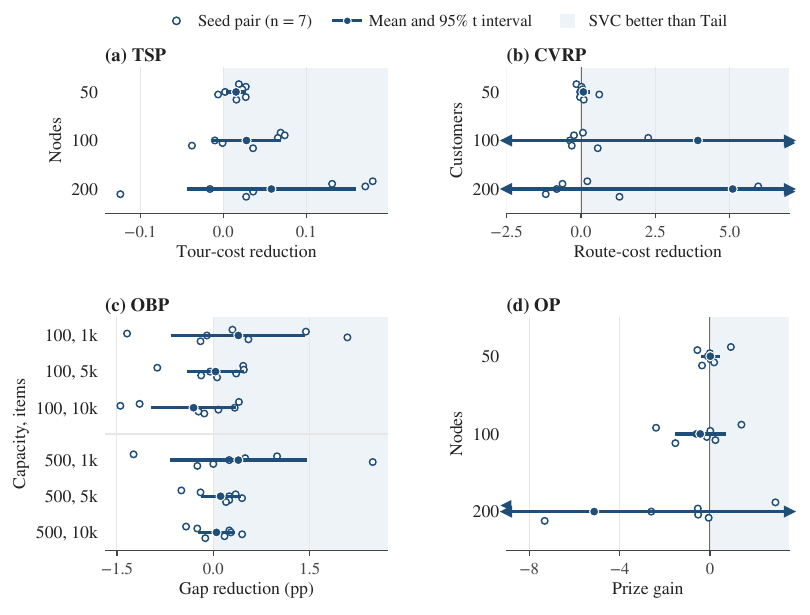}
\caption{Test performance of the programs selected by SVC and Tail at each test scale. SVC programs have lower mean cost than Tail programs at every TSP and CVRP size and a lower mean gap on five of six OBP settings. Each open circle is one of the seven matched search seeds; the filled marker and bar give the mean paired benefit and its descriptive 95\% Student-$t$ interval. Positive values (shaded) denote cost and gap reductions for TSP, CVRP and OBP, and prize increases for OP. OBP labels give bin capacity and number of items. Values beyond the CVRP and OP axes are marked by arrows: CVRP 100, one seed $+25.5$, interval $[-4.9, +12.8]$; CVRP 200, one seed $+30.9$, interval $[-5.6, +15.8]$; OP 200, one seed $-27.7$, interval $[-14.8, +4.5]$. Complete means and comparisons with all controls are given in \suppsection{app:program-tests}.}
\label{fig:selected-programs}
\end{figure}

Figure~\ref{fig:selected-programs} summarizes the SVC--Tail comparisons; complete six-condition means and paired contrasts are given in \suppsection{app:program-tests}.

On TSP, the programs obtained with SVC, which are selected through searches on 50-node instances, achieve the lowest mean tour cost among the six conditions at all three test sizes. Their mean costs are 6.24393, 8.70728, and 12.20907 at 50, 100, and 200 nodes, respectively. Relative to Tail, these correspond to reductions of 0.01537 [0.00352, 0.02722], 0.02797 [\ensuremath{-}0.01320, 0.06915], and 0.05797 [\ensuremath{-}0.04407, 0.16000], with lower costs in six, four, and five of the seven paired comparisons. The lowest mean cost is therefore achieved both at the search size and on instances with twice and four times as many nodes.

A similar pattern holds for CVRP under the common float64 evaluation protocol. SVC achieves the lowest mean route cost at 50 and 100 customers, while at 200 customers it has a lower mean cost than all four alternative reward conditions. Across the three sizes, its mean costs are 9.27922, 15.82433, and 28.29313, compared with 9.36023, 19.75206, and 33.39339 for Tail. The corresponding reductions are 0.08101 [\ensuremath{-}0.14855, 0.31057], 3.92773 [\ensuremath{-}4.91800, 12.77346], and 5.10026 [\ensuremath{-}5.64264, 15.84316]. Four of seven pairs favor SVC at each size; the larger mean reductions at 100 and 200 customers partly reflect one Tail program with substantially higher route costs than the other Tail programs. Frozen is the only condition with a lower mean cost than SVC at 200 customers, where it achieves 28.19590.

On OBP, SVC achieves the lowest mean gap on four of the six test sets, including the capacity-500 set containing 10,000-item sequences, which are longer than those used during search. Its mean gaps are lower than those of Tail on five sets and lower than those of Validity-only and Factorized on all six. For 1,000 items and capacity 100, the reduction relative to Validity-only is 0.46862 [0.09344, 0.84380] percentage points. The two sets on which SVC does not achieve the lowest mean gap reveal differences across both sequence length and capacity. For 10,000 items and capacity 100, Tail achieves a lower mean gap: 0.98240\% for SVC versus 0.67535\% for Tail, giving a paired benefit of \ensuremath{-}0.30704 [\ensuremath{-}0.97142, 0.35733] percentage points. For 1,000 items and capacity 500, Frozen and Native have lower mean gaps than SVC. All gaps are computed by aggregating bin counts and volume lower bounds within each test file.

On OP, SVC's small mean advantage over Tail is confined to the 50-node test set, which matches the size used during search. Mean prize is 15.03414 for SVC and 15.02151 for Tail, giving a paired benefit of 0.01263 [\ensuremath{-}0.42274, 0.44800]. At 100 and 200 nodes, SVC obtains mean prizes of 30.32694 and 50.40584, compared with 30.76118 and 55.53223 for Tail, with paired benefits of \ensuremath{-}0.43424 [\ensuremath{-}1.57354, 0.70506] and \ensuremath{-}5.12639 [\ensuremath{-}14.78146, 4.52868]. At both sizes, its mean prize is below that of every control, and only one of the seven pairs favors SVC over Tail at 200 nodes. The improvement observed at the search size therefore does not extend to larger OP instances.

Taken together, the search and test results support the use of SVC to discover heuristics that perform well beyond the instances on which they were selected. This is most evident on TSP and CVRP, where the lower mean search costs are accompanied by lower mean test costs than Tail at every evaluated size; the best mean OBP gaps on four test sets provide further support. These findings concern the performance of the selected programs.

\subsection{Generation under common contexts}
\label{sec:generation}

The preceding results show that SVC obtains heuristics with lower mean search and test costs than Tail on TSP and CVRP. We next examine the trained generators through their outputs under common contexts and their ability to support a newly initialized search. For the generation evaluation, the final checkpoints receive the same prompts and parent programs: 128 TSP contexts and 64 CVRP contexts drawn from a Frozen search, with four responses sampled per prompt. Parent-improvement yield measures the fraction of all responses that produce a valid program whose fitness exceeds that of the best parent in the prompt; best-of-4 measures the fraction of prompts for which at least one response meets this criterion. Each metric is computed for each checkpoint and then averaged over seven seeds. Frozen provides a single-checkpoint reference.

\begin{figure}[!t]
\centering
\includegraphics[width=\linewidth]{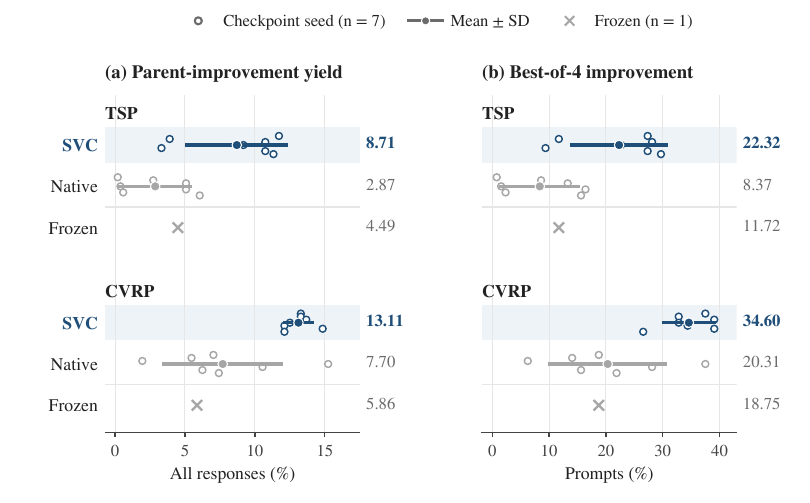}
\caption{Common-context generation on 128 TSP and 64 CVRP prompts, with four responses per prompt. SVC raises the mean parent-improvement yield over Native from 2.87\% to 8.71\% on TSP and from 7.70\% to 13.11\% on CVRP, and the mean best-of-4 rate from 8.37\% to 22.32\% and from 20.31\% to 34.60\%. Parent-improvement yield is the percentage of all responses, including invalid responses, that produce a valid program with fitness above the best parent in the prompt; best-of-4 is the percentage of prompts with at least one such response. Open circles show all seven checkpoint seeds; filled circles and bars show the mean and sample SD. Crosses denote one Frozen reference checkpoint per task; numbers give means or reference values. These gains occur despite lower validity (TSP: 81.11\% versus 96.46\%; CVRP: 79.91\% versus 95.09\%); SVC remains below Tail on both improvement measures (metrics for all evaluated conditions in \suppsection{app:common-generation}).}
\label{fig:common-generation}
\end{figure}

SVC generated parent improvements more frequently than Native on both tasks (Figure~\ref{fig:common-generation}).
 On TSP, parent-improvement yield increased from 2.87\% to 8.71\%, a difference of 5.83 percentage points with a descriptive 95\% paired interval of [1.37, 10.29]. Best-of-4 increased from 8.37\% to 22.32\%. On CVRP, the corresponding yield increased from 7.70\% to 13.11\%, a difference of 5.41 points [0.98, 9.85], and best-of-4 increased from 20.31\% to 34.60\%. These gains occurred alongside lower executable validity: 81.11\% versus 96.46\% on TSP and 79.91\% versus 95.09\% on CVRP. Because improvement yield includes invalid responses in its denominator, the higher yields reflect more successful improvements per sampling budget despite the lower validity.

The comparison with Tail gives a different ordering. SVC had lower mean parent-improvement yield on TSP, 8.71\% versus 12.61\%, and on CVRP, 13.11\% versus 13.84\%. The paired differences were \ensuremath{-}3.91 points [\ensuremath{-}8.75, 0.94] and \ensuremath{-}0.73 points [\ensuremath{-}3.28, 1.82], respectively. Tail also had higher mean best-of-4 and frontier-improvement yield, where the frontier is the best retained fitness before generation in the search that supplied the prompt. Frontier-improvement yields were 1.28\% versus 2.43\% on TSP and 3.24\% versus 3.57\% on CVRP; their paired intervals included zero. The remaining comparisons were task dependent: SVC exceeded Factorized in TSP parent-improvement yield and best-of-4, whereas its comparison with Validity-only was mixed on TSP and favored SVC on both measures on CVRP.

SVC also generated fewer valid responses that matched the best parent's fitness. With a tolerance of $10^{-7}$, these responses accounted for 7.47\% of valid SVC outputs on TSP, compared with 85.30\% for Native and 20.34\% for Tail. The corresponding CVRP proportions were 19.02\%, 61.24\%, and 30.17\%. SVC had a lower proportion than Native at every seed on both tasks, and than Tail at all seven TSP seeds and six of the seven CVRP seeds. The mean proportions were also lower when all responses formed the denominator. Together with the higher parent-improvement yield relative to Native, these results show that SVC checkpoints produce a different distribution of candidate scores under fixed evaluation contexts, with fewer parent-score ties and more successful improvements per sampling budget. Score ties describe fitness outcomes rather than code identity or algorithmic diversity; the validity and Tail comparisons above complete this picture of generation behavior.

Generation metrics for all evaluated conditions are provided in \suppsection{app:common-generation}, and TSP reward-variant comparisons in \suppsection{app:variant-generation}. These evaluations compare checkpoint outputs under identical prompts and parents with further training disabled. Factorized is included on TSP but was not evaluated on CVRP.

\subsection{Search after resetting the population}
\label{sec:reset}

\begin{figure}[!htbp]
\centering
\includegraphics[width=\linewidth]{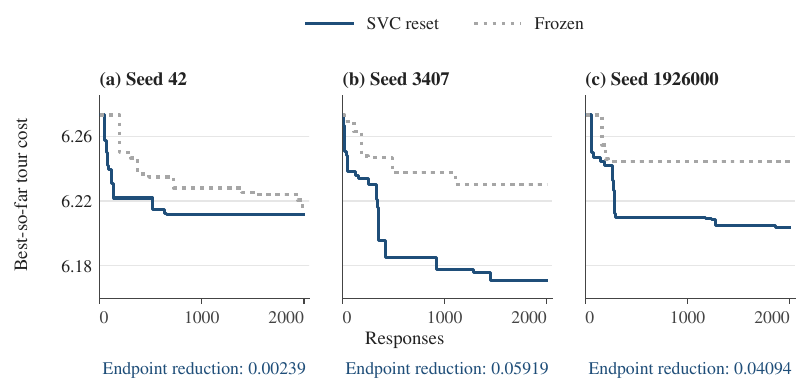}
\caption{Population-reset search on TSP for three seeds. SVC reset reuses the final SVC checkpoint after reinitializing the population and search history, with further training disabled; Frozen uses the base generator. Curves show best-so-far tour cost (lower is better) at all 500 groups of four responses. SVC reset ends below Frozen in all three pairs, with endpoint reductions of 0.002, 0.059 and 0.041 (Frozen minus SVC reset), and also has lower budget-averaged cost in every pair. The descriptive 95\% paired $t$ interval for the mean reduction includes zero. The loaded adapter did not apply to the first response group.}
\label{fig:reset-trajectories}
\end{figure}

Figure~\ref{fig:reset-trajectories} shows the three reset trajectories. Population-reset search further tests whether the trained checkpoints can support a new search. The final SVC checkpoints from three TSP seeds were loaded after reinitializing the population and search history, with further training disabled and a budget of 500 groups of four responses. Relative to the corresponding Frozen searches, final tour costs decreased by 0.00239, 0.05919, and 0.04094 for seeds 42, 3407, and 1926000, respectively; mean best-so-far cost also decreased in all three pairs. The mean final-cost reduction was 0.03417, with a descriptive 95\% paired interval of [\ensuremath{-}0.03787, 0.10621].

These results support the value of reusing SVC checkpoints: with the loaded parameters held fixed, all three searches achieved gains without retaining the earlier population or continuing training. They complement the preceding search and selected-program results by showing that the trained generator can also help a newly initialized TSP search discover good heuristics. The three-pair sample and the interval spanning zero leave uncertainty about the average benefit. The evaluation protocol documents the shared first response group before the loaded adapter took effect.

Paired reset values are given in Table~\ref{tab:reset-values}; execution details are specified in \suppsection{app:protocol-generation}.

\subsection{Ablating the reward construction}

We conduct ablation studies on TSP to evaluate the roles of SVC's reward components. Across seven seeds, SVC achieved lower mean final and budget-averaged costs than all four ablation variants (Table~\ref{tab:svc-ablation-main} and Table~\ref{tab:svc-ablation-supplementary}). It achieved lower final costs in six of seven paired runs against each of the added-penalty and Admission-only variants. The mean per-seed relative reductions were 0.307\% and 0.387\%, with descriptive 95\% paired intervals of [0.010\%, 0.604\%] and [0.017\%, 0.757\%], respectively. These results support weighting admitted responses by exposure without adding a separate failure penalty.

The three main variants respectively add CALM's penalties for invalid responses, replace exposure weights with binary admission rewards, and remove the distance component from the reward while retaining distance-based parent selection. Relative to No-distance, SVC's mean per-seed relative reduction in final cost was 0.189\%; the paired interval for this comparison and those for budget-averaged cost differences all included zero. The supplementary Rank+penalty variant replaces exposure weights with within-group rank weights and adds failure penalties (Table~\ref{tab:svc-ablation-supplementary}). These ablations are exploratory because TSP results informed the choice of SVC's reward formulation. Under common generation contexts, SVC also achieved higher mean parent- and frontier-improvement yields than Admission-only (Table~\ref{tab:internal-generation}). Selected-program tests for all four variants are reported in \suppsection{app:ablation-program-tests}.

\begin{table}[!htbp]
\centering
\caption{Reward ablations on TSP. SVC has the lowest mean endpoint and budget-averaged cost and a lower endpoint cost than the added-penalty and Admission-only variants in six of seven paired runs. The ablations are exploratory because TSP results informed the choice of SVC's reward formulation. Tour costs are mean $\pm$ sample SD over seven search seeds. Endpoints use 2,000 responses; budget means average the best-so-far cost over response groups 1--500. Bold marks the lowest mean among the displayed conditions; it does not denote significance.}
\label{tab:svc-ablation-main}
\small
\begin{tabular*}{\linewidth}{@{\extracolsep{\fill}}llccc@{}}
\toprule
 & & Endpoint & Budget & SVC \\
Condition & Change from SVC & cost $\downarrow$ & mean $\downarrow$ & W/T/L$^{\dagger}$ \\
\midrule
\textbf{SVC} & Reference & \textbf{6.193}\,{\scriptsize$\pm$0.023} & \textbf{6.214}\,{\scriptsize$\pm$0.021} & --- \\
\addlinespace[1pt]
Added-penalty & adds failure penalty & 6.213\,{\scriptsize$\pm$0.016} & 6.230\,{\scriptsize$\pm$0.017} & 6/0/1 \\
Admission-only & admission weight 1 & 6.218\,{\scriptsize$\pm$0.017} & 6.231\,{\scriptsize$\pm$0.010} & 6/0/1 \\
No-distance & no distance term & 6.205\,{\scriptsize$\pm$0.010} & 6.223\,{\scriptsize$\pm$0.009} & 5/0/2 \\
\bottomrule
\end{tabular*}

\par\medskip
{\footnotesize\raggedright $^{\dagger}$SVC endpoint wins/ties/losses against each variant, paired by seed. No-distance removes only reward distance credit and retains distance-based parent selection. Its endpoint relative-reduction interval and all budget-mean difference intervals include zero. The two-factor Rank+penalty variant is reported in \suppsection{app:ablation-details}.\par}
\end{table}

\begin{figure}[!htbp]
\centering
\includegraphics[width=\linewidth]{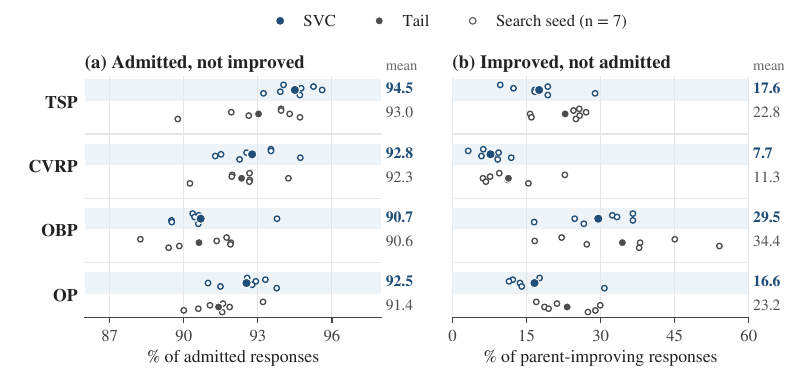}
\caption{Admission and parent improvement on SVC and Tail search trajectories. (a) Among admitted eligible responses, the proportion that did not improve on the best parent in the prompt. (b) Among improving eligible responses, the proportion not admitted. Open circles show seven seed-level proportions; filled circles and numbers show their equal-weight means. Eligibility requires a valid, evaluated, non-initialization response with finite fitness and a finite parent comparator. Improvement is strict; admission is a response-hash match to a newly retained program and does not count unique insertions. The two panels use different conditional denominators. Both conditions use the same admission rule but visit different states and generate different candidates; these descriptive differences motivate distinguishing admission from immediate improvement and do not isolate a causal advantage of the reward mapping. Results for all six conditions are given in Figure~\ref{fig:app-admission-all}.}
\label{fig:admission-divergence}
\end{figure}

\begin{samepage}
\subsection{Admission and parent improvement}
\label{sec:admission}
\looseness=-1

The admission records show how often retention and parent improvement diverge under the inherited search rules. For valid responses from non-initialization prompts with a finite parent comparator, improvement requires fitness strictly greater than that of the best parent in the prompt. Across the four tasks, the mean proportion of admitted SVC responses that did not improve on this parent ranged from 90.68\% to 94.50\%; the mean proportion of improving responses that were not admitted ranged from 7.71\% to 29.53\% (Figure~\ref{fig:admission-divergence}; each task averages seven seed-level proportions). A candidate can be admitted with a fitness value absent from the retained set, or rejected for repeating a retained value despite exceeding its parent comparator. This supports distinguishing admission from parent improvement in reward construction: SVC uses actual admission as a gate and exposure weights to reflect opportunities for subsequent use as a parent.

\end{samepage}

\subsection{Credit on shared records}
\label{sec:credit}

To examine the effect of reward construction on the training signal, we apply different reward mappings to the same recorded responses. On each of the 28 SVC trajectories, the SVC mapping yielded a lower proportion of zero-advantage groups than Admission-only and No-distance. The mean proportions on TSP were 7.43\%, 46.57\%, and 50.17\%, respectively; those across the other three tasks were 13.44\%, 36.93\%, and 51.55\%. Exposure weights can distinguish valid admitted responses within a group, and the distance component can assign credit to admitted candidates outside the main parent pool. These mechanisms extend the availability of nonzero feedback relative to the two internal variants; Tables~\ref{tab:credit-replay-tsp}--\ref{tab:credit-replay-non_tsp} report the other mappings.

SVC also assigned a larger share of positive advantage to responses whose programs were later used as parents. Across 21 non-TSP trajectories, these responses accounted for a mean of 37.45\% of the total positive advantage, compared with 27.53\% under Native and 33.81\% under Tail. Across all four tasks, SVC also yielded higher mean Spearman correlations between group-relative advantage and subsequent parent-use count than Native and Tail (Tables~\ref{tab:credit-replay-tsp}--\ref{tab:credit-replay-op_aco}). Correlations were computed among valid responses within each trajectory and averaged over the seven seeds per task; the task means ranged from 0.2976 to 0.3449 for SVC, compared with 0.1954--0.2435 for Native and 0.2385--0.2821 for Tail.

Across the 21 non-TSP trajectories, SVC's group-relative advantages yielded a mean area under the receiver operating characteristic curve (AUC) of 0.7673 for distinguishing ancestor-labeled from unlabeled responses among valid admitted responses. This label marks \mbox{programs} that were ancestors of later valid, evaluated candidates whose finite fitness strictly exceeded the finite best retained fitness before generation. The paired intervals for SVC's AUC differences from Native and Tail included zero. No-distance had a higher mean AUC of 0.8087 but more zero-advantage groups, illustrating a tradeoff between nonzero feedback coverage and discrimination of the ancestor label. Replay characterizes feedback and its association with subsequent use on fixed records, while complete search experiments evaluate search outcomes.

\subsection{OBP plateau diagnosis}

SVC retained nonzero task advantages during OBP plateaus as well. A plateau is a maximal interval of at least 100 consecutive response groups during which the best-so-far fitness reconstructed from valid generated candidates remains unchanged. Six of seven SVC trajectories contained plateaus. Averaged over these six trajectories, 11.80\% of plateau groups had zero task advantages, so most groups still provided nonzero feedback despite the unchanged best fitness. Applying Tail rewards to the same records yielded a proportion of 4.04\%, showing that the reward mapping further affects feedback availability within plateaus.

All-condition admission results are given in \suppsection{app:admission-events}, replay definitions and tables in \suppsection{app:replay}, and plateau diagnostics in \suppsection{app:plateau}. Figure~\ref{fig:app-shared-response} illustrates the reward mappings on the same four recorded responses.

\section{Discussion}
\label{sec:discussion}

SVC aims to make training feedback to the generator reflect candidates' potential use in subsequent search. Actual admission determines reward eligibility, while eligible responses are weighted by a short-term parent-exposure proxy derived from the existing parent-selection rules. The resulting response-level rewards, which combine retention and selection information, are available for online post-training after the current candidates have been evaluated, without waiting for their descendants to be evaluated. SVC achieves the lowest mean final costs on TSP and CVRP and the lowest mean final gap on OBP among the six local conditions. The programs obtained with SVC also have lower mean costs than those obtained with Tail at every evaluated TSP and CVRP size. Under common contexts on these two tasks, SVC achieves higher mean parent-improvement yields than Native, while gains over Frozen in three TSP population-reset searches further support the reuse of trained checkpoints. The credit analysis also shows that exposure weighting distinguishes admitted responses and provides nonzero task advantages in a larger proportion of response groups than the Admission-only and No-distance mappings.

In evolutionary search, retained programs are outputs of the search that can also serve as references for subsequent generation. SVC draws on this dual role by incorporating the reuse opportunities implied by parent-selection rules into training feedback, so that rewards account for candidates' roles in the search process. The multi-level evaluation shows that search outcomes and immediate generation behavior can have different relative orderings: SVC has lower mean final costs than Tail on TSP and CVRP, while Tail has higher mean parent-improvement yields under common contexts. The admission-gating mechanism aligns reward eligibility with search retention, while exposure weighting represents opportunities for subsequent parent use under the existing selection rules. This perspective motivates further study of exposure-based rewards that are adapted to different parent-selection strategies, together with evaluations over longer search budgets. The current results indicate the potential of this direction.

\section{Conclusion}
\label{sec:conclusion}

We introduced Search-value credit, which combines actual admission with a short-term parent-exposure proxy to construct rewards for online generator training in evolutionary automatic heuristic design. Among six local conditions, SVC achieves the lowest mean final costs on TSP and CVRP and the lowest mean final gap on OBP. Its selected programs also have lower mean costs than Tail's at every tested TSP and CVRP size, and its checkpoints produce more parent improvements per response than Native under common contexts on these tasks. The credit analysis shows that exposure weighting provides nonzero task advantages in more response groups than Admission-only and No-distance. These results support using search retention and selection information to construct generator training rewards. The benefits depend on the task and comparator: most paired search and test intervals include zero, and the small mean improvement over Tail at the OP search size does not extend to larger OP test instances.

\FloatBarrier
\section*{Data and code availability}
This study builds on the publicly available CALM implementation \citep{calmReleaseSoftware}. The code and experimental data supporting the results reported in this article are available to the editor and reviewers as a compressed archive upon request during peer review. The authors intend to make these materials publicly available upon acceptance of the manuscript.

\section*{Declaration of competing interests}
The authors declare that they have no known competing financial interests or personal relationships that could have appeared to influence the work reported in this paper.

\section*{Funding}
This research did not receive any specific grant from funding agencies in the public, commercial, or not-for-profit sectors.

\section*{CRediT authorship contribution statement}
\textbf{\FirstAuthorName:} Conceptualization, Methodology, Software, Investigation, Data curation, Visualization, Writing -- original draft.
\textbf{\SecondAuthorName:} Methodology, Formal analysis, Visualization, Writing -- original draft.
\textbf{\CorrespondingAuthorName:} Writing -- review \& editing, Supervision.

\section*{Declaration of generative AI and AI-assisted technologies in the manuscript preparation process}
During the preparation of this manuscript, the authors used Codex (OpenAI) with the GPT-6-Astra model to assist with drafting and revising text, organizing the manuscript, interpreting recorded results, refining the layout and presentation of figures and tables in the main manuscript and supplementary appendices, and preparing a Chinese translation for author review. Codex also assisted with reviewing and organizing code, preparing scripts for statistical checks and Matplotlib visualization, and checking references, formatting, and cross-references. Empirical figures were plotted from recorded numerical results; the method overview and the explicitly labeled constructed example were drawn programmatically. The authors take full responsibility for the content of this publication.

\FloatBarrier
\clearpage
\appendix
\counterwithin{figure}{section}
\counterwithin{table}{section}
\counterwithin{equation}{section}
\renewcommand{\thefigure}{\Alph{section}.\arabic{figure}}
\renewcommand{\thetable}{\Alph{section}.\arabic{table}}
\renewcommand{\theequation}{\Alph{section}.\arabic{equation}}
\section{Implementation details and worked example}
\label{app:svc-implementation}

This appendix describes how response records are matched and how SVC rewards and group-relative advantages are computed. The notation follows the method section: $v_{t,i}$ denotes validity, $a_{t,i}$ the response-level admission flag, $e_{t,i}$ eligibility for hypothetical insertion, and $w_{t,i}$ exposure. The ordered snapshot $\mathcal B_t$ contains the fitness values and idea descriptions of retained programs with finite fitness, saved immediately before the response group is processed. The experiments use four responses per group, a main parent pool of size $K=10$, weights $(1,1,2,4)$ for simplification, injection, replacement, and crossover, respectively, and a fitness tolerance of $\tau=10^{-7}$.

\subsection{Response records and actual admission}
\label{app:method-records}

Before evaluation, the observer saves the identities of the retained program objects and the ordered snapshot. CALM processes evaluation results in response order. It adds a program if the program passes its checks and no retained program has fitness $f'$ satisfying $|f-f'|<\tau$, where $f$ is the candidate's fitness. The comparison includes programs added earlier in the same group; admission requires neither improvement over the parent comparator nor membership in the main parent pool. The observer identifies newly retained programs by comparing object identities before and after the update, then matches their complete responses to sampled responses using SHA-1 hashes of the UTF-8 text, without removing whitespace or normalizing code. Thus, distinct responses with the same fitness can compete for admission when that fitness is absent from the retained set, whereas identical responses can share an admission flag arising from a single program insertion.

The observer also uses complete-response hashes to record whether a matching evaluation exists and to look up fitness. It processes the evaluation ledger in order, storing the mean performance of each non-error result and overwriting any earlier fitness for the same hash; execution failures returned as strings do not overwrite fitness. Identical responses can therefore share these fields, but each response's status is decoded from its own native CALM reward using the configured failure codes. SVC treats a response as valid only if this status is valid, a matching evaluation record exists, and the recorded fitness is finite. An invalid response with an admission flag passes the consistency check only if its hash occurs more than once in the group; its reward remains zero. Otherwise, the reward routine raises an error. Every valid response with an admission flag must also have finite, nonnegative exposure. The number of flagged responses can therefore differ from the number of distinct program objects inserted.

\Needspace{9\baselineskip}
\subsection{Computational procedure}

The pseudocode presents one SVC search and training iteration for a response group, from prompt construction and response generation through the actual search update, response matching, exposure computation, and policy update. Exposure is computed inside the response loop using the same saved snapshot for each candidate; admission flags reflect the actual sequential update. Repeating this procedure gives the online search and training loop.

\begin{figure}[!htbp]
\hrule\medskip
\noindent\textbf{SVC: one search and training iteration}\par\medskip
\hrule\medskip
\begin{Verbatim}[fontsize=\fontsize{9}{10.5}\selectfont,numbers=left,numbersep=5pt,xleftmargin=12pt]
Input: policy theta, search state M, group size G >= 2, K >= 1
       operator weights u0=(us,ui,ur,uc), tau=1e-7
Construct a prompt using CALM's operator/parent selection
Sample responses y[1:G] from the current policy
B <- ordered finite-fitness entries before processing y
I <- identities of all currently retained program objects
Apply CALM's pre-evaluation checks; evaluate extracted programs
Process results in response order; update M by CALM admission
U <- response hashes of retained objects whose identities are not in I
E,F <- evaluated hashes and scores from the ordered ledger
       (last non-error score per hash)
h[i] <- SHA1_UTF8(y[i]), for every response i
status[i] <- decode each native CALM reward using failure codes
for i = 1,...,G:
    v[i] <- status[i] is valid AND h[i] in E AND finite F.get(h[i])
    a[i] <- h[i] in U; repeated <- count(h[i] in h) > 1
    e[i] <- false; w[i] <- 0
    if v[i]:
        e[i] <- every b in B satisfies abs(F[h[i]] - f(b)) >= tau
    if e[i]:
        L <- B with (F[h[i]], idea of y[i]) appended
        N <- length(L); star <- N; m <- min(K,N)
        rho <- descending fitness ranks; break ties in L order
        H <- indices with rho <= m; p <- zero vector of size N
        p[k] <- (1/rho[k])/sum(l=1,...,m; 1/l), for k in H
        u <- u0; if N < 2: u[crossover] <- 0
        if N < K AND ui > 0: u[injection] <- max(u)
        c <- u[crossover]/sum(u) if sum(u) > 0, else 0
        T[k] <- lowercase word-token set of idea[k], for all k
        D <- 0
        for j in H:
            d[k] <- |T[k] minus T[j]|/|T[k]|, for all k
                    (use 0 when T[k] is empty)
            delta <- descending distance ranks; ties in L order
            q[k] <- (1/delta[k])/sum(l=1,...,N; 1/delta[l])
            if j != star AND d[star] > 0: D <- D+p[j]*q[star]
        w[i] <- p[star] + (c/2)*p[star] + (c/2)*D
    if a[i] AND NOT v[i] AND NOT repeated: raise record error
    if a[i] AND v[i]: require finite, nonnegative w[i]
    r[i] <- w[i] if a[i] AND v[i], else 0
reward32 <- float32(r); reward_mean <- mean(reward32)
s <- sqrt(sum((reward32[i]-reward_mean)^2 for i=1:G)/(G-1))
advantage[i] <- (reward32[i]-reward_mean) / (s+1e-4)
Update theta using inherited GRPO with KL regularization
Output: updated search state M and generator policy theta
\end{Verbatim}

\medskip\hrule
\end{figure}

Here, \texttt{F.get(h[i])} returns the recorded fitness if available; a missing value fails the finiteness check. The word-token extractor applies the regular expression \texttt{\string\b\string\w+\string\b} to lowercased idea descriptions and stores the distinct tokens as a set. Tied distances receive successive integer ranks in collection order, rather than a shared average rank. All $N$ entries contribute to the denominator used to normalize inverse ranks, including the primary parent and entries at zero distance. Excluding self-pairs and zero-distance pairs from the final sum does not renormalize this distribution. For the experimental operator weights, the probability of proposing crossover is zero for $N<2$, $4/11$ for $2\leq N<K$, and $1/2$ for $N\geq K$. The observer returns zero crossover probability when the total operator weight is zero; the native selector requires a positive total weight.

The observer saves the snapshot before each reward-function call and identifies newly retained objects after the search update, once per call. It then groups contiguous responses by prompt hash and computes rewards within each group. Each call in the main experiments contains one prompt group. If a call contains multiple prompt groups, those groups share the snapshot and admission matching from that call. Reproducing a record therefore requires knowing which call it belongs to and its position within the group, as well as its prompt identity.

\subsection{Numerical example}

\begin{figure}[!htbp]
\centering
\includegraphics[width=\linewidth]{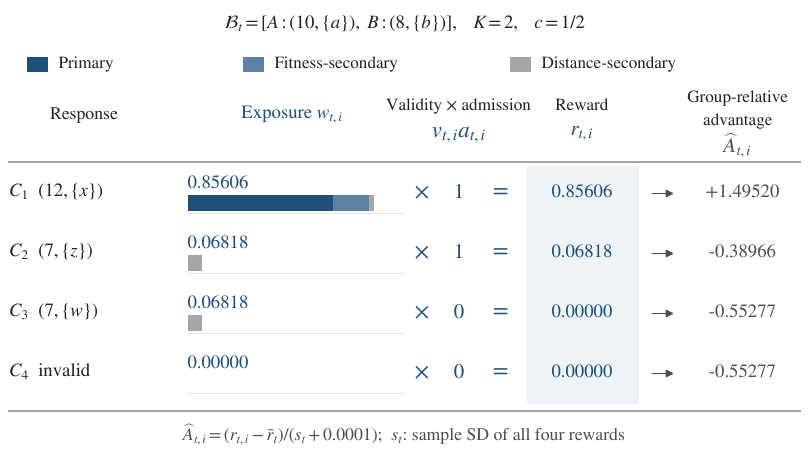}
\caption[Constructed SVC computation]{Constructed SVC example with $K=2$ (experimental $K=10$). The table reports the validity--admission product $v_{t,i}a_{t,i}$, reward, and group-relative advantage; the bars decompose exposure before admission gating. All hypothetical insertions use the same ordered snapshot $\mathcal B_t$. Responses $C_2$ and $C_3$ both qualify for hypothetical insertion ($e_{t,2}=e_{t,3}=1$), but only $C_2$ is admitted. Values are rounded to five decimal places for display. Advantages are calculated from unrounded rewards with sample standard deviation and stabilizer $10^{-4}$. These values are constructed, not experimental measurements.}
\label{fig:svc-worked-example}
\end{figure}

Figure~\ref{fig:svc-worked-example} uses a constructed four-response group to illustrate the computation. Its main parent pool is deliberately reduced to $K=2$; the experiments use $K=10$. The ordered snapshot is $A=(10,\{a\})$, $B=(8,\{b\})$, where each pair gives fitness and idea tokens. The distinct responses are $C_1=(12,\{x\})$, $C_2=(7,\{z\})$, $C_3=(7,\{w\})$, and invalid $C_4$. The first three responses pass the evaluation checks. Sequential processing admits $C_1$ and $C_2$, while $C_3$ repeats the fitness already added by $C_2$. Thus $\mathbf v_t=(1,1,1,0)$, $\mathbf a_t=(1,1,0,0)$, and $\mathbf e_t=(1,1,1,0)$.

The hypothetical collection for each eligible candidate has $N=3$ and $c=1/2$. From either original program, the other original program and the appended candidate both have distance one. Collection order places the candidate second among these tied entries, giving it a normalized inverse-rank weight of $(1/2)/(1+1/2+1/3)=3/11$ on the distance route. For $C_1$, the main parent pool is $(C_1,A)$ with probabilities $(2/3,1/3)$. Its three exposure components are therefore

\begin{equation}
w_{t,1}=\frac23+\frac14\frac23+\frac14\frac13\frac3{11}
    =\frac{113}{132}.
\label{eq:example-exposure-c1}
\end{equation}

When hypothetically inserted one at a time, candidates $C_2$ and $C_3$ lie outside the main parent pool and receive exposure only through the distance route:

\begin{equation}
w_{t,2}=w_{t,3}=\frac14\left(\frac23+\frac13\right)\frac3{11}
       =\frac3{44}.
\label{eq:example-exposure-distance}
\end{equation}

Actual admission gates these weights. Because $C_1$ and $C_2$ are admitted while $C_3$ and $C_4$ are not, the reward vector is $\mathbf r_t=(113/132,\,3/44,\,0,\,0)$. Its group mean is $\bar r_t\simeq0.2310606061$, and its sample standard deviation is $s_t\simeq0.4179044974$. Applying equation~(\ref{eq:svc-m7}) of the main article to these unrounded values, with stabilizer $10^{-4}$, gives
\begin{equation}
\widehat{\mathbf A}_t\simeq(1.495199,\,-0.389658,\,-0.552771,\,-0.552771).
\label{eq:example-advantages}
\end{equation}

The example separates three effects. The distance route gives $C_2$ positive reward even though it lies outside the main parent pool; its reward is below the group mean, so its advantage is negative. Candidate $C_3$ has positive hypothetical exposure but zero reward because it is not admitted. The invalid response also receives zero reward and negative advantage in this group. These are consequences of the constructed inputs, rather than measurements of search performance.

\subsection{Numerical precision and interpretation}

The observer returns raw rewards to the learner. The learner converts them to float32 and recomputes the group mean, sample standard deviation with denominator $G-1$, and advantage. The logging routine independently performs the same normalization in float64. In both paths, $10^{-4}$ is added after taking the square root. Logged advantages can therefore differ in their last digits from the training values; reproducing a float64 record is not a bitwise reproduction of a training update. This example uses unrounded float64 rewards for calculation. Constant rewards give zero task advantages, which does not remove the separate KL term or guarantee a zero optimizer update.

Exposure combines proposal weights in a fixed hypothetical collection. It omits competition with other newly generated candidates, crossover retries, rejection of repeated prompts, fallback to creation, subsequent reductions of the retained set, and how long a program remains in that set. The second-parent term based on fitness retains the proposal probability assigned to self-pairs; the distance term excludes self-pairs and zero-distance pairs after normalization. In addition, the proxy breaks distance ties using the snapshot order, with the candidate appended last. The native selector first sorts retained programs by fitness and does not explicitly request stable sorting when ranking distances. Exposure therefore serves as a short-term proxy for reuse opportunities. It does not give the exact probability of inclusion in the next successfully constructed prompt or quantify improvements in descendant programs or long-term return.

\FloatBarrier
\section{Experimental configuration and evaluation protocols}
\label{app:protocol}

\subsection{Tasks, data, and program interfaces}

Table~\ref{tab:protocol-data} lists the search and test instances, program roles, and solver budgets. CVRP and OP use fixed ant colony optimization (ACO) solvers with the same budgets during search and selected-program testing. CVRP sizes count customers, excluding the depot. The 22 data files contain no byte-identical instances shared between search and testing within a task; this does not establish statistical independence of data generation.

\begingroup
\small
\setlength{\LTpre}{6pt}
\setlength{\LTpost}{6pt}
\begin{longtable}{>{\raggedright\arraybackslash}p{0.10\dimexpr\linewidth-2\tabcolsep*4\relax}>{\raggedright\arraybackslash}p{0.25\dimexpr\linewidth-2\tabcolsep*4\relax}>{\raggedright\arraybackslash}p{0.32\dimexpr\linewidth-2\tabcolsep*4\relax}>{\raggedright\arraybackslash}p{0.33\dimexpr\linewidth-2\tabcolsep*4\relax}}
\caption{Task instances and fixed solver budgets. Test counts are per selected program and per setting.}\label{tab:protocol-data}\\
\toprule
Task & Search instances & Selected-program tests & Solver / metric \\
\midrule
\endfirsthead
\multicolumn{4}{l}{\small Table \thetable{} (continued)}\\
\toprule
Task & Search instances & Selected-program tests & Solver / metric \\
\midrule
\endhead
\bottomrule
\endfoot
TSP & 64 instances; 50 nodes & 50/100/200 nodes; 1,000 instances per size & Next-node rule; mean tour cost \\[3pt]
CVRP & 10 instances; 50 customers + depot & 50/100/200 customers + depot; 64 instances per size & ACO: 30 ants $\times$ 100 iterations; mean route cost \\[3pt]
OBP & 4 instances: 1k/5k items $\times$ capacities 100/500 & 1k/5k/10k items $\times$ capacities 100/500; 5 instances per setting & Bin-priority rule; gap to volume lower bound (\%) \\[3pt]
OP & 5 instances; 50 nodes & 50/100/200 nodes; 64 instances per size & ACO: 20 ants $\times$ 50 iterations; mean prize \\[3pt]
\end{longtable}
\endgroup

Prompts combine task and output requirements, operator instructions, and any selected parent programs with their performance and ideas; creation uses the task template. Responses must contain an idea and a deterministic Python function: \texttt{select\_next\_node\_v1} for TSP, \texttt{step} for OBP, or \texttt{heuristics} for CVRP and OP.

\subsection{Search, training, and execution settings}
\label{app:protocol-search}

SVC, Frozen, Native, Validity-only, Factorized, and Tail use Qwen2.5-7B-Instruct and CALM. Each task--condition pair has seven runs with seeds 42, 3407, 1926000, 2027, 271828, 314159, and 12345. Each run generates 500 groups of four responses, including invalid outputs. Frozen disables parameter updates while search states and prompts evolve. Four TSP variants each have seven runs: adding a failure penalty, using binary admission, removing distance credit while retaining distance-based parent selection, or combining rank weighting with a failure penalty. Shared settings appear in Table~\ref{tab:protocol-config}.

\begingroup
\small
\setlength{\LTpre}{6pt}
\setlength{\LTpost}{6pt}
\begin{longtable}{>{\raggedright\arraybackslash}p{0.30\dimexpr\linewidth-2\tabcolsep*2\relax}>{\raggedright\arraybackslash}p{0.70\dimexpr\linewidth-2\tabcolsep*2\relax}}
\caption{Local model, training, and search configuration. Training settings apply when parameter updates are enabled.}\label{tab:protocol-config}\\
\toprule
Setting & Value \\
\midrule
\endfirsthead
\multicolumn{2}{l}{\small Table \thetable{} (continued)}\\
\toprule
Setting & Value \\
\midrule
\endhead
\bottomrule
\endfoot
Base model & Qwen2.5-7B-Instruct; 4-bit loading \\[3pt]
LoRA & Rank 32; alpha 64; q/k/v/o and gate/up/down projections \\[3pt]
Training / optimizer & GRPO / adamw\_8bit \\[3pt]
Learning rate & 0.00005; constant schedule; no warmup \\[3pt]
Adam / decay & beta1 = 0.9; beta2 = 0.99; weight decay = 0.1 \\[3pt]
Batch / accumulation & Per-device batch 1; gradient accumulation 1; one prompt and four responses per call \\[3pt]
Gradient norm / policy KL & Maximum gradient norm 0.1; policy KL coefficient 0.04 \\[3pt]
Token limits & Prompt 2,048; response 1,024; model context 4,096 \\[3pt]
Temperature & 0.9 \\[3pt]
Response budget & 500 groups $\times$ 4 = 2,000 responses, including invalid outputs \\[3pt]
Primary parent pool & At most 10 retained programs ranked by fitness \\[3pt]
Operator weights & Simplification 1; injection 1; replacement 2; crossover 4 \\[3pt]
Population shrinkage & Stagnation threshold 25; random factor 0.0005 \\[3pt]
Candidate evaluation & 60-second execution timeout during search \\[3pt]
\end{longtable}
\endgroup

The parent-pool limit applies to primary parent selection, not the full retained set. Operator weights are adjusted to the search state before sampling; SVC is defined in Section~\ref{sec:svc} of the main article, with supplementary variants in \ref{app:credit-analysis}.

Recorded versions are PyTorch 2.5.1 \citep{paszke2019pytorch}, Transformers 4.49.0 \citep{wolf2020transformers}, TRL 0.15.1, Unsloth 2025.3.18, vLLM 0.7.3 \citep{kwon2023vllm}, PEFT 0.14.0, Ray 2.40.0 \citep{moritz2018ray}, and NumPy 1.26.4 \citep{harris2020numpy}. Among 168 main runs, 164 use Python 3.10.19 and four use 3.11.15. SVC has 21 runs on RTX 4090 and seven on RTX 3090; the controls have 136 and four. The RTX 3090 controls are CVRP Tail seed 1926000 and TSP Tail seeds 42, 3407, and 1926000. These three TSP runs use eight rather than four Ray workers and omit intermediate checkpoints. Seed pairing therefore does not imply identical execution environments.

We use the online model GPT-4o-mini as the API reference, with no parameter updates, seven search seeds for each of the four tasks (28 searches in total), the same response budget, and CALM interfaces. Task means give equal weight to the seven runs, and paired comparisons match runs by search seed. Each group shares its pre-generation context. Requests use temperature 0.9, top-$p=1$, and at most 1,024 output tokens. API prompts contain complete CALM messages; local Qwen prompts are left-truncated at 2,048 tokens, with a different tokenizer. No server-side sampling seed is sent: matched search seeds align initialization, not sampling noise. API results form a separate complete-system comparison, excluding selected-program and checkpoint evaluations.

API requests time out after 180 seconds, with up to five transport retries and backoffs of 30, 60, 120, 180, and 240 seconds. SDK and empty-content retries are disabled. The 28 runs required 56,009 attempts for 56,000 responses, including nine transport retries and no exhausted retries; retries do not enlarge the response budget.

\subsection{Selected-program tests}

Each main local run contributes its valid admitted program with the highest training fitness, taking the first matching record in a tie. Fitness is checked against the final search result, and program hashes link test records. The 168 programs yield 630 program--setting cells across 15 task-specific settings. All reported cells have complete finite-score instance records. An all-instance mean requires every requested instance to succeed; failures remain recorded and neither failures nor test performance trigger reselection.

TSP, OBP, and OP tests use a 300-second per-instance limit and worker seed $271828+i$ for zero-based instance index $i$. CVRP instead uses an outer task limit. CVRP and OP ACO generators use seed 42. All CVRP programs are tested on data cast to float64; the 50/100-customer files are stored as float64 and the 200-customer file as float32, whose lost precision cannot be restored by casting.

For OBP, define bins used $U_j$ and volume lower bound $L_j=\lceil\sum_k z_{jk}/Q_j\rceil$, where $z_{jk}$ is item size and $Q_j$ is capacity. Search fitness is minus the arithmetic mean of four gaps $100(U_j-L_j)/L_j$. Testing reports $100(\sum_j U_j-\sum_j L_j)/\sum_j L_j$ per file, weighting gaps by their lower bounds. The bound need not be attainable.

\subsection{Common-context generation and population reset}
\label{app:protocol-generation}

Common-context evaluation uses 128 TSP and 64 CVRP prompts from the corresponding Frozen seed-42 search, selected across operators and stages with a parent comparator. Each preserves the messages, parents, and donor group's pre-generation best retained fitness. These are search-derived, not unseen, contexts. Each checkpoint generates four responses per prompt at temperature 0.9, top-$p=1$, with a 1,024-token output limit and 60-second execution limit. Prompt/context limits are 2,048/4,096 tokens; no prompt truncation occurred.

Comparisons use the same host: TSP on RTX 3090 covers seven checkpoints each for SVC, Native, Validity-only, Factorized, and Tail; CVRP on RTX 4090 covers the same conditions except Factorized. One Frozen seed-42 checkpoint per task gives 36 TSP and 29 CVRP units. Internal TSP comparisons pair SVC with the penalty variant on RTX 3090 and with Admission-only and No-distance on RTX 4090.

Sampling uses seed $271828+i$ for prompt $i$. TSP Tail seeds 42, 3407, and 1926000 each combine two complementary 64-prompt segments; the second restarts offsets at 0--63 instead of 64--127. All 128 prompts are covered once by one checkpoint on the same host, but sampling streams differ from uninterrupted evaluation.

Reset uses final TSP SVC checkpoints 42, 3407, and 1926000, starts from the initial program, clears population and prompt history, and disables training. Each 500-group, four-response run is paired with Frozen. Optimizer steps, backward steps, and training tokens are zero; initial and final adapter checksums match the source. The adapter had not taken effect in vLLM for group one, whose responses match Frozen; groups two and three differ. Thus the adapter was not active for all 500 groups. The three pairs are reported individually and summarized with $n=3$.

\subsection{Shared-record replay and credit diagnostics}

Eight reward mappings are replayed on the same groups from 28 SVC trajectories without generation or training: SVC, Native, Validity-only, Factorized, Tail, the penalty variant, Admission-only, and No-distance. Four-response normalization uses sample standard deviation and stabilizer $10^{-4}$. Logs and replay use NumPy float64, versus float32 in the learner; rank statistics retain exact floating-point ties. Zero task advantage means a group's maximum absolute advantage is at most $10^{-10}$, not zero KL regularization or parameter change.

Admission--improvement counts use valid, non-initialization responses with finite parent comparators, counting admission flags rather than unique insertions. Later parent use and ancestor labels follow code identities and parent links in strictly later groups. Ancestor targets are valid, evaluated responses whose finite fitness strictly exceeds the finite best retained fitness before generation. Ancestor AUC uses valid admitted responses with code hashes and half credit for ties. Positive-advantage share is labeled responses' positive advantage divided by that of all trajectory responses. Shared code identities and unequal observation windows limit causal interpretation. TSP's seven trajectories are distinguished from the 21 non-TSP trajectories.

OBP plateaus require exactly constant best generated fitness for at least 100 consecutive groups. Within/outside rates use their respective group counts; absent plateaus have undefined rates, leaving six runs in the SVC within-plateau summary.

\Needspace{4\baselineskip}
\subsection{Metrics and statistical units}

Let $f^\star_{\ell,t}$ be best fitness after group $t$ in run $\ell$, with higher values preferred. Displayed quantities are $\mu_{\ell,t}=-f^\star_{\ell,t}$ for TSP, CVRP, and OBP, and $\mu_{\ell,t}=f^\star_{\ell,t}$ for OP. Table~\ref{tab:protocol-metrics} defines the metrics. Budget averages cover groups 1--500 without group zero, interpolation, or time weighting; groups 100/250/500 correspond to 400/1,000/2,000 responses.

\Needspace{4\baselineskip}
\begingroup
\small
\setlength{\LTpre}{6pt}
\setlength{\LTpost}{6pt}
\begin{longtable}{>{\raggedright\arraybackslash}p{0.32\dimexpr\linewidth-2\tabcolsep*2\relax}>{\raggedright\arraybackslash}p{0.68\dimexpr\linewidth-2\tabcolsep*2\relax}}
\caption{Search and common-context metrics. Proportions are computed per run or checkpoint before averaging over seeds.}\label{tab:protocol-metrics}\\
\toprule
Metric & Definition / denominator \\
\midrule
\endfirsthead
\multicolumn{2}{l}{\small Table \thetable{} (continued)}\\
\toprule
Metric & Definition / denominator \\
\midrule
\endhead
\bottomrule
\endfoot
Final task quantity & $\mu_{\ell,500}$ \\[3pt]
Budget average & $500^{-1}\sum_{t=1}^{500}\mu_{\ell,t}$ \\[3pt]
Late-search improvement & $f^\star_{\ell,500}-f^\star_{\ell,250}$; positive denotes within-run progress \\[3pt]
Validity rate & Valid finite-score responses / all responses \\[3pt]
Parent improvement yield & Valid responses strictly better than the best prompt parent / all responses \\[3pt]
Parent best-of-four & Prompts with at least one parent improvement / all prompts \\[3pt]
Frontier yield / best-of-four & Same response / prompt denominators, using the fixed donor frontier as comparator \\[3pt]
Parent ties (valid / all) & Valid responses within $10^{-7}$ of the parent comparator / valid responses or all responses \\[3pt]
Mean fitness among valid programs & Sum of valid-response fitness / valid responses; undefined if none are valid \\[3pt]
\end{longtable}
\endgroup

Validity requires passing interface and execution checks with a finite score. The parent comparator is the best prompt parent; the frontier is the donor group's fixed pre-generation best retained fitness. Best-of-four is the observed prompt-level proportion, not a mean best score or an unbiased pass-at-$k$ estimator.

Metrics are computed within each run, program, or checkpoint, then averaged equally across seeds. Benefits are control minus SVC for cost/gap and SVC minus control for prize/improvement rates. Generation-rate and OBP-gap differences use percentage points; a positive tie-rate difference denotes more ties. Relative benefits divide each seed's benefit by its absolute control value before averaging. Cross-size summaries average within each seed first.

Descriptive 95\% paired intervals use $\bar d\pm t_{n-1,0.975}s_d/\sqrt n$, with $n=7$ for main comparisons and $n=3$ for reset, assuming approximately normal seed differences without multiplicity adjustment. Pointwise trajectory intervals are not simultaneous bands. Responses, instances, repeated hosts, and scales do not add independent training runs; the single Frozen generation reference has no seven-pair interval.

\subsection{Run inventory}

Each task--condition--seed contributes one completed, audited full run, selected by completion and audit status rather than score. Mirrors and incomplete runs add no replicates, and every program and checkpoint evaluation is linked to the run used for that seed.

The inventory contains 224 searches: 168 main, 28 TSP variants, and 28 API references.

\subsection{Control reward definitions}

\label{app:control-rewards}

This subsection defines the five control conditions. As in the method, $v_{t,i}\in\{0,1\}$ indicates validity and $f_{t,i}$ denotes task fitness. Let $b_t$ denote the best prompt-base fitness when available. Conditions with parameter updates use the same GRPO normalization: rewards $\mathbf r_t$ are transformed to group-relative advantages $\widehat{\mathbf A}_t=\mathcal N_G(\mathbf r_t)$ by subtracting the group mean and dividing by the sample standard deviation plus stabilizer $10^{-4}$.

\paragraph{Frozen} Parameter updates are disabled; no training reward is used. Search states and prompts evolve normally.

\paragraph{Native CALM} This condition uses CALM's released reward mapping \citep{huang2026calm}. Typed failures receive penalties for missing rationale ($-1$), missing code ($-0.95$), missing required function (interface failure, $-0.90$), runtime failure ($-0.85$), and detected randomness ($-0.75$). Valid initialization candidates receive zero. Otherwise, let $f$ be the task score and $b$ the best prompt-base score. With
\begin{equation*}
\delta(f,b)=\mathrm{clip}\!\left(\frac{|f-b|}{\min\{|f|,|b|\}},10^{-10},1\right),
\end{equation*}
an improvement receives $1+\delta$, numerical equality receives zero, and a degradation receives $-3\delta/8$; the released base-equivalence branch, for a candidate identified as one of the prompt bases, receives $-3/5$. Native therefore combines failure severity, validity, and context-relative quality in one piecewise scale.

\paragraph{Validity-only} Rewards are $r_{t,i}=+1$ if $v_{t,i}=1$ (valid executable response with finite score) and $r_{t,i}=-1$ otherwise. No fitness comparisons are used.

\paragraph{Factorized Validity--Quality} The signed validity signal is $v^{\mathrm F}_{t,i}=2v_{t,i}-1\in\{-1,+1\}$. Let $V_t^{\mathrm F}$ contain valid non-initialization candidates with a prompt-base comparator, and let $\Delta_{t,i}=f_{t,i}-b_t$. The group-scaled quality component is
\begin{equation*}
q_{t,i}=\begin{cases}
\displaystyle\frac{\Delta_{t,i}}{\sqrt{|V_t^{\mathrm F}|^{-1}\sum_{j\in V_t^{\mathrm F}}\Delta_{t,j}^2}},&i\in V_t^{\mathrm F},\\[4pt]
0,&i\notin V_t^{\mathrm F}.
\end{cases}
\end{equation*}
Set $q_{t,i}=0$ when the eligible-set RMS is numerically zero or no valid candidate is eligible. The scalar reward is
\begin{equation*}
r_{t,i}=\tfrac{1}{2}v^{\mathrm F}_{t,i}+\tfrac{1}{2}q_{t,i}.
\end{equation*}
One joint GRPO normalization follows; this is not two separate losses.

\paragraph{Tail-weighted} This control adapts TTT-Discover's exponential concentration rule \citep{yuksekgonul2026tttdiscover} to CALM's contextual utility. A tie-aware midrank $u_{t,i}\in[0,1]$ places typed failures below valid candidates; among valid candidates, it favors scores absent from the pre-generation archive and responses not repeated within the group. Let $F_t$ be the best retained fitness before group generation (the frontier) and $V_t^{\mathrm T}$ the valid candidates for which this comparison is available. With $[x]_+=\max(0,x)$, RMS scaling preserves the relative magnitudes of positive frontier gaps within the group:
\begin{equation*}
\begin{aligned}
s_{g,t} &= \begin{cases}
\displaystyle\sqrt{|V_t^{\mathrm T}|^{-1}\sum_{j\in V_t^{\mathrm T}}[f_{t,j}-F_t]_+^2}, & V_t^{\mathrm T}\ne\varnothing,\\[4pt]
0, & V_t^{\mathrm T}=\varnothing,
\end{cases}\\[4pt]
g_{t,i} &= \begin{cases}
\displaystyle\frac{[f_{t,i}-F_t]_+}{s_{g,t}}, & i\in V_t^{\mathrm T}\text{ and }s_{g,t}>0,\\[4pt]
0, & \text{otherwise},
\end{cases}\\
w^{\mathrm{Tail}}_{t,i} &= u_{t,i}+\tfrac{1}{2}g_{t,i}.
\end{aligned}
\end{equation*}
The gap channel is zero when its denominator is zero. The superscript distinguishes the Tail utility $w^{\mathrm{Tail}}$ from SVC's parent-exposure weight $w$. Following TTT-Discover's concentration rule, but not its PUCT or leave-one-out objective, use
\begin{equation*}
\omega_{t,i}=\frac{\exp(\beta_t w^{\mathrm{Tail}}_{t,i})}{\sum_j\exp(\beta_t w^{\mathrm{Tail}}_{t,j})},\qquad r_{t,i}=G\omega_{t,i},
\end{equation*}
and choose $\beta_t\geq0$ to satisfy
\begin{equation*}
D_\mathrm{KL}(\boldsymbol\omega_t\|U_G)=\min\{\log 2,\log(G/k_t)\},
\end{equation*}
where $U_G$ is uniform over the $G$ responses and $k_t$ counts tied utility maxima. If the target equals the largest KL attainable under tied maxima, the limiting distribution is uniform over those maxima; a constant group returns uniform weights. This KL controls concentration over the completion group and is unrelated to the policy--reference KL coefficient in GRPO.

\FloatBarrier
\section{Complete search and selected-program test results}
\label{app:results}

\subsection{Search comparisons}
\label{app:search-results}

The tables in this appendix report the complete paired comparisons underlying the search results. SVC is compared with Frozen, Native CALM (Native), Validity-only, Factorized Validity--Quality (Factorized), and Tail-weighted (Tail), all using Qwen2.5-7B-Instruct. Each task and condition has seven search seeds: 42, 2027, 3407, 12345, 271828, 314159, and 1926000. Each search generates 500 groups of four responses, including invalid responses in the budget.

Tables~\ref{tab:svc-paired-final} and~\ref{tab:svc-paired-budget-mean} give endpoint and budget-averaged comparisons, respectively. The endpoint is the best-so-far task metric after 2,000 responses; the budget average is the arithmetic mean of that metric over groups 1--500 within each run. Tables~\ref{tab:svc-paired-g100} and~\ref{tab:svc-paired-g250} give the intermediate comparisons at 400 and 1,000 responses. Table~\ref{tab:svc-paired-second-half-gain} reports the difference in improvement from group 250 to group 500. This last quantity measures progress within each run and is distinct from endpoint performance. Together, these tables permit comparison of the endpoint ordering with performance over the search budget. The SVC--Tail paired endpoint benefits summarized in the main search-endpoint table (Table~\ref{tab:main-results} of the main article) correspond to the Tail rows of Table~\ref{tab:svc-paired-final}.

Tour cost, route cost, and OBP gap are minimized; OP prize is maximized. Paired benefit is control minus SVC for costs and gap, and SVC minus control for prize. Second-half improvement is measured in the increasing-fitness direction, and its paired difference is SVC improvement minus control improvement. OBP gaps are percentages and their differences are percentage points. Each interval is a descriptive 95\% paired interval, $\bar d\pm t_{6,0.975}s_d/\sqrt{7}$, where $s_d$ is the sample SD of the seven seed differences. These intervals use an approximately normal difference distribution as a working assumption. W/T/L denotes paired wins, ties, and losses for SVC, with tie tolerance $10^{-12}$.

The API experiments use GPT-4o-mini on all four tasks with the same seven search seeds per task, giving 28 searches. Each task mean averages the seven runs; the tables below report mean paired differences from SVC across the seven matched search seeds. Model parameters remain fixed during search. The request settings are specified in \ref{app:protocol-search}. The API is not included in the selected-program tests below.

\begingroup
\fontsize{10}{12}\selectfont
\setlength{\tabcolsep}{4pt}\renewcommand{\arraystretch}{1.05}
\setlength{\LTleft}{0pt}\setlength{\LTright}{0pt}
\setlength{\LTpre}{6pt}\setlength{\LTpost}{4pt}
\begin{longtable}{@{}lrrrc@{}}
\caption{Endpoint paired results at 2,000 responses. SVC has lower mean costs on TSP and CVRP and a lower mean gap on OBP than every listed comparator. Each comparison uses seven matched search seeds. Benefit is control minus SVC for costs and gap, and SVC minus control for prize. Positive values indicate cost or gap reductions, or prize increases.}\label{tab:svc-paired-final}\\
\toprule
Comparator & Mean benefit & Paired SD & Descriptive 95\% interval & W/T/L \\
\midrule
\endfirsthead
\multicolumn{5}{@{}l}{\small Table \thetable{} (continued)}\\
\toprule
Comparator & Mean benefit & Paired SD & Descriptive 95\% interval & W/T/L \\
\midrule
\endhead
\bottomrule
\endfoot
\bottomrule
\endlastfoot
\multicolumn{5}{@{}l}{\textbf{TSP} (cost)} \\*
Frozen & $+0.02834$ & 0.02725 & $[+0.00314,\ +0.05354]$ & 6/0/1 \\
Native & $+0.02219$ & 0.03074 & $[-0.00624,\ +0.05062]$ & 5/0/2 \\
Validity-only & $+0.02160$ & 0.03224 & $[-0.00822,\ +0.05142]$ & 5/0/2 \\
Factorized & $+0.03122$ & 0.02313 & $[+0.00983,\ +0.05261]$ & 7/0/0 \\
Tail & $+0.02263$ & 0.02475 & $[-0.00026,\ +0.04552]$ & 6/0/1 \\*
\textit{API system}$^\ast$ & $+0.04028$ & 0.02252 & $[+0.01946,\ +0.06111]$ & 7/0/0 \\
\addlinespace[6pt]
\multicolumn{5}{@{}l}{\textbf{CVRP} (cost)} \\*
Frozen & $+0.22376$ & 0.39193 & $[-0.13871,\ +0.58624]$ & 6/0/1 \\
Native & $+0.25872$ & 0.24970 & $[+0.02779,\ +0.48965]$ & 5/0/2 \\
Validity-only & $+0.17616$ & 0.28728 & $[-0.08953,\ +0.44185]$ & 5/0/2 \\
Factorized & $+0.38375$ & 0.30903 & $[+0.09794,\ +0.66956]$ & 7/0/0 \\
Tail & $+0.09246$ & 0.21601 & $[-0.10732,\ +0.29223]$ & 4/0/3 \\*
\textit{API system}$^\ast$ & $+0.16754$ & 0.15628 & $[+0.02300,\ +0.31207]$ & 6/0/1 \\
\addlinespace[6pt]
\multicolumn{5}{@{}l}{\textbf{OBP} (percentage points)} \\*
Frozen & $+0.41694$ & 0.47598 & $[-0.02327,\ +0.85714]$ & 6/0/1 \\
Native & $+0.25656$ & 0.37081 & $[-0.08638,\ +0.59950]$ & 6/0/1 \\
Validity-only & $+0.27793$ & 0.39835 & $[-0.09048,\ +0.64634]$ & 6/0/1 \\
Factorized & $+0.54876$ & 0.46835 & $[+0.11560,\ +0.98192]$ & 6/0/1 \\
Tail & $+0.03739$ & 0.16083 & $[-0.11136,\ +0.18613]$ & 3/0/4 \\*
\textit{API system}$^\ast$ & $+0.24411$ & 0.47571 & $[-0.19585,\ +0.68407]$ & 5/0/2 \\
\addlinespace[6pt]
\multicolumn{5}{@{}l}{\textbf{OP} (prize)} \\*
Frozen & $-0.04743$ & 0.19353 & $[-0.22641,\ +0.13155]$ & 4/0/3 \\
Native & $-0.01086$ & 0.15759 & $[-0.15660,\ +0.13489]$ & 3/0/4 \\
Validity-only & $-0.01171$ & 0.14026 & $[-0.14144,\ +0.11801]$ & 4/0/3 \\
Factorized & $+0.10686$ & 0.17349 & $[-0.05359,\ +0.26731]$ & 6/0/1 \\
Tail & $+0.00657$ & 0.33424 & $[-0.30255,\ +0.31569]$ & 3/0/4 \\*
\textit{API system}$^\ast$ & $+0.01343$ & 0.21597 & $[-0.18631,\ +0.21317]$ & 4/0/3 \\
\end{longtable}
\noindent{\footnotesize\raggedright Paired SD is the sample SD of the seven differences. Descriptive intervals are $\bar d\pm t_{6,0.975}s_d/\sqrt{7}$. W/T/L denotes SVC wins/ties/losses (tolerance $10^{-12}$). OBP differences are percentage points. $^\ast$API: GPT-4o-mini, with seven search seeds per task on all four tasks. Reported benefits are means of seven paired differences; request settings are given in \ref{app:protocol-search}.\par}
\endgroup

\begingroup
\fontsize{10}{12}\selectfont
\setlength{\tabcolsep}{4pt}\renewcommand{\arraystretch}{1.05}
\setlength{\LTleft}{0pt}\setlength{\LTright}{0pt}
\setlength{\LTpre}{6pt}\setlength{\LTpost}{4pt}
\begin{longtable}{@{}lrrrc@{}}
\caption{Paired budget-averaged best-so-far results. Each run is averaged over response groups 1--500, then the seven seeds are paired. Benefit is control minus SVC for costs and gap, and SVC minus control for prize. Positive values favor SVC.}\label{tab:svc-paired-budget-mean}\\
\toprule
Comparator & Mean benefit & Paired SD & Descriptive 95\% interval & W/T/L \\
\midrule
\endfirsthead
\multicolumn{5}{@{}l}{\small Table \thetable{} (continued)}\\
\toprule
Comparator & Mean benefit & Paired SD & Descriptive 95\% interval & W/T/L \\
\midrule
\endhead
\bottomrule
\endfoot
\bottomrule
\endlastfoot
\multicolumn{5}{@{}l}{\textbf{TSP} (cost)} \\*
Frozen & $+0.02056$ & 0.02185 & $[+0.00035,\ +0.04076]$ & 6/0/1 \\
Native & $+0.01351$ & 0.03111 & $[-0.01526,\ +0.04227]$ & 5/0/2 \\
Validity-only & $+0.01355$ & 0.02518 & $[-0.00974,\ +0.03684]$ & 5/0/2 \\
Factorized & $+0.02011$ & 0.01774 & $[+0.00371,\ +0.03651]$ & 6/0/1 \\
Tail & $+0.01292$ & 0.02278 & $[-0.00815,\ +0.03398]$ & 5/0/2 \\*
\textit{API system}$^\ast$ & $+0.02746$ & 0.02047 & $[+0.00853,\ +0.04639]$ & 6/0/1 \\
\addlinespace[6pt]
\multicolumn{5}{@{}l}{\textbf{CVRP} (cost)} \\*
Frozen & $+0.38053$ & 0.59676 & $[-0.17138,\ +0.93245]$ & 6/0/1 \\
Native & $+0.29357$ & 0.32911 & $[-0.01081,\ +0.59794]$ & 5/0/2 \\
Validity-only & $+0.19515$ & 0.35050 & $[-0.12901,\ +0.51930]$ & 5/0/2 \\
Factorized & $+0.58757$ & 0.31092 & $[+0.30002,\ +0.87512]$ & 7/0/0 \\
Tail & $+0.11282$ & 0.22060 & $[-0.09120,\ +0.31684]$ & 5/0/2 \\*
\textit{API system}$^\ast$ & $+0.01561$ & 0.16299 & $[-0.13513,\ +0.16635]$ & 3/0/4 \\
\addlinespace[6pt]
\multicolumn{5}{@{}l}{\textbf{OBP} (percentage points)} \\*
Frozen & $+0.17227$ & 0.54416 & $[-0.33100,\ +0.67553]$ & 5/0/2 \\
Native & $+0.09676$ & 0.40336 & $[-0.27629,\ +0.46980]$ & 3/0/4 \\
Validity-only & $+0.07366$ & 0.41400 & $[-0.30923,\ +0.45655]$ & 4/0/3 \\
Factorized & $+0.30266$ & 0.37437 & $[-0.04357,\ +0.64889]$ & 6/0/1 \\
Tail & $-0.07828$ & 0.39641 & $[-0.44490,\ +0.28834]$ & 3/0/4 \\*
\textit{API system}$^\ast$ & $+0.14273$ & 0.37119 & $[-0.20056,\ +0.48603]$ & 5/0/2 \\
\addlinespace[6pt]
\multicolumn{5}{@{}l}{\textbf{OP} (prize)} \\*
Frozen & $+0.07425$ & 0.19796 & $[-0.10883,\ +0.25732]$ & 4/0/3 \\
Native & $-0.00732$ & 0.19085 & $[-0.18382,\ +0.16918]$ & 4/0/3 \\
Validity-only & $-0.01934$ & 0.21769 & $[-0.22067,\ +0.18199]$ & 4/0/3 \\
Factorized & $+0.13004$ & 0.23567 & $[-0.08792,\ +0.34800]$ & 5/0/2 \\
Tail & $-0.03513$ & 0.35515 & $[-0.36359,\ +0.29333]$ & 2/0/5 \\*
\textit{API system}$^\ast$ & $-0.06347$ & 0.17205 & $[-0.22259,\ +0.09565]$ & 3/0/4 \\
\end{longtable}
\noindent{\footnotesize\raggedright The arithmetic budget average includes no group 0, smoothing, trapezoidal integration or time weighting. Paired SD uses the seven differences. Descriptive intervals are $\bar d\pm t_{6,0.975}s_d/\sqrt{7}$. W/T/L denotes SVC wins/ties/losses (tolerance $10^{-12}$); OBP differences use percentage points. $^\ast$API: GPT-4o-mini, with seven search seeds per task on all four tasks. Reported benefits are means of seven paired differences; request settings are given in \ref{app:protocol-search}.\par}
\endgroup

\begingroup
\fontsize{10}{12}\selectfont
\setlength{\tabcolsep}{4pt}\renewcommand{\arraystretch}{1.05}
\setlength{\LTleft}{0pt}\setlength{\LTright}{0pt}
\setlength{\LTpre}{6pt}\setlength{\LTpost}{4pt}
\begin{longtable}{@{}lrrrc@{}}
\caption{Paired best-so-far results at 400 responses (group 100). Each comparison uses seven matched search seeds. Benefit is control minus SVC for costs and gap, and SVC minus control for prize. Positive values favor SVC.}\label{tab:svc-paired-g100}\\
\toprule
Comparator & Mean benefit & Paired SD & Descriptive 95\% interval & W/T/L \\
\midrule
\endfirsthead
\multicolumn{5}{@{}l}{\small Table \thetable{} (continued)}\\
\toprule
Comparator & Mean benefit & Paired SD & Descriptive 95\% interval & W/T/L \\
\midrule
\endhead
\bottomrule
\endfoot
\bottomrule
\endlastfoot
\multicolumn{5}{@{}l}{\textbf{TSP} (cost)} \\*
Frozen & $+0.01060$ & 0.03448 & $[-0.02129,\ +0.04248]$ & 5/0/2 \\
Native & $+0.00685$ & 0.04505 & $[-0.03481,\ +0.04851]$ & 4/0/3 \\
Validity-only & $+0.00975$ & 0.03123 & $[-0.01913,\ +0.03864]$ & 4/1/2 \\
Factorized & $+0.01257$ & 0.03303 & $[-0.01799,\ +0.04312]$ & 5/0/2 \\
Tail & $+0.00282$ & 0.03517 & $[-0.02970,\ +0.03535]$ & 5/0/2 \\*
\textit{API system}$^\ast$ & $+0.01300$ & 0.03326 & $[-0.01776,\ +0.04376]$ & 5/0/2 \\
\addlinespace[6pt]
\multicolumn{5}{@{}l}{\textbf{CVRP} (cost)} \\*
Frozen & $+0.46275$ & 0.89408 & $[-0.36414,\ +1.28964]$ & 5/0/2 \\
Native & $+0.43774$ & 0.33517 & $[+0.12777,\ +0.74772]$ & 7/0/0 \\
Validity-only & $+0.28132$ & 0.41760 & $[-0.10490,\ +0.66753]$ & 6/0/1 \\
Factorized & $+0.87578$ & 0.50092 & $[+0.41251,\ +1.33906]$ & 7/0/0 \\
Tail & $+0.14614$ & 0.27652 & $[-0.10960,\ +0.40187]$ & 6/0/1 \\*
\textit{API system}$^\ast$ & $+0.10784$ & 0.20504 & $[-0.08180,\ +0.29747]$ & 5/0/2 \\
\addlinespace[6pt]
\multicolumn{5}{@{}l}{\textbf{OBP} (percentage points)} \\*
Frozen & $-0.13676$ & 0.57110 & $[-0.66495,\ +0.39142]$ & 3/0/4 \\
Native & $-0.01694$ & 0.54254 & $[-0.51870,\ +0.48483]$ & 3/1/3 \\
Validity-only & $-0.21291$ & 0.33512 & $[-0.52285,\ +0.09702]$ & 1/1/5 \\
Factorized & $+0.10068$ & 0.37071 & $[-0.24216,\ +0.44353]$ & 3/0/4 \\
Tail & $-0.02720$ & 0.59237 & $[-0.57505,\ +0.52065]$ & 3/1/3 \\*
\textit{API system}$^\ast$ & $+0.07218$ & 0.59594 & $[-0.47897,\ +0.62334]$ & 5/0/2 \\
\addlinespace[6pt]
\multicolumn{5}{@{}l}{\textbf{OP} (prize)} \\*
Frozen & $+0.42343$ & 0.50320 & $[-0.04195,\ +0.88881]$ & 7/0/0 \\
Native & $+0.12429$ & 0.29368 & $[-0.14732,\ +0.39589]$ & 4/0/3 \\
Validity-only & $+0.04314$ & 0.24667 & $[-0.18499,\ +0.27128]$ & 4/0/3 \\
Factorized & $+0.21771$ & 0.36675 & $[-0.12147,\ +0.55690]$ & 5/0/2 \\
Tail & $+0.06829$ & 0.46613 & $[-0.36281,\ +0.49938]$ & 3/0/4 \\*
\textit{API system}$^\ast$ & $+0.06343$ & 0.24490 & $[-0.16307,\ +0.28993]$ & 4/0/3 \\
\end{longtable}
\noindent{\footnotesize\raggedright Paired SD is the sample SD of the differences. Descriptive intervals are $\bar d\pm t_{6,0.975}s_d/\sqrt{7}$; these pointwise intervals are not simultaneous curve bands. W/T/L denotes SVC wins/ties/losses (tolerance $10^{-12}$); OBP differences use percentage points. $^\ast$API: GPT-4o-mini, with seven search seeds per task on all four tasks. Reported benefits are means of seven paired differences; request settings are given in \ref{app:protocol-search}.\par}
\endgroup

\begingroup
\fontsize{10}{12}\selectfont
\setlength{\tabcolsep}{4pt}\renewcommand{\arraystretch}{1.05}
\setlength{\LTleft}{0pt}\setlength{\LTright}{0pt}
\setlength{\LTpre}{6pt}\setlength{\LTpost}{4pt}
\begin{longtable}{@{}lrrrc@{}}
\caption{Paired best-so-far results at 1,000 responses (group 250). Each comparison uses seven matched search seeds. Benefit is control minus SVC for costs and gap, and SVC minus control for prize. Positive values favor SVC.}\label{tab:svc-paired-g250}\\
\toprule
Comparator & Mean benefit & Paired SD & Descriptive 95\% interval & W/T/L \\
\midrule
\endfirsthead
\multicolumn{5}{@{}l}{\small Table \thetable{} (continued)}\\
\toprule
Comparator & Mean benefit & Paired SD & Descriptive 95\% interval & W/T/L \\
\midrule
\endhead
\bottomrule
\endfoot
\bottomrule
\endlastfoot
\multicolumn{5}{@{}l}{\textbf{TSP} (cost)} \\*
Frozen & $+0.02990$ & 0.01724 & $[+0.01395,\ +0.04584]$ & 7/0/0 \\
Native & $+0.02035$ & 0.03069 & $[-0.00804,\ +0.04873]$ & 6/0/1 \\
Validity-only & $+0.01777$ & 0.02773 & $[-0.00787,\ +0.04341]$ & 5/0/2 \\
Factorized & $+0.02405$ & 0.01542 & $[+0.00979,\ +0.03831]$ & 7/0/0 \\
Tail & $+0.02019$ & 0.02462 & $[-0.00257,\ +0.04296]$ & 5/0/2 \\*
\textit{API system}$^\ast$ & $+0.03470$ & 0.01976 & $[+0.01642,\ +0.05298]$ & 7/0/0 \\
\addlinespace[6pt]
\multicolumn{5}{@{}l}{\textbf{CVRP} (cost)} \\*
Frozen & $+0.31016$ & 0.48948 & $[-0.14253,\ +0.76285]$ & 6/0/1 \\
Native & $+0.39189$ & 0.40422 & $[+0.01805,\ +0.76573]$ & 5/0/2 \\
Validity-only & $+0.23952$ & 0.36427 & $[-0.09737,\ +0.57642]$ & 5/0/2 \\
Factorized & $+0.56061$ & 0.39519 & $[+0.19511,\ +0.92610]$ & 7/0/0 \\
Tail & $+0.10480$ & 0.19838 & $[-0.07867,\ +0.28827]$ & 5/0/2 \\*
\textit{API system}$^\ast$ & $+0.15049$ & 0.17214 & $[-0.00871,\ +0.30969]$ & 5/0/2 \\
\addlinespace[6pt]
\multicolumn{5}{@{}l}{\textbf{OBP} (percentage points)} \\*
Frozen & $+0.12652$ & 0.74104 & $[-0.55882,\ +0.81187]$ & 4/0/3 \\
Native & $+0.09977$ & 0.65129 & $[-0.50258,\ +0.70212]$ & 3/0/4 \\
Validity-only & $+0.09088$ & 0.59902 & $[-0.46312,\ +0.64487]$ & 4/0/3 \\
Factorized & $+0.35992$ & 0.55479 & $[-0.15318,\ +0.87302]$ & 5/0/2 \\
Tail & $-0.09444$ & 0.69450 & $[-0.73675,\ +0.54787]$ & 3/0/4 \\*
\textit{API system}$^\ast$ & $+0.27977$ & 0.64921 & $[-0.32065,\ +0.88019]$ & 5/0/2 \\
\addlinespace[6pt]
\multicolumn{5}{@{}l}{\textbf{OP} (prize)} \\*
Frozen & $+0.00171$ & 0.16413 & $[-0.15008,\ +0.15351]$ & 2/0/5 \\
Native & $+0.05286$ & 0.21162 & $[-0.14286,\ +0.24857]$ & 5/0/2 \\
Validity-only & $+0.02943$ & 0.20754 & $[-0.16251,\ +0.22137]$ & 5/0/2 \\
Factorized & $+0.16686$ & 0.22325 & $[-0.03961,\ +0.37332]$ & 5/0/2 \\
Tail & $-0.01229$ & 0.33579 & $[-0.32284,\ +0.29827]$ & 3/0/4 \\*
\textit{API system}$^\ast$ & $-0.00400$ & 0.18096 & $[-0.17136,\ +0.16336]$ & 3/0/4 \\
\end{longtable}
\noindent{\footnotesize\raggedright Paired SD is the sample SD of the differences. Descriptive intervals are $\bar d\pm t_{6,0.975}s_d/\sqrt{7}$; these pointwise intervals are not simultaneous curve bands. W/T/L denotes SVC wins/ties/losses (tolerance $10^{-12}$); OBP differences use percentage points. $^\ast$API: GPT-4o-mini, with seven search seeds per task on all four tasks. Reported benefits are means of seven paired differences; request settings are given in \ref{app:protocol-search}.\par}
\endgroup

\begingroup
\fontsize{10}{12}\selectfont
\setlength{\tabcolsep}{4pt}\renewcommand{\arraystretch}{1.05}
\setlength{\LTleft}{0pt}\setlength{\LTright}{0pt}
\setlength{\LTpre}{6pt}\setlength{\LTpost}{4pt}
\begin{longtable}{@{}lrrrc@{}}
\caption{Exploratory comparison of second-half improvement. Within each run, gain is internal fitness at group 500 minus fitness at group 250. The paired difference is SVC gain minus control gain, summarized over seven matched search seeds.}\label{tab:svc-paired-second-half-gain}\\
\toprule
Comparator & Mean benefit & Paired SD & Descriptive 95\% interval & W/T/L \\
\midrule
\endfirsthead
\multicolumn{5}{@{}l}{\small Table \thetable{} (continued)}\\
\toprule
Comparator & Mean benefit & Paired SD & Descriptive 95\% interval & W/T/L \\
\midrule
\endhead
\bottomrule
\endfoot
\bottomrule
\endlastfoot
\multicolumn{5}{@{}l}{\textbf{TSP} (cost)} \\*
Frozen & $-0.00156$ & 0.01181 & $[-0.01248,\ +0.00936]$ & 3/0/4 \\
Native & $+0.00184$ & 0.01233 & $[-0.00957,\ +0.01325]$ & 4/0/3 \\
Validity-only & $+0.00383$ & 0.00708 & $[-0.00272,\ +0.01037]$ & 4/0/3 \\
Factorized & $+0.00717$ & 0.00950 & $[-0.00162,\ +0.01596]$ & 5/1/1 \\
Tail & $+0.00244$ & 0.00541 & $[-0.00257,\ +0.00744]$ & 4/0/3 \\*
\textit{API system}$^\ast$ & $+0.00559$ & 0.00763 & $[-0.00147,\ +0.01264]$ & 6/0/1 \\
\addlinespace[6pt]
\multicolumn{5}{@{}l}{\textbf{CVRP} (cost)} \\*
Frozen & $-0.08640$ & 0.12528 & $[-0.20226,\ +0.02946]$ & 2/0/5 \\
Native & $-0.13317$ & 0.21591 & $[-0.33285,\ +0.06651]$ & 2/0/5 \\
Validity-only & $-0.06336$ & 0.10823 & $[-0.16346,\ +0.03673]$ & 1/0/6 \\
Factorized & $-0.17686$ & 0.30610 & $[-0.45995,\ +0.10624]$ & 2/1/4 \\
Tail & $-0.01234$ & 0.07173 & $[-0.07868,\ +0.05400]$ & 2/1/4 \\*
\textit{API system}$^\ast$ & $+0.01705$ & 0.05201 & $[-0.03106,\ +0.06515]$ & 4/1/2 \\
\addlinespace[6pt]
\multicolumn{5}{@{}l}{\textbf{OBP} (percentage points)} \\*
Frozen & $+0.29041$ & 0.43595 & $[-0.11277,\ +0.69360]$ & 5/0/2 \\
Native & $+0.15679$ & 0.44186 & $[-0.25186,\ +0.56544]$ & 6/0/1 \\
Validity-only & $+0.18705$ & 0.53069 & $[-0.30375,\ +0.67786]$ & 5/0/2 \\
Factorized & $+0.18884$ & 0.43459 & $[-0.21309,\ +0.59077]$ & 5/0/2 \\
Tail & $+0.13183$ & 0.58694 & $[-0.41100,\ +0.67465]$ & 4/0/3 \\*
\textit{API system}$^\ast$ & $-0.03566$ & 0.45905 & $[-0.46021,\ +0.38889]$ & 4/0/3 \\
\addlinespace[6pt]
\multicolumn{5}{@{}l}{\textbf{OP} (prize)} \\*
Frozen & $-0.04914$ & 0.13805 & $[-0.17682,\ +0.07853]$ & 3/0/4 \\
Native & $-0.06371$ & 0.15180 & $[-0.20411,\ +0.07668]$ & 2/0/5 \\
Validity-only & $-0.04114$ & 0.10213 & $[-0.13560,\ +0.05331]$ & 2/1/4 \\
Factorized & $-0.06000$ & 0.15441 & $[-0.20280,\ +0.08280]$ & 3/1/3 \\
Tail & $+0.01886$ & 0.06898 & $[-0.04494,\ +0.08265]$ & 4/1/2 \\*
\textit{API system}$^\ast$ & $+0.01743$ & 0.06066 & $[-0.03867,\ +0.07353]$ & 4/1/2 \\
\end{longtable}
\noindent{\footnotesize\raggedright Positive values mean greater second-half improvement for SVC, not necessarily a better endpoint or budget average. These differences do not establish learned long-term planning. Paired SD uses the seven differences. Descriptive intervals are $\bar d\pm t_{6,0.975}s_d/\sqrt{7}$. W/T/L counts positive/zero/negative differences (tolerance $10^{-12}$); OBP units are percentage points. $^\ast$API: GPT-4o-mini, with seven search seeds per task on all four tasks. Reported benefits are means of seven paired differences; request settings are given in \ref{app:protocol-search}.\par}
\endgroup

\subsection{Budget gaps and endpoint ranks}
\label{app:budget-ranks}

\begin{figure}[!htbp]
\centering
\includegraphics[width=\linewidth]{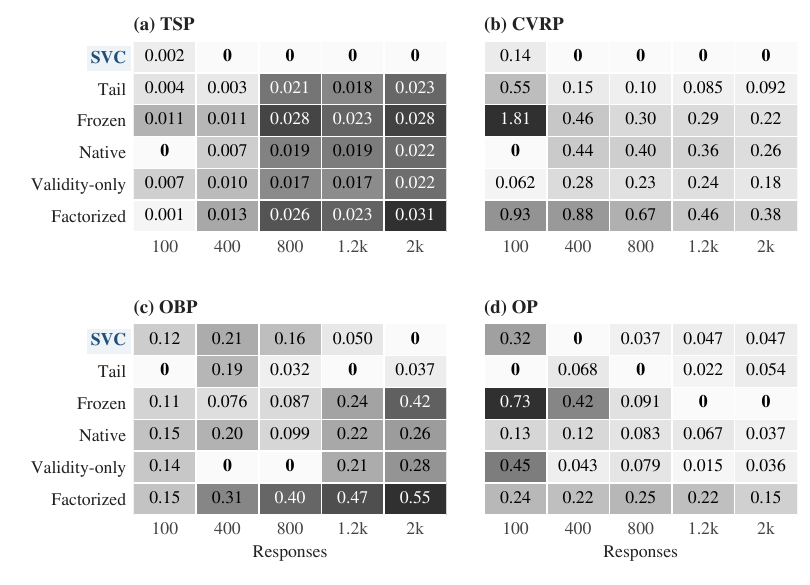}
\caption{%
Gaps to the best condition mean at 100, 400, 800, 1,200 and 2,000 responses. Each cell gives the excess tour cost (TSP), excess route cost (CVRP), excess gap in percentage points (OBP), or prize deficit (OP), relative to the best mean among the six conditions at that budget. Means use seven search seeds. Darker gray indicates a larger gap; the scale runs from zero to the largest displayed gap within each task and is fixed across that task's budgets. Shades are not comparable across tasks. Bold zeros mark the best observed means, without implying statistical significance. SVC is listed first with a navy label. SVC has the best mean at the four displayed budgets from 400 to 2,000 responses on TSP and CVRP.
}
\label{fig:app-search-budget-gaps}
\end{figure}

\begin{figure}[!htbp]
\centering
\includegraphics[width=\linewidth]{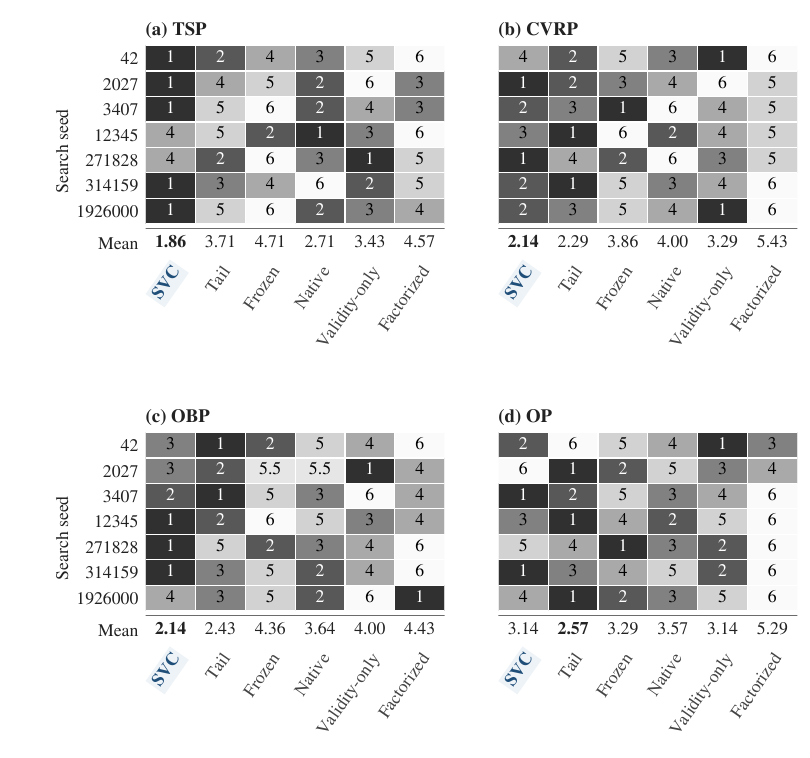}
\caption{%
Endpoint ranks after 2,000 responses within each matched search seed. Rank 1 is best and rank 6 worst, with smaller tour cost, route cost and gap preferred and larger prize preferred. Values tied within $10^{-12}$ receive their average rank. Darker gray indicates a better rank on a shared 1--6 scale. SVC is the first column with a navy label; all seven seeds and six conditions are shown. The unshaded bottom row gives the arithmetic mean of the seven displayed ranks, rounded to two decimals. Bold mean ranks mark the lowest mean rank within each task: SVC on TSP, CVRP and OBP, and Tail on OP. Ranks describe ordering rather than effect size or statistical significance.
}
\label{fig:app-seed-endpoint-ranks}
\end{figure}

Figure~\ref{fig:app-search-budget-gaps} compares each condition's mean with the best condition mean at five response budgets. Figure~\ref{fig:app-seed-endpoint-ranks} reports endpoint ranks within each matched seed. These summaries describe budget-dependent mean differences and seed-level ordering, respectively.

\FloatBarrier
\subsection{Selected-program tests}
\label{app:program-tests}

Tables~\ref{tab:svc-heldout-means-routing} and~\ref{tab:svc-heldout-means-obp} report all six local conditions at all 15 task-specific test settings. TSP and OP use 50, 100, and 200 nodes; CVRP uses 50, 100, and 200 customers plus a depot. OBP combines 1,000, 5,000, and 10,000 items with bin capacities 100 and 500. Each program is selected by training fitness and evaluated without reselection on the test data. There are 168 selected programs and 630 program--setting cells.

Each program metric aggregates 1,000 instances for TSP, 64 for CVRP or OP, and five for each OBP setting. For OBP, the gap is $100(\sum_j U_j-\sum_j L_j)/\sum_j L_j$, where $U_j$ and $L_j$ are the bin count and volume lower bound. Search instead averages the four individual instance gaps. All compared CVRP programs use float64 execution. Every requested instance in the reported cells has a finite valid result. Means and sample SDs summarize seven program metrics with equal seed weights; test instances and test settings do not increase the number of independent search runs.

Tables~\ref{tab:svc-heldout-paired-tsp}--\ref{tab:svc-heldout-paired-op_aco} give all five control comparisons at every setting, using the same benefit direction and paired-interval definition as the search tables. Large program-level differences remain included: the Tail program from CVRP seed 271828 has mean costs of 41.23354 and 58.50423 at 100 and 200 customers, and the SVC program from OP seed 2027 has mean prize 28.68409 at 200 nodes. These valid program outcomes contribute to the reported means and SDs. The tests measure the performance of selected programs on new instances and sizes.

\begingroup
\fontsize{10}{12}\selectfont
\setlength{\tabcolsep}{3.5pt}
\renewcommand{\arraystretch}{1.08}
\setlength{\LTleft}{0pt}\setlength{\LTright}{0pt}
\setlength{\LTpre}{8pt}\setlength{\LTpost}{2pt}
\begin{longtable}{@{\extracolsep{\fill}}lrrr@{}}
\caption{Selected-program test results for TSP, CVRP and OP. SVC has the best mean at all TSP sizes and at 50 and 100 CVRP customers. Values are mean $\pm$ sample SD over seven training seeds. Columns give nodes for TSP/OP and customers for CVRP (with an additional depot). Arrows indicate the favorable direction.}\label{tab:svc-heldout-means-routing}\\
\endfirsthead
\multicolumn{4}{@{}l}{\small Table~\thetable{} (continued)}\\
\toprule
\endhead
\bottomrule
\endfoot
\toprule
\multicolumn{4}{@{}l}{\textbf{TSP}: Tour cost, $\downarrow$} \\*
Condition & 50 & 100 & 200 \\*
\midrule
SVC & $\mathbf{6.24393}\,\pm\,0.01510$ & $\mathbf{8.70728}\,\pm\,0.04047$ & $\mathbf{12.20907}\,\pm\,0.10220$ \\
Frozen & $6.24762\,\pm\,0.01501$ & $8.71067\,\pm\,0.05461$ & $12.22683\,\pm\,0.10112$ \\
Native & $6.25442\,\pm\,0.00846$ & $8.72629\,\pm\,0.03133$ & $12.25553\,\pm\,0.06593$ \\
Validity-only & $6.26550\,\pm\,0.01387$ & $8.72841\,\pm\,0.02312$ & $12.21699\,\pm\,0.08227$ \\
Factorized & $6.26030\,\pm\,0.00940$ & $8.71780\,\pm\,0.02586$ & $12.22228\,\pm\,0.08007$ \\
Tail & $6.25930\,\pm\,0.00938$ & $8.73525\,\pm\,0.02939$ & $12.26704\,\pm\,0.07454$ \\
\addlinespace[7pt]
\noalign{\penalty-10000}
\multicolumn{4}{@{}l}{\textbf{CVRP}: Route cost, $\downarrow$} \\*
Condition & 50 & 100 & 200 \\*
\midrule
SVC & $\mathbf{9.27922}\,\pm\,0.16930$ & $\mathbf{15.82433}\,\pm\,0.22526$ & $28.29313\,\pm\,0.66528$ \\
Frozen & $9.41610\,\pm\,0.35308$ & $15.84684\,\pm\,0.49413$ & $\mathbf{28.19590}\,\pm\,0.66462$ \\
Native & $9.44442\,\pm\,0.12592$ & $16.09634\,\pm\,0.39022$ & $28.81669\,\pm\,0.74785$ \\
Validity-only & $9.38443\,\pm\,0.17705$ & $15.96770\,\pm\,0.23478$ & $28.57215\,\pm\,0.32426$ \\
Factorized & $9.55955\,\pm\,0.28232$ & $16.27478\,\pm\,0.77523$ & $29.17247\,\pm\,1.59662$ \\
Tail & $9.36023\,\pm\,0.11288$ & $19.75206\,\pm\,9.52121$ & $33.39339\,\pm\,11.37542$ \\
\addlinespace[7pt]
\multicolumn{4}{@{}l}{\textbf{OP}: Prize, $\uparrow$} \\*
Condition & 50 & 100 & 200 \\*
\midrule
SVC & $15.03414\,\pm\,0.22633$ & $30.32694\,\pm\,0.89665$ & $50.40584\,\pm\,9.92413$ \\
Frozen & $15.14775\,\pm\,0.05169$ & $30.64696\,\pm\,0.47778$ & $54.45595\,\pm\,2.42635$ \\
Native & $\mathbf{15.16239}\,\pm\,0.09441$ & $30.47622\,\pm\,0.77829$ & $53.41457\,\pm\,3.84400$ \\
Validity-only & $15.01973\,\pm\,0.27263$ & $30.33508\,\pm\,1.08211$ & $54.06588\,\pm\,3.00240$ \\
Factorized & $15.05818\,\pm\,0.22532$ & $30.39946\,\pm\,0.44504$ & $53.84365\,\pm\,1.74839$ \\
Tail & $15.02151\,\pm\,0.32462$ & $\mathbf{30.76118}\,\pm\,0.51407$ & $\mathbf{55.53223}\,\pm\,1.18130$ \\
\end{longtable}
\endgroup
\par\vspace{6pt}
\noindent\begin{minipage}{\linewidth}\fontsize{10}{12}\selectfont\raggedright
Bold marks the best observed mean among the six conditions at that setting; it does not denote statistical significance. Each program is selected using training fitness. Its test metric is first averaged over 1,000 instances for TSP or 64 for CVRP/OP; the table then averages seven program metrics. All requested instances yielded finite valid scores. CVRP uses float64 execution; the high Tail costs remain included. Scientific units differ between tasks.
\end{minipage}\par\medskip

\begingroup
\fontsize{10}{12}\selectfont
\setlength{\tabcolsep}{3.5pt}
\renewcommand{\arraystretch}{1.08}
\setlength{\LTleft}{0pt}\setlength{\LTright}{0pt}
\setlength{\LTpre}{8pt}\setlength{\LTpost}{2pt}
\begin{longtable}{@{\extracolsep{\fill}}lrrr@{}}
\caption{Selected-program OBP test results across item counts and bin capacities. Values are gap percentages, reported as mean $\pm$ sample SD over seven training seeds; lower is better. Columns give item counts, with 1k denoting 1,000 items.}\label{tab:svc-heldout-means-obp}\\
\endfirsthead
\multicolumn{4}{@{}l}{\small Table~\thetable{} (continued)}\\
\toprule
\endhead
\bottomrule
\endfoot
\toprule
\multicolumn{4}{@{}l}{\textbf{OBP; capacity 100}: Gap (\%), $\downarrow$} \\*
Condition & 1k & 5k & 10k \\*
\midrule
SVC & $\mathbf{3.35132}\,\pm\,0.76561$ & $\mathbf{1.07811}\,\pm\,0.32785$ & $0.98240\,\pm\,0.63768$ \\
Frozen & $4.29565\,\pm\,0.71902$ & $2.19160\,\pm\,1.49337$ & $1.76817\,\pm\,1.61842$ \\
Native & $3.72763\,\pm\,0.61612$ & $1.77988\,\pm\,1.13484$ & $1.23459\,\pm\,1.29828$ \\
Validity-only & $3.81994\,\pm\,0.49985$ & $1.67235\,\pm\,1.18394$ & $1.26522\,\pm\,1.31124$ \\
Factorized & $4.16075\,\pm\,0.57226$ & $2.24395\,\pm\,1.61730$ & $1.88500\,\pm\,1.78956$ \\
Tail & $3.74183\,\pm\,0.49678$ & $1.11348\,\pm\,0.25191$ & $\mathbf{0.67535}\,\pm\,0.27624$ \\
\addlinespace[7pt]
\multicolumn{4}{@{}l}{\textbf{OBP; capacity 500}: Gap (\%), $\downarrow$} \\*
Condition & 1k & 5k & 10k \\*
\midrule
SVC & $0.53173\,\pm\,0.54418$ & $\mathbf{0.36211}\,\pm\,0.15106$ & $\mathbf{0.33471}\,\pm\,0.12199$ \\
Frozen & $\mathbf{0.38993}\,\pm\,0.40157$ & $0.44022\,\pm\,0.18470$ & $0.35964\,\pm\,0.19246$ \\
Native & $0.46083\,\pm\,0.41592$ & $0.41892\,\pm\,0.23287$ & $0.35251\,\pm\,0.20117$ \\
Validity-only & $0.60262\,\pm\,0.55221$ & $0.44022\,\pm\,0.21556$ & $0.47002\,\pm\,0.40433$ \\
Factorized & $0.70897\,\pm\,0.86468$ & $0.52542\,\pm\,0.12787$ & $0.42017\,\pm\,0.14564$ \\
Tail & $0.92166\,\pm\,0.85616$ & $0.47572\,\pm\,0.20262$ & $0.38456\,\pm\,0.20697$ \\
\end{longtable}
\endgroup
\par\vspace{6pt}
\noindent\begin{minipage}{\linewidth}\fontsize{10}{12}\selectfont\raggedright
Within each program and setting, the metric is $100(\sum_j U_j-\sum_j L_j)/\sum_j L_j$ over five instances, where $U_j$ is the number of bins used and $L_j$ is the volume lower bound. It is not the unweighted mean of per-instance gaps. Bold marks the lowest observed mean among the six conditions, not statistical significance. All requested instances are included.
\end{minipage}\par\medskip

\begingroup
\fontsize{10}{12}\selectfont
\setlength{\tabcolsep}{3.5pt}
\renewcommand{\arraystretch}{1.08}
\setlength{\LTleft}{0pt}\setlength{\LTright}{0pt}
\setlength{\LTpre}{8pt}\setlength{\LTpost}{2pt}
\begin{longtable}{@{\extracolsep{\fill}}lrrrc@{}}
\caption{TSP selected-program paired comparisons. Benefit is the control minus SVC, in tour cost; positive values favor SVC. Every setting includes all five controls.}\label{tab:svc-heldout-paired-tsp}\\
\toprule
Control & Mean benefit & Paired SD & 95\% interval & W/T/L \\
\midrule
\endfirsthead
\multicolumn{5}{@{}l}{\small Table~\thetable{} (continued)}\\
\toprule
Control & Mean benefit & Paired SD & 95\% interval & W/T/L \\
\midrule
\endhead
\bottomrule
\endfoot
\multicolumn{5}{@{}l}{\textbf{50 nodes}} \\*
Frozen & $+0.00369$ & 0.02478 & $[-0.01923,\ +0.02661]$ & 3/0/4 \\
Native & $+0.01049$ & 0.02100 & $[-0.00893,\ +0.02991]$ & 5/0/2 \\
Validity-only & $+0.02158$ & 0.02724 & $[-0.00361,\ +0.04676]$ & 5/0/2 \\
Factorized & $+0.01638$ & 0.01893 & $[-0.00113,\ +0.03388]$ & 6/0/1 \\
Tail & $+0.01537$ & 0.01282 & $[+0.00352,\ +0.02722]$ & 6/0/1 \\
\addlinespace[6pt]
\multicolumn{5}{@{}l}{\textbf{100 nodes}} \\*
Frozen & $+0.00339$ & 0.06215 & $[-0.05409,\ +0.06087]$ & 4/0/3 \\
Native & $+0.01901$ & 0.05925 & $[-0.03578,\ +0.07381]$ & 4/0/3 \\
Validity-only & $+0.02113$ & 0.05838 & $[-0.03286,\ +0.07513]$ & 4/0/3 \\
Factorized & $+0.01052$ & 0.05075 & $[-0.03642,\ +0.05746]$ & 3/0/4 \\
Tail & $+0.02797$ & 0.04452 & $[-0.01320,\ +0.06915]$ & 4/0/3 \\
\addlinespace[6pt]
\multicolumn{5}{@{}l}{\textbf{200 nodes}} \\*
Frozen & $+0.01776$ & 0.13042 & $[-0.10286,\ +0.13838]$ & 4/0/3 \\
Native & $+0.04646$ & 0.12503 & $[-0.06918,\ +0.16209]$ & 5/0/2 \\
Validity-only & $+0.00792$ & 0.16168 & $[-0.14161,\ +0.15745]$ & 4/0/3 \\
Factorized & $+0.01321$ & 0.13778 & $[-0.11422,\ +0.14064]$ & 3/0/4 \\
Tail & $+0.05797$ & 0.11032 & $[-0.04407,\ +0.16000]$ & 5/0/2 \\
\end{longtable}
\endgroup
\par\vspace{6pt}
\noindent\begin{minipage}{\linewidth}\fontsize{10}{12}\selectfont\raggedright
Each point estimate averages seven training seeds after aggregating instances within each selected program. Descriptive intervals are $\bar d\pm t_{6,0.975}s_d/\sqrt{7}$. W/T/L counts SVC wins/ties/losses (tolerance $10^{-12}$). Scales and test instances do not add independent training repetitions.
\end{minipage}\par\medskip

\begingroup
\fontsize{10}{12}\selectfont
\setlength{\tabcolsep}{3.5pt}
\renewcommand{\arraystretch}{1.08}
\setlength{\LTleft}{0pt}\setlength{\LTright}{0pt}
\setlength{\LTpre}{8pt}\setlength{\LTpost}{2pt}
\begin{longtable}{@{\extracolsep{\fill}}lrrrc@{}}
\caption{CVRP selected-program paired comparisons. Benefit is the control minus SVC, in route cost; positive values favor SVC. Every setting includes all five controls.}\label{tab:svc-heldout-paired-cvrp_aco}\\
\toprule
Control & Mean benefit & Paired SD & 95\% interval & W/T/L \\
\midrule
\endfirsthead
\multicolumn{5}{@{}l}{\small Table~\thetable{} (continued)}\\
\toprule
Control & Mean benefit & Paired SD & 95\% interval & W/T/L \\
\midrule
\endhead
\bottomrule
\endfoot
\multicolumn{5}{@{}l}{\textbf{50 customers}} \\*
Frozen & $+0.13687$ & 0.39834 & $[-0.23153,\ +0.50527]$ & 5/0/2 \\
Native & $+0.16520$ & 0.22276 & $[-0.04082,\ +0.37122]$ & 5/0/2 \\
Validity-only & $+0.10520$ & 0.27667 & $[-0.15067,\ +0.36108]$ & 5/0/2 \\
Factorized & $+0.28033$ & 0.33579 & $[-0.03022,\ +0.59089]$ & 6/0/1 \\
Tail & $+0.08101$ & 0.24822 & $[-0.14855,\ +0.31057]$ & 4/0/3 \\
\addlinespace[6pt]
\multicolumn{5}{@{}l}{\textbf{100 customers}} \\*
Frozen & $+0.02251$ & 0.68605 & $[-0.61199,\ +0.65700]$ & 2/0/5 \\
Native & $+0.27201$ & 0.38856 & $[-0.08735,\ +0.63136]$ & 4/0/3 \\
Validity-only & $+0.14337$ & 0.16085 & $[-0.00539,\ +0.29214]$ & 6/0/1 \\
Factorized & $+0.45045$ & 0.96376 & $[-0.44088,\ +1.34179]$ & 6/0/1 \\
Tail & $+3.92773$ & 9.56455 & $[-4.91800,\ +12.77346]$ & 4/0/3 \\
\addlinespace[6pt]
\multicolumn{5}{@{}l}{\textbf{200 customers}} \\*
Frozen & $-0.09723$ & 1.15393 & $[-1.16444,\ +0.96997]$ & 2/0/5 \\
Native & $+0.52356$ & 0.87877 & $[-0.28917,\ +1.33628]$ & 4/0/3 \\
Validity-only & $+0.27902$ & 0.44584 & $[-0.13332,\ +0.69135]$ & 5/0/2 \\
Factorized & $+0.87933$ & 2.08436 & $[-1.04838,\ +2.80705]$ & 6/0/1 \\
Tail & $+5.10026$ & 11.61588 & $[-5.64264,\ +15.84316]$ & 4/0/3 \\
\end{longtable}
\endgroup
\par\vspace{6pt}
\noindent\begin{minipage}{\linewidth}\fontsize{10}{12}\selectfont\raggedright
Each point estimate averages seven training seeds after aggregating instances within each selected program. Descriptive intervals are $\bar d\pm t_{6,0.975}s_d/\sqrt{7}$. W/T/L counts SVC wins/ties/losses (tolerance $10^{-12}$). Scales and test instances do not add independent training repetitions.
\end{minipage}\par\medskip

\begingroup
\fontsize{10}{12}\selectfont
\setlength{\tabcolsep}{3.5pt}
\renewcommand{\arraystretch}{1.08}
\setlength{\LTleft}{0pt}\setlength{\LTright}{0pt}
\setlength{\LTpre}{8pt}\setlength{\LTpost}{2pt}
\begin{longtable}{@{\extracolsep{\fill}}lrrrc@{}}
\caption{OBP selected-program paired comparisons at bin capacity 100. Benefit is the control minus SVC, in percentage points; positive values favor SVC. Every setting includes all five controls.}\label{tab:svc-heldout-paired-obp-1}\\
\toprule
Control & Mean benefit & Paired SD & 95\% interval & W/T/L \\
\midrule
\endfirsthead
\multicolumn{5}{@{}l}{\small Table~\thetable{} (continued)}\\
\toprule
Control & Mean benefit & Paired SD & 95\% interval & W/T/L \\
\midrule
\endhead
\bottomrule
\endfoot
\multicolumn{5}{@{}l}{\textbf{100: 1k (capacity: items)}} \\*
Frozen & $+0.94433$ & 1.19077 & $[-0.15694,\ +2.04561]$ & 5/0/2 \\
Native & $+0.37631$ & 1.10276 & $[-0.64357,\ +1.39620]$ & 4/0/3 \\
Validity-only & $+0.46862$ & 0.40567 & $[+0.09344,\ +0.84380]$ & 6/0/1 \\
Factorized & $+0.80943$ & 1.24741 & $[-0.34423,\ +1.96309]$ & 5/0/2 \\
Tail & $+0.39051$ & 1.12588 & $[-0.65075,\ +1.43178]$ & 4/0/3 \\
\addlinespace[6pt]
\multicolumn{5}{@{}l}{\textbf{100: 5k (capacity: items)}} \\*
Frozen & $+1.11348$ & 1.32396 & $[-0.11097,\ +2.33794]$ & 6/0/1 \\
Native & $+0.70176$ & 0.83341 & $[-0.06901,\ +1.47254]$ & 7/0/0 \\
Validity-only & $+0.59424$ & 1.32534 & $[-0.63150,\ +1.81997]$ & 6/0/1 \\
Factorized & $+1.16583$ & 1.51538 & $[-0.23566,\ +2.56733]$ & 4/0/3 \\
Tail & $+0.03537$ & 0.47645 & $[-0.40527,\ +0.47601]$ & 4/0/3 \\
\addlinespace[6pt]
\multicolumn{5}{@{}l}{\textbf{100: 10k (capacity: items)}} \\*
Frozen & $+0.78577$ & 1.67462 & $[-0.76299,\ +2.33454]$ & 4/0/3 \\
Native & $+0.25219$ & 1.24913 & $[-0.90306,\ +1.40744]$ & 4/0/3 \\
Validity-only & $+0.28282$ & 1.05047 & $[-0.68870,\ +1.25435]$ & 5/0/2 \\
Factorized & $+0.90261$ & 1.98641 & $[-0.93451,\ +2.73973]$ & 4/0/3 \\
Tail & $-0.30704$ & 0.71836 & $[-0.97142,\ +0.35733]$ & 3/0/4 \\
\end{longtable}
\endgroup
\par\vspace{6pt}
\noindent\begin{minipage}{\linewidth}\fontsize{10}{12}\selectfont\raggedright
Each point estimate averages seven training seeds after aggregating instances within each selected program. Descriptive intervals are $\bar d\pm t_{6,0.975}s_d/\sqrt{7}$. W/T/L counts SVC wins/ties/losses (tolerance $10^{-12}$). Scales and test instances do not add independent training repetitions. OBP metrics aggregate five instances using the ratio of summed bin excess to summed volume lower bounds; differences use percentage points.
\end{minipage}\par\medskip

\begingroup
\fontsize{10}{12}\selectfont
\setlength{\tabcolsep}{3.5pt}
\renewcommand{\arraystretch}{1.08}
\setlength{\LTleft}{0pt}\setlength{\LTright}{0pt}
\setlength{\LTpre}{8pt}\setlength{\LTpost}{2pt}
\begin{longtable}{@{\extracolsep{\fill}}lrrrc@{}}
\caption{OBP selected-program paired comparisons at bin capacity 500. Benefit is the control minus SVC, in percentage points; positive values favor SVC. Every setting includes all five controls.}\label{tab:svc-heldout-paired-obp-2}\\
\toprule
Control & Mean benefit & Paired SD & 95\% interval & W/T/L \\
\midrule
\endfirsthead
\multicolumn{5}{@{}l}{\small Table~\thetable{} (continued)}\\
\toprule
Control & Mean benefit & Paired SD & 95\% interval & W/T/L \\
\midrule
\endhead
\bottomrule
\endfoot
\multicolumn{5}{@{}l}{\textbf{500: 1k (capacity: items)}} \\*
Frozen & $-0.14179$ & 0.85787 & $[-0.93519,\ +0.65160]$ & 2/2/3 \\
Native & $-0.07090$ & 0.53054 & $[-0.56157,\ +0.41977]$ & 2/3/2 \\
Validity-only & $+0.07090$ & 0.85616 & $[-0.72092,\ +0.86271]$ & 3/1/3 \\
Factorized & $+0.17724$ & 1.17016 & $[-0.90497,\ +1.25946]$ & 3/2/2 \\
Tail & $+0.38993$ & 1.15375 & $[-0.67711,\ +1.45698]$ & 4/1/2 \\
\addlinespace[6pt]
\multicolumn{5}{@{}l}{\textbf{500: 5k (capacity: items)}} \\*
Frozen & $+0.07810$ & 0.27352 & $[-0.17486,\ +0.33107]$ & 4/1/2 \\
Native & $+0.05680$ & 0.22307 & $[-0.14950,\ +0.26310]$ & 3/1/3 \\
Validity-only & $+0.07810$ & 0.29384 & $[-0.19365,\ +0.34986]$ & 4/1/2 \\
Factorized & $+0.16331$ & 0.20636 & $[-0.02754,\ +0.35415]$ & 4/1/2 \\
Tail & $+0.11360$ & 0.33674 & $[-0.19783,\ +0.42504]$ & 5/0/2 \\
\addlinespace[6pt]
\multicolumn{5}{@{}l}{\textbf{500: 10k (capacity: items)}} \\*
Frozen & $+0.02493$ & 0.23777 & $[-0.19498,\ +0.24483]$ & 4/0/3 \\
Native & $+0.01780$ & 0.15811 & $[-0.12842,\ +0.16403]$ & 3/1/3 \\
Validity-only & $+0.13531$ & 0.40316 & $[-0.23755,\ +0.50817]$ & 4/0/3 \\
Factorized & $+0.08546$ & 0.20547 & $[-0.10457,\ +0.27548]$ & 5/0/2 \\
Tail & $+0.04985$ & 0.31855 & $[-0.24476,\ +0.34446]$ & 4/0/3 \\
\end{longtable}
\endgroup
\par\vspace{6pt}
\noindent\begin{minipage}{\linewidth}\fontsize{10}{12}\selectfont\raggedright
Each point estimate averages seven training seeds after aggregating instances within each selected program. Descriptive intervals are $\bar d\pm t_{6,0.975}s_d/\sqrt{7}$. W/T/L counts SVC wins/ties/losses (tolerance $10^{-12}$). Scales and test instances do not add independent training repetitions. OBP metrics aggregate five instances using the ratio of summed bin excess to summed volume lower bounds; differences use percentage points.
\end{minipage}\par\medskip

\begingroup
\fontsize{10}{12}\selectfont
\setlength{\tabcolsep}{3.5pt}
\renewcommand{\arraystretch}{1.08}
\setlength{\LTleft}{0pt}\setlength{\LTright}{0pt}
\setlength{\LTpre}{8pt}\setlength{\LTpost}{2pt}
\begin{longtable}{@{\extracolsep{\fill}}lrrrc@{}}
\caption{OP selected-program paired comparisons. Benefit is SVC minus the control, in prize; positive values favor SVC. Every setting includes all five controls.}\label{tab:svc-heldout-paired-op_aco}\\
\toprule
Control & Mean benefit & Paired SD & 95\% interval & W/T/L \\
\midrule
\endfirsthead
\multicolumn{5}{@{}l}{\small Table~\thetable{} (continued)}\\
\toprule
Control & Mean benefit & Paired SD & 95\% interval & W/T/L \\
\midrule
\endhead
\bottomrule
\endfoot
\multicolumn{5}{@{}l}{\textbf{50 nodes}} \\*
Frozen & $-0.11360$ & 0.26630 & $[-0.35989,\ +0.13269]$ & 3/0/4 \\
Native & $-0.12824$ & 0.19457 & $[-0.30819,\ +0.05171]$ & 3/0/4 \\
Validity-only & $+0.01441$ & 0.23915 & $[-0.20676,\ +0.23559]$ & 4/0/3 \\
Factorized & $-0.02404$ & 0.24479 & $[-0.25043,\ +0.20236]$ & 5/0/2 \\
Tail & $+0.01263$ & 0.47075 & $[-0.42274,\ +0.44800]$ & 3/0/4 \\
\addlinespace[6pt]
\multicolumn{5}{@{}l}{\textbf{100 nodes}} \\*
Frozen & $-0.32002$ & 1.11789 & $[-1.35390,\ +0.71386]$ & 2/0/5 \\
Native & $-0.14928$ & 1.13171 & $[-1.19594,\ +0.89739]$ & 3/0/4 \\
Validity-only & $-0.00814$ & 1.11841 & $[-1.04250,\ +1.02622]$ & 4/0/3 \\
Factorized & $-0.07252$ & 1.03724 & $[-1.03181,\ +0.88677]$ & 4/0/3 \\
Tail & $-0.43424$ & 1.23188 & $[-1.57354,\ +0.70506]$ & 3/0/4 \\
\addlinespace[6pt]
\multicolumn{5}{@{}l}{\textbf{200 nodes}} \\*
Frozen & $-4.05011$ & 10.14280 & $[-13.43063,\ +5.33041]$ & 2/0/5 \\
Native & $-3.00872$ & 10.13347 & $[-12.38062,\ +6.36317]$ & 2/0/5 \\
Validity-only & $-3.66004$ & 10.63135 & $[-13.49240,\ +6.17232]$ & 3/0/4 \\
Factorized & $-3.43781$ & 10.96278 & $[-13.57669,\ +6.70107]$ & 4/0/3 \\
Tail & $-5.12639$ & 10.43966 & $[-14.78146,\ +4.52868]$ & 1/0/6 \\
\end{longtable}
\endgroup
\par\vspace{6pt}
\noindent\begin{minipage}{\linewidth}\fontsize{10}{12}\selectfont\raggedright
Each point estimate averages seven training seeds after aggregating instances within each selected program. Descriptive intervals are $\bar d\pm t_{6,0.975}s_d/\sqrt{7}$. W/T/L counts SVC wins/ties/losses (tolerance $10^{-12}$). Scales and test instances do not add independent training repetitions.
\end{minipage}\par\medskip

\FloatBarrier
\section{Generation and population-reset search}
\label{app:generation-reset}
\providecommand{\generationProtocolRef}{\ref{app:protocol}}

SVC uses actual admission and a short-term parent-exposure proxy to turn search feedback into generator training rewards. This appendix examines the resulting checkpoints under common generation contexts, compares reward variants, and evaluates checkpoint reuse after resetting the population. Together, these evaluations connect the reward design to generation behavior and subsequent search performance. Evaluation protocols and statistical definitions are given in \generationProtocolRef.

\subsection{Generation under common contexts}
\label{app:common-generation}

Under the same prompts and parent programs, SVC produces parent improvements more frequently than Native on both TSP and CVRP (Tables~\ref{tab:common-tsp} and~\ref{tab:common-cvrp_aco}). Mean parent-improvement yield is 8.71\% versus 2.87\% on TSP and 13.11\% versus 7.70\% on CVRP. The corresponding parent best-of-4 rates are 22.32\% versus 8.37\% and 34.60\% versus 20.31\%. Improvement yield counts successful improvements among all sampled responses, including invalid outputs in the denominator. SVC therefore produces more successful improvements per response budget despite its lower validity, which is 81.11\% versus 96.46\% on TSP and 79.91\% versus 95.09\% on CVRP.

SVC also produces a smaller proportion of responses whose fitness matches that of the best parent. Averaged across checkpoints, these ties account for 7.47\% of valid TSP responses and 19.02\% of valid CVRP responses, compared with 85.30\% and 61.24\% for Native. The difference persists when all responses form the denominator: 6.28\% versus 82.53\% on TSP and 15.51\% versus 58.82\% on CVRP. Thus, relative to Native, SVC combines higher mean parent-improvement yields with fewer parent-score ties. The tie rates describe fitness outcomes, rather than code identity or algorithmic diversity. Relative to Tail, SVC has lower mean parent and frontier improvement yields and best-of-4 rates on both tasks; the generation advantage is therefore specific to the comparison.

The complete tables report seven metrics. A response is valid when its extracted program passes execution checks and returns finite fitness. Parent improvement requires fitness strictly above that of the best parent in the prompt; frontier improvement requires fitness above the best value retained before generation in the search that supplied the prompt. Both yields divide improving responses by all responses, whereas each best-of-4 rate is the fraction of prompts with at least one corresponding improvement among four responses. A parent-score tie has an absolute fitness difference below $10^{-7}$ and is reported using both valid-response and all-response denominators.

Each checkpoint receives 128 TSP prompts or 64 CVRP prompts from the corresponding Frozen seed-42 search. Trained conditions report the mean and sample standard deviation of seven checkpoint-level rates; Frozen provides one reference per task, without a standard deviation or seven-pair interval. TSP is evaluated on RTX 3090 and CVRP on RTX 4090; Factorized is not evaluated on CVRP. For TSP Tail seeds 42, 3407, and 1926000, two complementary 64-prompt segments cover each prompt once, but the second segment reuses the first segment's sampling-seed range. Their sampling streams thus differ from uninterrupted evaluations. Segments and evaluations repeated on another host do not add training seeds.

\begingroup
\small
\setlength{\LTleft}{0pt}\setlength{\LTright}{0pt}
\setlength{\LTpre}{6pt}\setlength{\LTpost}{4pt}
\begin{longtable}{@{}lrrr@{}}
\caption{Complete TSP common-context metrics (percent). Each trained condition summarizes seven checkpoint seeds, each evaluated on 128 fixed prompts with four responses per prompt. Entries are mean $\pm$ sample SD; Frozen reports one reference checkpoint without an SD.}\label{tab:common-tsp}\\
\toprule

\endfirsthead
\multicolumn{4}{@{}l}{\small Table \thetable{} (continued)}\\
\toprule

\endhead
\bottomrule
\endfoot
\bottomrule
\endlastfoot
Condition & Valid responses & Parent improvement & Best-of-4\\*
\midrule
\noalign{\penalty10000}
SVC & $81.110\pm15.025$ & $8.705\pm3.571$ & $22.321\pm8.352$\\*
Tail & $89.955\pm4.202$ & $12.612\pm4.240$ & $31.362\pm7.279$\\*
Native & $96.456\pm2.123$ & $2.874\pm2.531$ & $8.371\pm6.851$\\*
Validity-only & $94.838\pm1.020$ & $7.840\pm1.341$ & $22.768\pm3.564$\\*
Factorized & $96.345\pm2.223$ & $2.372\pm1.005$ & $7.031\pm3.664$\\*
\midrule
\noalign{\penalty10000}
Frozen (reference) & $59.961$ & $4.492$ & $11.719$\\
\addlinespace[6pt]
Condition & Frontier breakthrough & Frontier best-of-4 & \\*
\midrule
\noalign{\penalty10000}
SVC & $1.283\pm0.780$ & $4.464\pm2.806$ & \\*
Tail & $2.427\pm1.421$ & $7.478\pm3.217$ & \\*
Native & $0.502\pm0.626$ & $1.562\pm1.914$ & \\*
Validity-only & $1.451\pm0.372$ & $4.576\pm1.590$ & \\*
Factorized & $0.363\pm0.307$ & $1.339\pm1.253$ & \\*
\midrule
\noalign{\penalty10000}
Frozen (reference) & $0.781$ & $2.344$ & \\
\addlinespace[6pt]
\pagebreak[4]
Condition & Ties / valid responses & Ties / all responses & \\*
\midrule
\noalign{\penalty10000}
SVC & $7.470\pm3.403$ & $6.278\pm3.348$ & \\*
Tail & $20.343\pm7.698$ & $18.276\pm6.733$ & \\*
Native & $85.304\pm15.419$ & $82.533\pm16.237$ & \\*
Validity-only & $31.891\pm12.209$ & $30.329\pm11.777$ & \\*
Factorized & $86.744\pm8.641$ & $83.705\pm9.875$ & \\*
\midrule
\noalign{\penalty10000}
Frozen (reference) & $24.430$ & $14.648$ & \\
\end{longtable}

\noindent{\footnotesize\raggedright Valid responses must pass execution checks and have finite fitness. Parent improvement is strict improvement over the best parent; frontier breakthrough is strict improvement over the prompt donor's best retained fitness before generation. The two yields divide by all responses. Best-of-4 metrics divide successful four-response groups by all prompts. Ties satisfy an absolute parent-fitness difference below $10^{-7}$; their two denominators are explicitly separated. A lower tie rate does not establish less copying or greater algorithm diversity. TSP and CVRP evaluations use their respective search instances and fixed context banks. The TSP Tail checkpoints for seeds 42, 3407 and 1926000 combine complementary halves with reused second-half sampling seeds. All condition summaries weight checkpoint seeds equally.\par}
\endgroup

\begingroup
\small
\setlength{\LTleft}{0pt}\setlength{\LTright}{0pt}
\setlength{\LTpre}{6pt}\setlength{\LTpost}{4pt}
\begin{longtable}{@{}lrrr@{}}
\caption{Complete CVRP common-context metrics (percent). Each trained condition summarizes seven checkpoint seeds, each evaluated on 64 fixed prompts with four responses per prompt. Entries are mean $\pm$ sample SD; Frozen reports one reference checkpoint without an SD. Dashes mark the unevaluated Factorized condition.}\label{tab:common-cvrp_aco}\\
\toprule

\endfirsthead
\multicolumn{4}{@{}l}{\small Table \thetable{} (continued)}\\
\toprule

\endhead
\bottomrule
\endfoot
\bottomrule
\endlastfoot
Condition & Valid responses & Parent improvement & Best-of-4\\*
\midrule
\noalign{\penalty10000}
SVC & $79.911\pm14.499$ & $13.114\pm0.979$ & $34.598\pm4.459$\\*
Tail & $90.681\pm6.280$ & $13.839\pm2.138$ & $37.277\pm5.806$\\*
Native & $95.089\pm3.204$ & $7.701\pm4.193$ & $20.312\pm10.166$\\*
Validity-only & $94.643\pm3.073$ & $9.431\pm3.743$ & $26.786\pm8.338$\\*
Factorized & -- & -- & --\\*
\midrule
\noalign{\penalty10000}
Frozen (reference) & $64.062$ & $5.859$ & $18.750$\\
\addlinespace[6pt]
Condition & Frontier breakthrough & Frontier best-of-4 & \\*
\midrule
\noalign{\penalty10000}
SVC & $3.237\pm1.001$ & $10.045\pm2.972$ & \\*
Tail & $3.571\pm0.857$ & $11.161\pm2.459$ & \\*
Native & $2.009\pm1.284$ & $6.473\pm3.427$ & \\*
Validity-only & $2.511\pm0.953$ & $8.036\pm2.459$ & \\*
Factorized & -- & -- & \\*
\midrule
\noalign{\penalty10000}
Frozen (reference) & $2.344$ & $6.250$ & \\
\addlinespace[6pt]
\pagebreak[4]
Condition & Ties / valid responses & Ties / all responses & \\*
\midrule
\noalign{\penalty10000}
SVC & $19.019\pm4.144$ & $15.513\pm5.316$ & \\*
Tail & $30.167\pm8.771$ & $27.734\pm8.870$ & \\*
Native & $61.239\pm29.299$ & $58.817\pm29.251$ & \\*
Validity-only & $46.619\pm14.096$ & $44.420\pm14.546$ & \\*
Factorized & -- & -- & \\*
\midrule
\noalign{\penalty10000}
Frozen (reference) & $37.805$ & $24.219$ & \\
\end{longtable}

\noindent{\footnotesize\raggedright Valid responses must pass execution checks and have finite fitness. Parent improvement is strict improvement over the best parent; frontier breakthrough is strict improvement over the prompt donor's best retained fitness before generation. The two yields divide by all responses. Best-of-4 metrics divide successful four-response groups by all prompts. Ties satisfy an absolute parent-fitness difference below $10^{-7}$; their two denominators are explicitly separated. A lower tie rate does not establish less copying or greater algorithm diversity. TSP and CVRP evaluations use their respective search instances and fixed context banks. The TSP Tail checkpoints for seeds 42, 3407 and 1926000 combine complementary halves with reused second-half sampling seeds. All condition summaries weight checkpoint seeds equally.\par}
\endgroup

\subsection{Generation with reward variants}
\label{app:variant-generation}

The internal TSP comparisons examine how the reward components relate to generation after training (Table~\ref{tab:internal-generation}). Admission-only replaces exposure weighting with a binary reward for actual admission; No-distance removes the reward's distance term while retaining distance-based parent selection; the added-penalty variant adds CALM's penalties for invalid responses. All checkpoints receive the same prompts and parent programs. Each comparison pairs seven training seeds on one evaluation host: RTX 3090 for added penalties and RTX 4090 for the other two. The RTX 4090 SVC evaluations reuse the same seven checkpoints, so the two hosts' SVC means are reported separately.

Relative to Admission-only, SVC has higher mean parent-improvement yield, frontier-improvement yield, and both best-of-4 rates. The parent-yield difference is $+0.64$ percentage points, with a descriptive 95\% paired interval of $[-3.82, 5.10]$. This direction is consistent with a generation benefit from weighting admitted responses by exposure, although the interval leaves the size and direction of the average effect uncertain. Against added penalties and No-distance, the parent-yield differences are $-2.73$ points $[-7.63, 2.16]$ and $-0.92$ points $[-4.63, 2.79]$, respectively; frontier yield and both best-of-4 means are also lower. All 15 paired intervals include zero.

These generation results complement the complete-search ablations, in which SVC has lower mean endpoint and budget-average costs than these three variants. These search advantages are not accompanied by higher means on every immediate generation metric. This distinction matters for SVC because its reward uses admission and a proxy for subsequent use as a parent, while the improvement metrics here measure immediate fitness gains. The comparisons remain exploratory: TSP results informed the reward design, and common evaluation contexts do not make the checkpoints' training histories identical.

\begingroup
\small
\setlength{\LTleft}{0pt}\setlength{\LTright}{0pt}
\setlength{\LTpre}{6pt}\setlength{\LTpost}{4pt}
\begin{longtable}{@{}lrrrr@{}}
\caption{Internal common-context generation on TSP. Each comparison uses seven matched checkpoint seeds evaluated on the same host, with 128 prompts and four responses per prompt. SVC and control columns are mean percentages; differences and descriptive 95\% paired intervals are in percentage points (SVC minus control).}\label{tab:internal-generation}\\
\toprule

\endfirsthead
\multicolumn{5}{@{}l}{\small Table \thetable{} (continued)}\\
\toprule

\endhead
\bottomrule
\endfoot
\bottomrule
\endlastfoot
Metric & SVC & Control & Difference & 95\% interval\\*
\midrule
\noalign{\penalty10000}
\multicolumn{5}{l}{\textit{Added penalty (RTX 3090)}}\\*
Validity & 81.11 & 89.31 & $-8.20$ & $[-21.64, +5.23]$\\*
Parent yield & 8.71 & 11.44 & $-2.73$ & $[-7.63, +2.16]$\\*
Parent best-of-4 & 22.32 & 29.69 & $-7.37$ & $[-18.55, +3.82]$\\*
Frontier yield & 1.28 & 2.40 & $-1.12$ & $[-2.54, +0.31]$\\*
Frontier best-of-4 & 4.46 & 7.03 & $-2.57$ & $[-5.62, +0.49]$\\
\addlinespace
\multicolumn{5}{l}{\textit{Admission-only (RTX 4090)}}\\*
Validity & 80.61 & 89.62 & $-9.01$ & $[-26.39, +8.37]$\\*
Parent yield & 8.93 & 8.29 & $+0.64$ & $[-3.82, +5.10]$\\*
Parent best-of-4 & 24.33 & 23.21 & $+1.12$ & $[-9.66, +11.89]$\\*
Frontier yield & 1.48 & 0.95 & $+0.53$ & $[-0.52, +1.58]$\\*
Frontier best-of-4 & 4.91 & 3.57 & $+1.34$ & $[-1.99, +4.67]$\\
\addlinespace
\multicolumn{5}{l}{\textit{No-distance (RTX 4090)}}\\*
Validity & 80.61 & 77.57 & $+3.04$ & $[-8.47, +14.55]$\\*
Parent yield & 8.93 & 9.85 & $-0.92$ & $[-4.63, +2.79]$\\*
Parent best-of-4 & 24.33 & 27.12 & $-2.79$ & $[-10.81, +5.23]$\\*
Frontier yield & 1.48 & 1.79 & $-0.31$ & $[-1.41, +0.80]$\\*
Frontier best-of-4 & 4.91 & 6.14 & $-1.23$ & $[-4.61, +2.16]$\\
\addlinespace
\end{longtable}

\noindent{\footnotesize\raggedright Validity and both yields divide by all 512 responses, including invalid responses. The two best-of-4 metrics divide successful response groups by all 128 prompts. Parent improvement requires fitness above that of the best parent; frontier improvement requires fitness above the best retained value before generation in the search that supplied the prompt. Intervals use $\bar d\pm t_{6,0.975}s_d/\sqrt{7}$, with sample SD $s_d$, assuming approximately normally distributed seed differences. All 15 intervals include zero. The RTX 4090 SVC evaluation repeats the same seven checkpoints and adds no training runs.\par}
\endgroup

\subsection{Reusing SVC checkpoints after population reset}

Population-reset search tests whether an SVC-trained generator can help a newly initialized search. Each of three final TSP checkpoints is loaded after resetting the retained population and prompt history, with further training disabled. The search generates 500 groups of four responses and is paired with Frozen using the same search seed. Figure~\ref{fig:reset-trajectories} of the main article shows the full trajectories, and Table~\ref{tab:reset-values} gives the paired costs and benefits. Endpoint cost is the best-so-far tour cost after group 500; budget-average cost averages this quantity over groups 1--500. Benefits are Frozen cost minus SVC-reset cost.

\begingroup
\small
\setlength{\LTleft}{0pt}\setlength{\LTright}{0pt}
\setlength{\LTpre}{6pt}\setlength{\LTpost}{4pt}
\begin{longtable}{@{}lrrr@{}}
\caption{Population-reset costs and paired benefits on TSP. Each row represents one matched search seed. Cost benefit is Frozen minus SVC reset, so positive values favor SVC reset. The budget average is the arithmetic mean of best-so-far tour cost over groups 1--500.}\label{tab:reset-values}\\
\toprule

\endfirsthead
\multicolumn{4}{@{}l}{\small Table \thetable{} (continued)}\\
\toprule

\endhead
\bottomrule
\endfoot
\bottomrule
\endlastfoot
Search seed & Frozen & SVC reset & Cost benefit\\*
\midrule
\noalign{\penalty10000}
\multicolumn{4}{l}{\textit{Endpoint}}\\*
42 & 6.213915 & 6.211530 & +0.002385\\*
3407 & 6.230332 & 6.171146 & +0.059186\\*
1926000 & 6.244498 & 6.203561 & +0.040937\\
\addlinespace
\multicolumn{4}{l}{\textit{Budget average}}\\*
42 & 6.234162 & 6.216481 & +0.017681\\*
3407 & 6.238371 & 6.188369 & +0.050002\\*
1926000 & 6.246967 & 6.213293 & +0.033674\\
\addlinespace

\addlinespace[6pt]
Paired benefit & Mean & Sample SD & Descriptive 95\% interval\\*
\midrule
\noalign{\penalty10000}
Endpoint & +0.034170 & 0.028999 & [-0.037868, +0.106207]\\*
Budget average & +0.033786 & 0.016161 & [-0.006359, +0.073931]\\
\end{longtable}

\noindent{\footnotesize\raggedright The paired summaries use three search seeds. Descriptive 95\% intervals are $\bar d\pm t_{2,0.975}s_d/\sqrt{3}$, where $s_d$ is the sample SD. Both intervals include zero. All three paired differences are positive for both metrics, with no ties at tolerance $10^{-12}$. No group 0, time weighting or trapezoidal integration is used. The first-group adapter delay described in \ref{app:protocol-generation} also applies; the evidence concerns three population-reset searches with further training disabled.\par}
\endgroup

All three SVC-reset searches achieve lower endpoint and budget-average costs than Frozen. Endpoint reductions are 0.002385, 0.059186, and 0.040937 for seeds 42, 3407, and 1926000, respectively. These gains show the practical value of reusing the trained checkpoints in the observed searches: the generator helps obtain lower costs without carrying over the earlier population or receiving further training. This supports the use of SVC to turn search experience into reusable generator parameters. The mean endpoint reduction is 0.034170, with a descriptive 95\% paired interval of $[-0.037868, 0.106207]$; the mean budget-average reduction is 0.033786 with interval $[-0.006359, 0.073931]$. Both intervals include zero, so the three pairs leave uncertainty about the average benefit across checkpoints.

The loaded adapter took effect after the first response group. This group matches Frozen, whereas groups two and three differ. Recorded optimizer steps, backward steps, and training tokens are zero, and adapter parameter checksums remain unchanged and match the source checkpoint. As specified in \generationProtocolRef, the statistical units are the three seed pairs.

\FloatBarrier
\section{Supplementary credit analysis and ablation details}
\label{app:credit-analysis}

This appendix supplements the main-text credit analysis with ablation details, admission events for all conditions, and reward replay on recorded responses. SVC implementation details are given in \ref{app:svc-implementation}; replay protocols and statistical conventions are given in \ref{app:protocol}. The search and checkpoint evaluations are reported in \ref{app:results} and \ref{app:generation-reset}, respectively.

\subsection{Supplementary ablation details}
\label{app:ablation-details}

Table~\ref{tab:svc-ablation-supplementary} supplies paired differences for the ablations in the main text. Binary admission assigns reward one to every valid admitted response, including a response whose exposure is zero.

For Rank+penalty, valid admitted responses with exposure greater than $10^{-12}$ receive $1+\mathrm{rank}/\max(|\mathcal I_t|-1,1)$, using zero-based within-group midranks over the qualifying response-index set $\mathcal I_t$. Exposure ties use tolerance $10^{-12}$; a single qualifying response receives reward one. Other valid responses receive zero, and invalid responses receive the native CALM negative failure reward, or $-1$ if unavailable.

Paired benefit is defined as comparator cost minus SVC cost for the same seed. Relative benefits are computed by dividing each paired benefit by the corresponding comparator cost before averaging across seeds. The mean relative endpoint reductions are 0.307\% against Added-penalty and 0.387\% against Admission-only, with descriptive 95\% paired intervals of [0.010\%, 0.604\%] and [0.017\%, 0.757\%], respectively. Against No-distance, the mean relative reduction is 0.189\%, and its paired interval includes zero. These relative reductions correspond to the main-text percentages; Table~\ref{tab:svc-ablation-supplementary} reports absolute cost differences. Common-context evaluations of the trained variants are reported in \ref{app:variant-generation}.

\begingroup
\begin{table}[!htbp]\centering
\caption{Supplementary two-factor variant and paired TSP comparisons. SVC has lower mean endpoint and budget-averaged cost than every variant.}
\label{tab:svc-ablation-supplementary}
\begingroup\fontsize{10}{12}\selectfont
\setlength{\tabcolsep}{4pt}\renewcommand{\arraystretch}{1.15}
\begin{tabular*}{\linewidth}{@{\extracolsep{\fill}}lrr@{}}
\toprule
Condition & Endpoint $\downarrow$ & Budget mean $\downarrow$ \\
\midrule
SVC & $6.19345\pm0.02295$ & $6.21416\pm0.02075$ \\
Rank + penalty & $6.21089\pm0.02037$ & $6.23233\pm0.02507$ \\
\bottomrule
\end{tabular*}
\endgroup
\par\vspace{12pt}
{\small\bfseries Paired cost reduction: variant minus SVC\par}\vspace{4pt}
\begingroup\fontsize{10}{12}\selectfont
\setlength{\tabcolsep}{4pt}\renewcommand{\arraystretch}{1.15}
\begin{tabular*}{\linewidth}{@{\extracolsep{\fill}}lrrc@{}}
\toprule
Variant & Mean benefit & Descriptive 95\% interval & W/T/L \\
\midrule
\multicolumn{4}{@{}l}{\textit{Endpoint}} \\[2pt]
+ Failure penalty & $+0.01906$ & $[+0.00059,\ +0.03754]$ & 6/0/1 \\
Binary admission & $+0.02407$ & $[+0.00103,\ +0.04711]$ & 6/0/1 \\
No reward distance & $+0.01177$ & $[-0.01213,\ +0.03567]$ & 5/0/2 \\
Rank + penalty & $+0.01744$ & $[-0.01494,\ +0.04982]$ & 5/0/2 \\
\addlinespace[5pt]
\multicolumn{4}{@{}l}{\textit{Budget mean}} \\[2pt]
+ Failure penalty & $+0.01631$ & $[-0.00364,\ +0.03626]$ & 5/0/2 \\
Binary admission & $+0.01672$ & $[-0.00425,\ +0.03770]$ & 6/0/1 \\
No reward distance & $+0.00877$ & $[-0.01616,\ +0.03370]$ & 5/0/2 \\
Rank + penalty & $+0.01817$ & $[-0.01843,\ +0.05477]$ & 5/0/2 \\
\addlinespace[5pt]
\bottomrule
\end{tabular*}
\endgroup
\par\vspace{6pt}\begin{minipage}{\linewidth}
\fontsize{10}{12}\selectfont\raggedright
Rank + penalty replaces positive admitted exposure weights with within-group rank weights and adds the failure penalty: two factors change relative to SVC. Absolute costs are means $\pm$ sample SD over seven seeds. Paired benefit is variant cost minus SVC cost for the same seed, so positive values favor SVC. Intervals are descriptive 95\% Student-$t$ intervals with six degrees of freedom. W/T/L counts positive/zero/negative benefits, with tolerance $10^{-12}$. An interval containing zero does not establish equivalence. The results do not isolate effects on model parameters or support a causal decomposition of the two-factor variant.
\end{minipage}\end{table}

\par\endgroup

\subsection{Selected-program tests for the TSP ablations}
\label{app:ablation-program-tests}

The programs selected by the SVC searches and the four reward variants are evaluated on separate TSP test sets with 50, 100, and 200 nodes, using 1,000 instances per size.
Each of the seven search seeds contributes its valid admitted program with the highest training fitness; test performance is not used for reselection.
The 35 selected programs yield 105 program--size cells, each with 1,000 finite valid instance results.
Tour costs are first averaged within each program and test size, then summarized across the seven matched search seeds.

Table~\ref{tab:ablation-program-means} reports all five conditions at all three sizes.
SVC has lower mean costs than the added-penalty, Admission-only, and No-distance variants at every size.
Table~\ref{tab:ablation-program-paired} gives the corresponding paired cost reductions and seed-wise relative reductions averaged over the three sizes.
The mean relative cost reduction, averaged across sizes within each seed, is positive against all four variants, although all four paired intervals include zero. Rank+penalty has lower mean costs than SVC at 100 and 200 nodes.

\begin{table}[!htbp]
\centering\small
\caption{TSP selected-program tests for the reward ablations. Costs are mean $\pm$ sample SD over seven selected programs, one per search seed; lower is better. Each program is evaluated on 1,000 test instances per size.}
\label{tab:ablation-program-means}
\begin{tabular*}{\linewidth}{@{\extracolsep{\fill}}lccc@{}}
\toprule
Condition & 50 nodes & 100 nodes & 200 nodes \\
\midrule
SVC & $\mathbf{6.24393}\pm0.01510$ & $8.70728\pm0.04047$ & $12.20907\pm0.10220$ \\
Added-penalty & $6.25866\pm0.01312$ & $8.72603\pm0.03222$ & $12.24813\pm0.08257$ \\
Admission-only & $6.26320\pm0.01575$ & $8.72431\pm0.01708$ & $12.21985\pm0.07260$ \\
No-distance & $6.25297\pm0.01098$ & $8.72708\pm0.01943$ & $12.25846\pm0.05178$ \\
Rank+penalty & $6.25453\pm0.01409$ & $\mathbf{8.70330}\pm0.03407$ & $\mathbf{12.19817}\pm0.16786$ \\
\bottomrule
\end{tabular*}
\par\medskip
\begin{minipage}{\linewidth}\footnotesize\raggedright
Bold marks the lowest observed mean among these five conditions at each size, without denoting significance.
Searches use 50-node instances. No-distance removes only the reward's distance contribution; distance-based parent selection is retained.
Rank+penalty changes both the exposure weighting and the failure penalty.
The instances and test sizes do not increase the number of independent search seeds.
\end{minipage}
\end{table}

\FloatBarrier
\begingroup\small
\setlength{\LTleft}{0pt}\setlength{\LTright}{0pt}
\setlength{\LTpre}{8pt}\setlength{\LTpost}{6pt}
\begin{longtable}{@{\extracolsep{\fill}}llrcc@{}}
\caption{Paired benefits in the TSP ablation program tests. Positive benefits favor SVC. Each comparison uses seven matched search seeds.}\label{tab:ablation-program-paired}\\
\toprule
Test size & Comparator & Mean benefit & 95\% interval & W/T/L \\
\midrule
\endfirsthead
\multicolumn{5}{l}{\small Table \thetable{} (continued)}\\
\toprule
Test size & Comparator & Mean benefit & 95\% interval & W/T/L \\
\midrule
\endhead
\bottomrule\endfoot
\multicolumn{5}{l}{\textit{Absolute tour-cost reduction: comparator minus SVC}}\\[3pt]
50 & Added-penalty & $+0.01473$ & $[+0.00480,\,+0.02465]$ & 7/0/0 \\
50 & Admission-only & $+0.01928$ & $[-0.00352,\,+0.04208]$ & 6/0/1 \\
50 & No-distance & $+0.00904$ & $[-0.00812,\,+0.02620]$ & 6/0/1 \\
50 & Rank+penalty & $+0.01060$ & $[-0.00697,\,+0.02818]$ & 4/0/3 \\
\addlinespace[3pt]
100 & Added-penalty & $+0.01875$ & $[-0.00611,\,+0.04362]$ & 5/0/2 \\
100 & Admission-only & $+0.01704$ & $[-0.02194,\,+0.05602]$ & 5/0/2 \\
100 & No-distance & $+0.01980$ & $[-0.02791,\,+0.06751]$ & 4/0/3 \\
100 & Rank+penalty & $-0.00397$ & $[-0.05287,\,+0.04492]$ & 3/0/4 \\
\addlinespace[3pt]
200 & Added-penalty & $+0.03906$ & $[-0.03996,\,+0.11808]$ & 5/0/2 \\
200 & Admission-only & $+0.01078$ & $[-0.09818,\,+0.11974]$ & 5/0/2 \\
200 & No-distance & $+0.04939$ & $[-0.06889,\,+0.16768]$ & 4/0/3 \\
200 & Rank+penalty & $-0.01090$ & $[-0.20685,\,+0.18505]$ & 2/0/5 \\
\addlinespace[3pt]
\multicolumn{5}{l}{\textit{Relative reduction averaged across sizes within each seed (\%)}}\\*[3pt]
50/100/200 & Added-penalty & $+0.256$ & $[-0.072,\,+0.584]$ & 6/0/1 \\
50/100/200 & Admission-only & $+0.196$ & $[-0.316,\,+0.708]$ & 5/0/2 \\
50/100/200 & No-distance & $+0.257$ & $[-0.325,\,+0.839]$ & 4/0/3 \\
50/100/200 & Rank+penalty & $+0.005$ & $[-0.755,\,+0.765]$ & 3/0/4 \\
\end{longtable}
\endgroup
\noindent\begin{minipage}{\linewidth}\footnotesize\raggedright
For each test size, benefit is comparator cost minus SVC cost for the same search seed.
The pooled relative benefit first divides each cost reduction by its corresponding comparator cost and multiplies by 100, then averages the three test sizes within each seed before summarizing the seven seeds.
Intervals are descriptive 95\% Student-$t$ intervals with six degrees of freedom.
All four pooled intervals include zero.
W/T/L counts positive/zero/negative paired benefits with tolerance $10^{-12}$.
\end{minipage}

\subsection{Admission-event summary}
\label{app:admission-events}

\begin{figure}[!htbp]
\centering
\includegraphics[width=\linewidth]{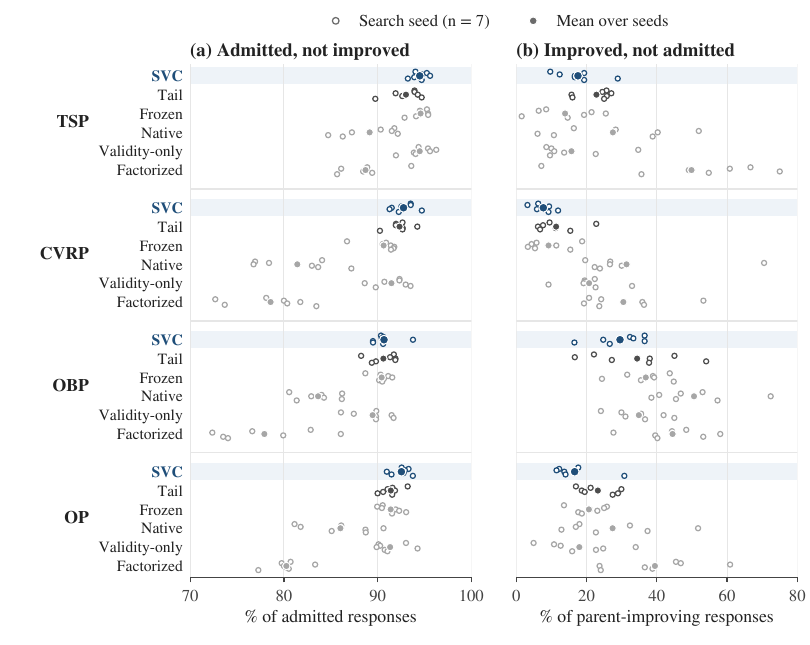}
\caption{%
Divergence between admission and parent improvement on each condition's own search trajectories. (a) Share of admitted responses that did not improve on the best parent in their prompt. (b) Share of parent-improving responses that were not admitted. Open circles are the seven search seeds; filled markers are seed means. Responses are counted when valid, evaluated, with a finite parent comparator and outside initialization; improvement requires fitness strictly above the best parent. Admission is the logged response-hash match to a newly retained program, so repeated responses can share one admission. These proportions describe each condition's own trajectories and do not establish a causal effect of the reward mapping.
}
\label{fig:app-admission-all}
\end{figure}

Figure~\ref{fig:app-admission-all} extends the main-text comparison to all six conditions. On CVRP, OBP and OP, SVC has both the highest share of admitted responses that do not improve on the best parent and the lowest share of improving responses that are not admitted among the six conditions. These descriptive proportions do not isolate a causal effect of the reward mapping. Eligibility follows \ref{app:protocol}: valid, non-initialization responses with a finite parent comparator. Counts refer to responses, including duplicates that share admission flags.

\subsection{Empirical four-response example}

\begin{figure}[!htbp]
\centering
\includegraphics[width=\linewidth]{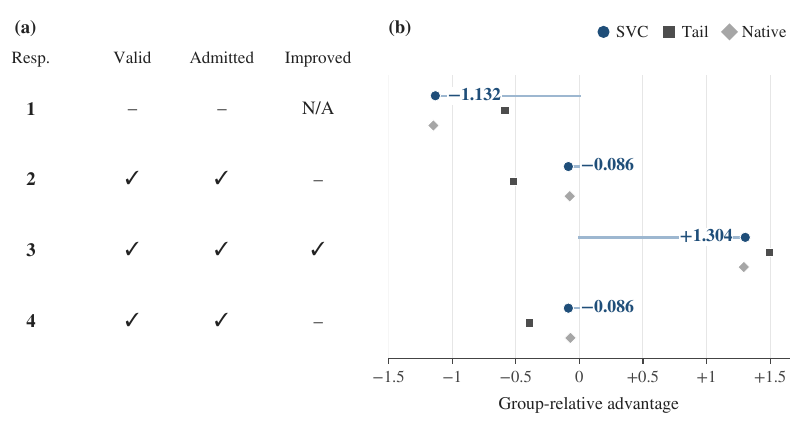}
\caption{%
Group-relative advantages for the same four recorded responses from TSP SVC training (Qwen2.5-7B-Instruct, seed 42, group $t=2$). (a) Recorded response attributes (ticks: yes; dashes: no); improvement requires the candidate's fitness to strictly exceed the best matched-parent fitness, and N/A denotes an invalid response. (b) SVC assigns equal advantages to admitted non-improvers 2 and 4 and its lowest advantage to invalid response 1; Tail separates responses 2 and 4, while Native remains close to SVC. Each mapping uses all four responses: $\widehat A_{t,i}=(r_{t,i}-\bar r_t)/(s_t+10^{-4})$, with sample SD $s_t$ (denominator 3). This is the first chronological complete group containing an admitted valid non-improver and a valid improver, both with finite parent comparators (71/500 groups meet the rule). Raw reward scales do not measure update strength; this fixed-record example does not establish overall benefit.
}
\label{fig:app-shared-response}
\end{figure}

Figure~\ref{fig:app-shared-response} shows an empirical response group, complementing the constructed example in \ref{app:svc-implementation}. It is group 2 of the TSP SVC run with seed 42: the earliest complete four-response group in that run containing both a valid admitted response that does not improve on its parent comparator and a valid response that does. Both responses must have finite parent comparators. The rule was fixed for figure construction, without experimental preregistration. In this run, 71 of 500 groups (14.2\%) meet the rule. This frequency describes one trajectory; the 500 groups are not independent runs.

All four responses enter each normalization, including the invalid response. Rewards are centered by their group mean and divided by their sample standard deviation plus $10^{-4}$; the sample-variance denominator is three. Responses 2 and 4 have positive SVC rewards but negative group-relative advantages because their rewards fall below the group mean. Native and SVC advantages differ modestly in this example. The raw reward scales of the mappings cannot be compared as measures of update strength, and the example does not estimate a search benefit.

\subsection{Labels and statistics for fixed-record replay}
\label{app:replay}

Eight mappings are applied to the same 28 SVC trajectories: Native, Validity-only, Factorized, Tail-weighted, SVC, added penalty, Admission-only, and No-distance. SVC follows the method and \ref{app:svc-implementation}, using the recorded admission flags and exposure values. The three SVC variants correspond to the main-text ablations. Rank+penalty is a separate TSP training variant and is not among the eight replay mappings. As specified in \ref{app:protocol}, replay holds the responses and search history fixed while replacing the reward mapping; it does not retrain the model or generate a new trajectory.

Later parent use counts occurrences of a response's code hash (SHA-1) in the parent-code lists of strictly later groups, including repeated slots. For ancestor labels, each code hash is linked to the parents recorded in its first group with an admission flag. Targets must be valid, evaluated responses with finite fitness strictly above the finite best retained fitness before generation. Starting from each target group's parents, these links are traced backward. Responses carrying a reached hash in that code's first admitted group receive the ancestor label only if that group strictly precedes the target group. Multiple responses can share a label, and several targets may occur in one group. Code hashes identify exact code, not semantic uniqueness.

AUC measures how well the group-relative advantage ranks ancestor-labeled responses above unlabeled responses among valid admitted responses with a code hash. Tied advantages receive half credit; AUC is undefined if either label class is absent. For a binary label $\lambda_{t,i}$, the positive-advantage share is

\begin{equation}
P_\lambda=\frac{\sum_{t,i} \lambda_{t,i}\max(\widehat A_{t,i},0)}{\sum_{t,i}\max(\widehat A_{t,i},0)}.\label{eq:positive-advantage-share}
\end{equation}

The sums cover all groups $t$ and all response indices $i$ in one trajectory. $P_\lambda$ is undefined when its denominator is zero. $P_U$ uses the label that code is used at least once as a later parent, and $P_D$ uses the ancestor label. These quantities are shares of positive advantage magnitude, rather than percentages of responses or actual gradient contributions. Spearman correlation relates advantage to later-use count among valid responses; it is undefined with fewer than three responses or when either ranked variable is constant. Undefined metric values are left missing and excluded from the corresponding summaries; a paired contrast requires both values to be defined. All reported entries are defined, giving seven trajectories per task and 21 in the non-TSP collection. A zero-task-advantage group satisfies $\max_i|\widehat A_{t,i}|\leq10^{-10}$; its fraction $P_0$ uses all 500 groups. Constant task rewards do not remove the separate KL term or imply a zero parameter update.

Each metric is calculated per trajectory before averaging. Tables~\ref{tab:credit-replay-tsp}--\ref{tab:credit-replay-non_tsp} provide all eight mappings for each task and the 21-trajectory non-TSP collection. Tables~\ref{tab:credit-paired-tsp} and~\ref{tab:credit-paired-non_tsp} give paired contrasts for the three metrics emphasized in the main text. Intervals are descriptive Student-$t$ intervals. The non-TSP collection contains three fixed tasks with seven trajectories each and does not support inference to a population of tasks.

Code first admitted earlier has a longer remaining observation window. The labels are not adjusted for this difference, so the absence of recorded later use for code admitted near the end of a run can reflect limited follow-up. Without candidate-removal interventions, replay describes signal allocation and associations with subsequent recorded events, not causal contributions or long-term returns. SVC's lower zero-task-advantage rate than Admission-only on every trajectory also does not imply nonzero feedback in every group where Admission-only provides it: CVRP seed 42 has 17 groups with nonzero task advantages under Admission-only but zero task advantages under SVC.

\begingroup
\fontsize{9}{11}\selectfont
\setlength{\tabcolsep}{3pt}
\renewcommand{\arraystretch}{1.0}
\setlength{\LTleft}{0pt}\setlength{\LTright}{0pt}
\setlength{\LTpre}{8pt}\setlength{\LTpost}{2pt}
\begin{longtable}{@{\extracolsep{\fill}}lrrrrr@{}}
\caption{Eight-mapping replay on TSP SVC trajectories.}\label{tab:credit-replay-tsp}\\
\toprule
Mapping & AUC & $P_U$ (\%) & $P_D$ (\%) & $\rho$ & $P_0$ (\%) \\
\midrule
\endfirsthead
\multicolumn{6}{@{}l}{\small Table~\thetable{} (continued)}\\
\toprule
Mapping & AUC & $P_U$ (\%) & $P_D$ (\%) & $\rho$ & $P_0$ (\%) \\
\midrule
\endhead
\bottomrule
\endfoot
Native & $0.8132\pm0.0589$ & $26.82\pm1.89$ & $4.40\pm1.99$ & $0.2087\pm0.0246$ & $5.86\pm4.24$ \\
Validity-only & $0.5129\pm0.0661$ & $21.14\pm2.20$ & $2.71\pm2.29$ & $0.0214\pm0.0256$ & $72.34\pm5.84$ \\
Factorized & $0.7648\pm0.0546$ & $26.10\pm1.85$ & $3.64\pm1.62$ & $0.2000\pm0.0190$ & $5.49\pm4.36$ \\
Tail & $0.7925\pm0.0408$ & $33.15\pm1.93$ & $5.25\pm2.35$ & $0.2385\pm0.0282$ & $4.83\pm4.08$ \\
\textbf{SVC} & $\mathbf{0.8206\pm0.0373}$ & $\mathbf{35.65\pm2.17}$ & $\mathbf{5.46\pm2.68}$ & $\mathbf{0.2976\pm0.0339}$ & $\mathbf{7.43\pm4.34}$ \\
+ penalty & $0.8132\pm0.0429$ & $32.31\pm1.66$ & $4.87\pm2.18$ & $0.2780\pm0.0287$ & $5.51\pm3.99$ \\
Admission-only & $0.5519\pm0.0756$ & $23.19\pm2.14$ & $2.96\pm1.93$ & $0.0890\pm0.0373$ & $46.57\pm6.08$ \\
No-distance & $0.8607\pm0.0419$ & $54.44\pm3.65$ & $9.22\pm3.99$ & $0.3319\pm0.0318$ & $50.17\pm3.32$ \\
\end{longtable}
\endgroup
\par\medskip\noindent\begin{minipage}{\linewidth}\fontsize{9}{11}\selectfont\raggedright
Means $\pm$ sample SD over 7 trajectories. AUC: ranking of ancestor-labeled above unlabeled responses by advantage, among valid admitted responses with code hashes. $P_U$ and $P_D$: shares of positive advantage assigned to later-used and ancestor-labeled responses, respectively (percent). $\rho$: Spearman correlation between advantage and later-use count among valid responses. $P_0$: zero-task-advantage groups (percent of 500 groups). All eight mappings use the same SVC records. Bold identifies SVC, not column-wise best values.
\end{minipage}\par\smallskip

\begingroup
\fontsize{9}{11}\selectfont
\setlength{\tabcolsep}{3pt}
\renewcommand{\arraystretch}{1.0}
\setlength{\LTleft}{0pt}\setlength{\LTright}{0pt}
\setlength{\LTpre}{8pt}\setlength{\LTpost}{2pt}
\begin{longtable}{@{\extracolsep{\fill}}lrrrrr@{}}
\caption{Eight-mapping replay on CVRP SVC trajectories.}\label{tab:credit-replay-cvrp_aco}\\
\toprule
Mapping & AUC & $P_U$ (\%) & $P_D$ (\%) & $\rho$ & $P_0$ (\%) \\
\midrule
\endfirsthead
\multicolumn{6}{@{}l}{\small Table~\thetable{} (continued)}\\
\toprule
Mapping & AUC & $P_U$ (\%) & $P_D$ (\%) & $\rho$ & $P_0$ (\%) \\
\midrule
\endhead
\bottomrule
\endfoot
Native & $0.7839\pm0.0437$ & $27.54\pm1.41$ & $4.99\pm1.72$ & $0.2302\pm0.0119$ & $4.97\pm2.50$ \\
Validity-only & $0.5626\pm0.0394$ & $21.96\pm2.02$ & $3.43\pm0.99$ & $0.0364\pm0.0258$ & $64.23\pm3.78$ \\
Factorized & $0.7360\pm0.0287$ & $27.37\pm1.33$ & $4.30\pm1.56$ & $0.2170\pm0.0114$ & $5.89\pm2.58$ \\
Tail & $0.7905\pm0.0330$ & $34.15\pm2.01$ & $5.70\pm1.86$ & $0.2549\pm0.0175$ & $4.54\pm2.51$ \\
\textbf{SVC} & $\mathbf{0.7980\pm0.0393}$ & $\mathbf{36.83\pm1.75}$ & $\mathbf{6.07\pm1.93}$ & $\mathbf{0.3230\pm0.0255}$ & $\mathbf{9.09\pm2.88}$ \\
+ penalty & $0.7803\pm0.0347$ & $31.54\pm1.40$ & $5.18\pm1.70$ & $0.2919\pm0.0229$ & $5.80\pm2.39$ \\
Admission-only & $0.6448\pm0.0594$ & $25.63\pm1.52$ & $4.51\pm1.12$ & $0.1401\pm0.0114$ & $41.00\pm3.89$ \\
No-distance & $0.8267\pm0.0256$ & $54.91\pm2.69$ & $9.96\pm3.20$ & $0.3640\pm0.0294$ & $48.26\pm2.72$ \\
\end{longtable}
\endgroup
\par\medskip\noindent\begin{minipage}{\linewidth}\fontsize{9}{11}\selectfont\raggedright
Means $\pm$ sample SD over 7 trajectories. AUC: ranking of ancestor-labeled above unlabeled responses by advantage, among valid admitted responses with code hashes. $P_U$ and $P_D$: shares of positive advantage assigned to later-used and ancestor-labeled responses, respectively (percent). $\rho$: Spearman correlation between advantage and later-use count among valid responses. $P_0$: zero-task-advantage groups (percent of 500 groups). All eight mappings use the same SVC records. Bold identifies SVC, not column-wise best values.
\end{minipage}\par\smallskip

\begingroup
\fontsize{9}{11}\selectfont
\setlength{\tabcolsep}{3pt}
\renewcommand{\arraystretch}{1.0}
\setlength{\LTleft}{0pt}\setlength{\LTright}{0pt}
\setlength{\LTpre}{8pt}\setlength{\LTpost}{2pt}
\begin{longtable}{@{\extracolsep{\fill}}lrrrrr@{}}
\caption{Eight-mapping replay on OBP SVC trajectories.}\label{tab:credit-replay-obp}\\
\toprule
Mapping & AUC & $P_U$ (\%) & $P_D$ (\%) & $\rho$ & $P_0$ (\%) \\
\midrule
\endfirsthead
\multicolumn{6}{@{}l}{\small Table~\thetable{} (continued)}\\
\toprule
Mapping & AUC & $P_U$ (\%) & $P_D$ (\%) & $\rho$ & $P_0$ (\%) \\
\midrule
\endhead
\bottomrule
\endfoot
Native & $0.7549\pm0.0520$ & $26.86\pm1.25$ & $5.47\pm1.99$ & $0.1954\pm0.0278$ & $13.63\pm6.68$ \\
Validity-only & $0.5025\pm0.0365$ & $20.59\pm2.44$ & $3.08\pm1.20$ & $0.0329\pm0.0289$ & $71.11\pm3.45$ \\
Factorized & $0.7355\pm0.0336$ & $23.35\pm1.20$ & $4.46\pm1.81$ & $0.1599\pm0.0324$ & $7.43\pm3.54$ \\
Tail & $0.7456\pm0.0413$ & $33.08\pm1.29$ & $6.52\pm2.20$ & $0.2681\pm0.0176$ & $6.89\pm3.32$ \\
\textbf{SVC} & $\mathbf{0.7442\pm0.0612}$ & $\mathbf{37.72\pm2.53}$ & $\mathbf{7.24\pm3.15}$ & $\mathbf{0.3412\pm0.0310}$ & $\mathbf{16.00\pm4.24}$ \\
+ penalty & $0.7404\pm0.0666$ & $32.59\pm2.18$ & $6.32\pm2.86$ & $0.3132\pm0.0361$ & $10.77\pm3.56$ \\
Admission-only & $0.5440\pm0.0869$ & $27.31\pm2.31$ & $4.78\pm2.29$ & $0.2322\pm0.0423$ & $32.49\pm5.04$ \\
No-distance & $0.8045\pm0.0686$ & $59.79\pm4.77$ & $12.72\pm5.88$ & $0.3760\pm0.0368$ & $55.34\pm3.00$ \\
\end{longtable}
\endgroup
\par\medskip\noindent\begin{minipage}{\linewidth}\fontsize{9}{11}\selectfont\raggedright
Means $\pm$ sample SD over 7 trajectories. AUC: ranking of ancestor-labeled above unlabeled responses by advantage, among valid admitted responses with code hashes. $P_U$ and $P_D$: shares of positive advantage assigned to later-used and ancestor-labeled responses, respectively (percent). $\rho$: Spearman correlation between advantage and later-use count among valid responses. $P_0$: zero-task-advantage groups (percent of 500 groups). All eight mappings use the same SVC records. Bold identifies SVC, not column-wise best values.
\end{minipage}\par\smallskip

\begingroup
\fontsize{9}{11}\selectfont
\setlength{\tabcolsep}{3pt}
\renewcommand{\arraystretch}{1.0}
\setlength{\LTleft}{0pt}\setlength{\LTright}{0pt}
\setlength{\LTpre}{8pt}\setlength{\LTpost}{2pt}
\begin{longtable}{@{\extracolsep{\fill}}lrrrrr@{}}
\caption{Eight-mapping replay on OP SVC trajectories.}\label{tab:credit-replay-op_aco}\\
\toprule
Mapping & AUC & $P_U$ (\%) & $P_D$ (\%) & $\rho$ & $P_0$ (\%) \\
\midrule
\endfirsthead
\multicolumn{6}{@{}l}{\small Table~\thetable{} (continued)}\\
\toprule
Mapping & AUC & $P_U$ (\%) & $P_D$ (\%) & $\rho$ & $P_0$ (\%) \\
\midrule
\endhead
\bottomrule
\endfoot
Native & $0.7536\pm0.0412$ & $28.18\pm1.81$ & $4.30\pm1.81$ & $0.2435\pm0.0340$ & $8.54\pm3.91$ \\
Validity-only & $0.5623\pm0.0433$ & $23.11\pm3.74$ & $3.72\pm1.92$ & $0.0458\pm0.0456$ & $71.71\pm5.51$ \\
Factorized & $0.7464\pm0.0505$ & $27.66\pm1.91$ & $4.07\pm1.69$ & $0.2225\pm0.0390$ & $10.00\pm4.67$ \\
Tail & $0.7211\pm0.0440$ & $34.21\pm2.52$ & $5.16\pm1.78$ & $0.2821\pm0.0296$ & $8.43\pm3.96$ \\
\textbf{SVC} & $\mathbf{0.7596\pm0.0455}$ & $\mathbf{37.80\pm1.42}$ & $\mathbf{5.73\pm2.52}$ & $\mathbf{0.3449\pm0.0158}$ & $\mathbf{15.23\pm4.99}$ \\
+ penalty & $0.7424\pm0.0454$ & $32.80\pm1.55$ & $4.78\pm2.10$ & $0.3158\pm0.0131$ & $10.00\pm3.90$ \\
Admission-only & $0.6301\pm0.0541$ & $24.82\pm3.19$ & $4.39\pm1.62$ & $0.1711\pm0.0495$ & $37.31\pm6.15$ \\
No-distance & $0.7948\pm0.0431$ & $56.03\pm1.31$ & $9.29\pm4.15$ & $0.3800\pm0.0209$ & $51.06\pm3.48$ \\
\end{longtable}
\endgroup
\par\medskip\noindent\begin{minipage}{\linewidth}\fontsize{9}{11}\selectfont\raggedright
Means $\pm$ sample SD over 7 trajectories. AUC: ranking of ancestor-labeled above unlabeled responses by advantage, among valid admitted responses with code hashes. $P_U$ and $P_D$: shares of positive advantage assigned to later-used and ancestor-labeled responses, respectively (percent). $\rho$: Spearman correlation between advantage and later-use count among valid responses. $P_0$: zero-task-advantage groups (percent of 500 groups). All eight mappings use the same SVC records. Bold identifies SVC, not column-wise best values.
\end{minipage}\par\smallskip

\begingroup
\fontsize{9}{11}\selectfont
\setlength{\tabcolsep}{3pt}
\renewcommand{\arraystretch}{1.0}
\setlength{\LTleft}{0pt}\setlength{\LTright}{0pt}
\setlength{\LTpre}{8pt}\setlength{\LTpost}{2pt}
\begin{longtable}{@{\extracolsep{\fill}}lrrrrr@{}}
\caption{Eight-mapping replay on Non-TSP SVC trajectories.}\label{tab:credit-replay-non_tsp}\\
\toprule
Mapping & AUC & $P_U$ (\%) & $P_D$ (\%) & $\rho$ & $P_0$ (\%) \\
\midrule
\endfirsthead
\multicolumn{6}{@{}l}{\small Table~\thetable{} (continued)}\\
\toprule
Mapping & AUC & $P_U$ (\%) & $P_D$ (\%) & $\rho$ & $P_0$ (\%) \\
\midrule
\endhead
\bottomrule
\endfoot
Native & $0.7641\pm0.0458$ & $27.53\pm1.53$ & $4.92\pm1.82$ & $0.2230\pm0.0325$ & $9.05\pm5.75$ \\
Validity-only & $0.5424\pm0.0476$ & $21.89\pm2.88$ & $3.41\pm1.38$ & $0.0384\pm0.0332$ & $69.02\pm5.40$ \\
Factorized & $0.7393\pm0.0371$ & $26.12\pm2.47$ & $4.28\pm1.61$ & $0.1998\pm0.0406$ & $7.77\pm3.91$ \\
Tail & $0.7524\pm0.0478$ & $33.81\pm1.97$ & $5.79\pm1.94$ & $0.2684\pm0.0240$ & $6.62\pm3.55$ \\
\textbf{SVC} & $\mathbf{0.7673\pm0.0524}$ & $\mathbf{37.45\pm1.91}$ & $\mathbf{6.35\pm2.54}$ & $\mathbf{0.3364\pm0.0256}$ & $\mathbf{13.44\pm5.04}$ \\
+ penalty & $0.7544\pm0.0516$ & $32.31\pm1.75$ & $5.43\pm2.25$ & $0.3070\pm0.0268$ & $8.86\pm3.88$ \\
Admission-only & $0.6063\pm0.0792$ & $25.92\pm2.54$ & $4.56\pm1.66$ & $0.1811\pm0.0534$ & $36.93\pm6.02$ \\
No-distance & $0.8087\pm0.0485$ & $56.91\pm3.76$ & $10.65\pm4.58$ & $0.3733\pm0.0291$ & $51.55\pm4.18$ \\
\end{longtable}
\endgroup
\par\medskip\noindent\begin{minipage}{\linewidth}\fontsize{9}{11}\selectfont\raggedright
Means $\pm$ sample SD over 21 trajectories. AUC: ranking of ancestor-labeled above unlabeled responses by advantage, among valid admitted responses with code hashes. $P_U$ and $P_D$: shares of positive advantage assigned to later-used and ancestor-labeled responses, respectively (percent). $\rho$: Spearman correlation between advantage and later-use count among valid responses. $P_0$: zero-task-advantage groups (percent of 500 groups). All eight mappings use the same SVC records. Bold identifies SVC, not column-wise best values. The 21 trajectories comprise seven seeds each of CVRP, OBP, and OP; this fixed-task descriptive summary does not represent 21 independent tasks.
\end{minipage}\par\smallskip

\begingroup
\fontsize{9}{11}\selectfont
\setlength{\tabcolsep}{3pt}
\renewcommand{\arraystretch}{1.0}
\setlength{\LTleft}{0pt}\setlength{\LTright}{0pt}
\setlength{\LTpre}{8pt}\setlength{\LTpost}{2pt}
\begin{longtable}{@{\extracolsep{\fill}}llrr@{}}
\caption{Paired replay contrasts on TSP records.}\label{tab:credit-paired-tsp}\\
\toprule
Comparator & Metric & SVC minus comparator & Descriptive 95\% interval \\
\midrule
\endfirsthead
\multicolumn{4}{@{}l}{\small Table~\thetable{} (continued)}\\
\toprule
Comparator & Metric & SVC minus comparator & Descriptive 95\% interval \\
\midrule
\endhead
\bottomrule
\endfoot
Native & AUC & $+0.0074$ & $[-0.0334,\ +0.0483]$ \\
Native & $P_U$ (pp) & $+8.83$ & $[+7.78,\ +9.89]$ \\
Native & $P_0$ (pp) & $+1.57$ & $[+0.75,\ +2.39]$ \\
Validity-only & AUC & $+0.3077$ & $[+0.2291,\ +0.3863]$ \\
Validity-only & $P_U$ (pp) & $+14.50$ & $[+12.50,\ +16.51]$ \\
Validity-only & $P_0$ (pp) & $-64.91$ & $[-72.09,\ -57.74]$ \\
Factorized & AUC & $+0.0559$ & $[-0.0003,\ +0.1120]$ \\
Factorized & $P_U$ (pp) & $+9.55$ & $[+7.93,\ +11.16]$ \\
Factorized & $P_0$ (pp) & $+1.94$ & $[+1.26,\ +2.62]$ \\
Tail & AUC & $+0.0282$ & $[+0.0169,\ +0.0394]$ \\
Tail & $P_U$ (pp) & $+2.50$ & $[+1.67,\ +3.33]$ \\
Tail & $P_0$ (pp) & $+2.60$ & $[+1.75,\ +3.45]$ \\
+ penalty & AUC & $+0.0074$ & $[-0.0033,\ +0.0181]$ \\
+ penalty & $P_U$ (pp) & $+3.34$ & $[+2.19,\ +4.48]$ \\
+ penalty & $P_0$ (pp) & $+1.91$ & $[+1.21,\ +2.62]$ \\
Admission-only & AUC & $+0.2687$ & $[+0.1881,\ +0.3493]$ \\
Admission-only & $P_U$ (pp) & $+12.46$ & $[+10.68,\ +14.24]$ \\
Admission-only & $P_0$ (pp) & $-39.14$ & $[-46.14,\ -32.14]$ \\
No-distance & AUC & $-0.0400$ & $[-0.0550,\ -0.0250]$ \\
No-distance & $P_U$ (pp) & $-18.79$ & $[-21.24,\ -16.35]$ \\
No-distance & $P_0$ (pp) & $-42.74$ & $[-45.53,\ -39.96]$ \\
\end{longtable}
\endgroup
\par\medskip\noindent\begin{minipage}{\linewidth}\fontsize{9}{11}\selectfont\raggedright
Differences pair mappings on the same trajectory. Positive AUC and later-use share differences mean larger values under SVC; a negative difference in the zero-task-advantage rate means more groups with nonzero task advantages. pp denotes percentage points. Student-$t$ intervals use six degrees of freedom for TSP and 20 for the fixed non-TSP collection. The non-TSP intervals do not support inference to a population of tasks.
\end{minipage}\par\smallskip

\begingroup
\fontsize{9}{11}\selectfont
\setlength{\tabcolsep}{3pt}
\renewcommand{\arraystretch}{1.0}
\setlength{\LTleft}{0pt}\setlength{\LTright}{0pt}
\setlength{\LTpre}{8pt}\setlength{\LTpost}{2pt}
\begin{longtable}{@{\extracolsep{\fill}}llrr@{}}
\caption{Paired replay contrasts on Non-TSP records.}\label{tab:credit-paired-non_tsp}\\
\toprule
Comparator & Metric & SVC minus comparator & Descriptive 95\% interval \\
\midrule
\endfirsthead
\multicolumn{4}{@{}l}{\small Table~\thetable{} (continued)}\\
\toprule
Comparator & Metric & SVC minus comparator & Descriptive 95\% interval \\
\midrule
\endhead
\bottomrule
\endfoot
Native & AUC & $+0.0031$ & $[-0.0114,\ +0.0177]$ \\
Native & $P_U$ (pp) & $+9.92$ & $[+8.94,\ +10.90]$ \\
Native & $P_0$ (pp) & $+4.39$ & $[+2.59,\ +6.19]$ \\
Validity-only & AUC & $+0.2248$ & $[+0.2006,\ +0.2491]$ \\
Validity-only & $P_U$ (pp) & $+15.56$ & $[+14.13,\ +16.99]$ \\
Validity-only & $P_0$ (pp) & $-55.58$ & $[-57.64,\ -53.52]$ \\
Factorized & AUC & $+0.0280$ & $[+0.0040,\ +0.0519]$ \\
Factorized & $P_U$ (pp) & $+11.32$ & $[+10.02,\ +12.63]$ \\
Factorized & $P_0$ (pp) & $+5.67$ & $[+4.31,\ +7.02]$ \\
Tail & AUC & $+0.0149$ & $[-0.0005,\ +0.0303]$ \\
Tail & $P_U$ (pp) & $+3.64$ & $[+2.76,\ +4.51]$ \\
Tail & $P_0$ (pp) & $+6.82$ & $[+5.55,\ +8.09]$ \\
+ penalty & AUC & $+0.0129$ & $[+0.0071,\ +0.0187]$ \\
+ penalty & $P_U$ (pp) & $+5.14$ & $[+4.74,\ +5.54]$ \\
+ penalty & $P_0$ (pp) & $+4.58$ & $[+3.91,\ +5.25]$ \\
Admission-only & AUC & $+0.1610$ & $[+0.1370,\ +0.1849]$ \\
Admission-only & $P_U$ (pp) & $+11.53$ & $[+10.51,\ +12.54]$ \\
Admission-only & $P_0$ (pp) & $-23.50$ & $[-27.04,\ -19.95]$ \\
No-distance & AUC & $-0.0414$ & $[-0.0517,\ -0.0310]$ \\
No-distance & $P_U$ (pp) & $-19.46$ & $[-20.92,\ -18.01]$ \\
No-distance & $P_0$ (pp) & $-38.11$ & $[-40.00,\ -36.23]$ \\
\end{longtable}
\endgroup
\par\medskip\noindent\begin{minipage}{\linewidth}\fontsize{9}{11}\selectfont\raggedright
Differences pair mappings on the same trajectory. Positive AUC and later-use share differences mean larger values under SVC; a negative difference in the zero-task-advantage rate means more groups with nonzero task advantages. pp denotes percentage points. Student-$t$ intervals use six degrees of freedom for TSP and 20 for the fixed non-TSP collection. The non-TSP intervals do not support inference to a population of tasks.
\end{minipage}\par\smallskip

\subsection{OBP plateau diagnostics}
\label{app:plateau}

A plateau is a maximal interval of at least 100 consecutive groups during which the best-so-far fitness reconstructed from valid generated responses remains exactly unchanged. Unlike the complete-search trajectories in \ref{app:results}, this reconstruction omits the initial population's best fitness and therefore describes stagnation in generated fitness. Within- and outside-plateau zero-task-advantage rates use their respective group counts as denominators. If a trajectory has no plateau, its within-plateau rate is undefined.

Table~\ref{tab:credit-obp-summary} supplements the main-text diagnostics with outside-plateau rates and Tail training records. Each condition contributes six defined within-plateau rates and seven outside-plateau rates, one per trajectory in each region. Training conditions have their own trajectories; Tail replay on SVC records changes only rewards and advantages.

Window-based diagnostics use five consecutive 100-group windows per trajectory to summarize valid-response counts, admission and fitness-repeat rates, new-best response counts, and zero-task-advantage rates. Fitness repetition requires an absolute difference below $10^{-7}$ from a pre-generation retained fitness. Repeat and admission rates divide by valid responses in the window; new-best counts include all flagged responses. Repeated fitness does not imply code copying or an absence of algorithmic diversity.

\begin{keeptable}
\caption{Mean OBP zero-task-advantage rates.}\label{tab:credit-obp-summary}
\fontsize{9}{11}\selectfont\setlength{\tabcolsep}{3pt}\renewcommand{\arraystretch}{1.0}
\begin{tabular*}{\linewidth}{@{\extracolsep{\fill}}llrrr@{}}
\toprule
Training records & Region & $n$ & Own (\%) & Tail replay (\%) \\
\midrule
SVC & Plateau & 6 & 11.80 & 4.04 \\
SVC & Other & 7 & 18.33 & 8.67 \\
Tail & Plateau & 6 & 7.33 & 7.33 \\
Tail & Other & 7 & 7.84 & 7.84 \\
\bottomrule
\end{tabular*}
\par\vspace{2pt}
\begin{minipage}{\linewidth}\fontsize{9}{11}\selectfont\raggedright
Each mean gives equal weight to the runs with a defined regional rate. SVC seed 12345 and Tail seed 314159 have no plateau. Each training condition therefore contributes six plateau rates and seven non-plateau rates. Differences between conditions reflect their own generated trajectories; replay on SVC records only changes the signal.
\end{minipage}
\end{keeptable}

\FloatBarrier
\clearpage
\begingroup
\setlength{\bibsep}{10pt}\renewcommand{\bibfont}{\small}
\bibliography{references}
\endgroup
\end{document}